\documentclass[10pt,twocolumn,letterpaper]{article}
\usepackage{iccv}
\usepackage[accsupp]{axessibility}  % Improves PDF readability for those with disabilities.
\usepackage{times}
\usepackage{epsfig}
\usepackage{graphicx}
\usepackage{amsmath}
\usepackage{amssymb}
\usepackage{color, colortbl}
\usepackage[dvipsnames]{xcolor}
\usepackage{array,multirow}
\usepackage{tikz}
\usepackage[normalem]{ulem}
\useunder{\uline}{\ul}{}
\usepackage{booktabs}
\usepackage{float}
\usepackage{adjustbox}
\usepackage{tabularx}
\usepackage{txfonts}
\usepackage{nicematrix}
\usepackage{mathtools} 
\usepackage{pifont}% http://ctan.org/pkg/pifont
\usepackage{enumitem}
\usepackage{subcaption}
\usepackage[normalem]{ulem}
\usepackage[first=0,last=9]{lcg}
\usepackage[dvipsnames]{xcolor}
\usepackage{graphicx}
\usepackage{caption}
\usepackage{amsmath}
\usepackage{array,multirow}
\usepackage{amssymb}
\usepackage{tikz}
\usepackage{ltablex}
\usepackage{supertabular}                           \usepackage{multicol}        
\usepackage{dblfloatfix} 
\newcommand{\Cooking}{\includegraphics[scale=0.027]{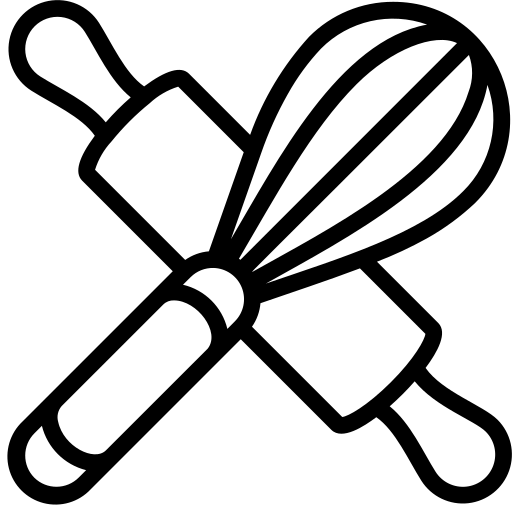}}%
\usepackage{tikz}
\newcommand{\Sewing}{\includegraphics[scale=0.027]{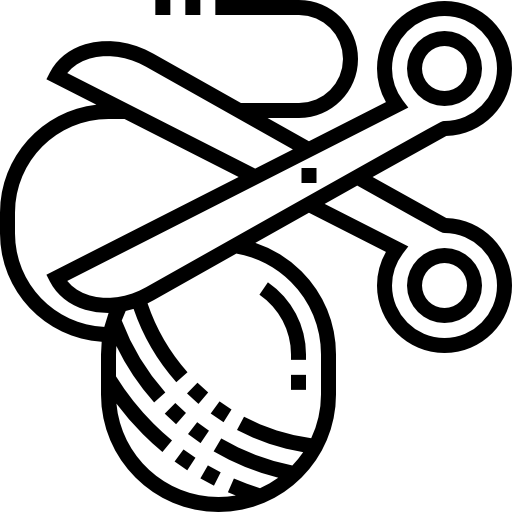}}%
\newcommand{\Mechanic}{\includegraphics[scale=0.027]{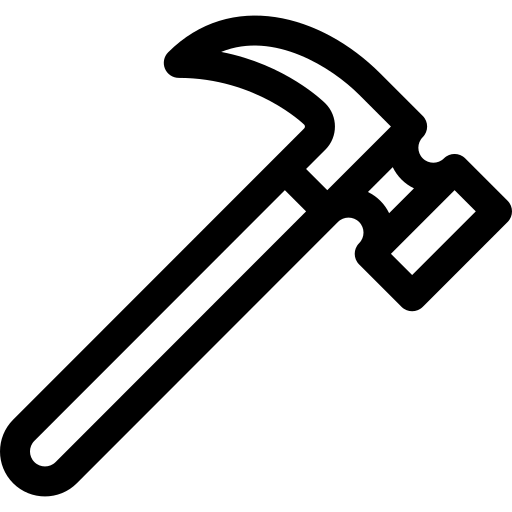}}%

\usepackage[normalem]{ulem}
\useunder{\uline}{\ul}{}
\usepackage{booktabs}
\usepackage{float}
\usepackage{adjustbox}
\usepackage{longtable}
\usepackage{tabularx}
\usepackage{txfonts}
\usepackage{mathtools} 
\usepackage{amsmath}
\usepackage{xcolor-material}
\usepackage{pgfplots}
\usepackage{pgfplots}
\usepackage{adjustbox}
\usepackage[accsupp]{axessibility} 

\usepackage{xcolor-material}
\usepackage{tikz}
\usepackage{amssymb}% http://ctan.org/pkg/amssymb
\usepackage{pifont}% http://ctan.org/pkg/pifont
\newcommand{\cmark}{\ding{51}}%
\newcommand{\xmark}{\ding{55}}%
\usepackage[normalem]{ulem}
\useunder{\uline}{\ul}{}
\usepackage{amssymb}
\usepackage[first=0,last=9]{lcg}

\definecolor{LightCyan}{rgb}{0.88,1,1}
\usepackage{enumitem}
\usepackage{color, colortbl}
\usepackage[pagebackref,breaklinks,colorlinks]{hyperref}
\usepackage{subfloat}
\definecolor{lightpink}{rgb}{1.0, 0.71, 0.76}
\definecolor{magenta(dye)}{rgb}{0.79, 0.08, 0.48}
\definecolor{lightblue}{rgb}{0.7, 0.77, 0.89}
\definecolor{lightblue}{rgb}{0.7, 0.77, 0.89}
\definecolor{cooking}{HTML}{CCA2D2}%cca2d25f
\definecolor{building}{HTML}{FAAED6} %faaed678
\definecolor{arts}{HTML}{96BFE0} %96bfe0ff
\definecolor{cleaning}{HTML}{DF9E79} %df9e79ff
\definecolor{mechanic}{HTML}{FFFF84} %ffff84ff
\definecolor{gardening}{HTML}{FFC082} %ffc082ff
\definecolor{playing}{HTML}{AED8A2} %a3d8a2ff
\definecolor{shopping}{HTML}{CDCDCD} %cdcdcdff
\definecolor{sport}{HTML}{F18687} %f18687ff
\definecolor{knitting}{HTML}{A8EEF5} %a8eef5ff
\newcommand{\dataset}{ARGO1M}
\newcommand{\method}{CIR}

\newcommand{\India}{\includegraphics[scale=0.17]{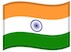}}%
\newcommand{\Building}{\includegraphics[scale=0.04]{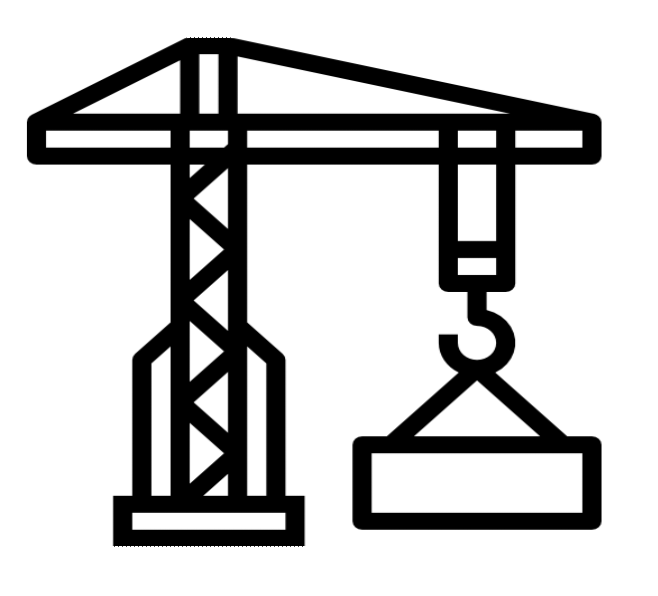}}%
\newcommand{\Saudi}{\includegraphics[scale=0.15]{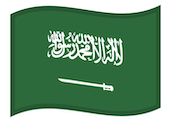}}%
\newcommand{\US}{\includegraphics[scale=0.15]{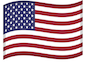}}%
\newcommand{\Knitting}{\includegraphics[scale=0.7]{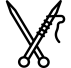}}%
\newcommand{\Sport}{\includegraphics[scale=0.7]{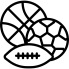}}%
\newcommand{\Shopping}{\includegraphics[scale=0.8]{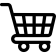}}%
\newcommand{\Cleaning}{\includegraphics[scale=0.9]{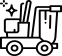}}%
\newcommand{\Playing}{\includegraphics[scale=0.7]{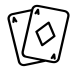}}%
\newcommand{\Gardening}{\includegraphics[scale=0.8]{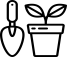}}%
\newcommand{\Japan}{\includegraphics[scale=0.15]{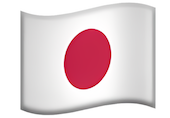}}%
\newcommand{\Italy}{\includegraphics[scale=0.15]{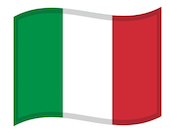}}%
\newcommand{\Colombia}{\includegraphics[scale=0.15]{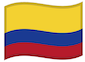}}%

\useunder{\uline}{\ul}{}
\definecolor{LightCyan}{rgb}{0.88,1,1}
\usepackage[capitalize]{cleveref}

\definecolor{lightpink}{rgb}{1.0, 0.71, 0.76}
\definecolor{magenta(dye)}{rgb}{0.79, 0.08, 0.48}
\definecolor{lightblue}{rgb}{0.7, 0.77, 0.89}
\definecolor{lightblue}{rgb}{0.7, 0.77, 0.89}

\crefname{section}{Section}{Secs.}
\Crefname{section}{Section}{Sections}
\Crefname{table}{Table}{Tables}
\crefname{table}{Table}{Tabs.}
\crefname{equation}{Eq.}{Tabs.}

\iccvfinalcopy % *** Uncomment this line for the final submission

\begin{document}

%%%%%%%%% TITLE
\title{What can a cook in Italy teach a mechanic in India?
\\Action Recognition Generalisation Over Scenarios and Locations}

\author{Chiara Plizzari\textsuperscript{$\clubsuit$}\thanks{Work carried during Chiara's research visit to the University of Bristol}
%{\tt\small chiara.plizzari@polito.it}
\and \quad
Toby Perrett\textsuperscript{$\vardiamondsuit$}
%{\tt\small toby.perrett@bristol.ac.uk}
\and \quad
Barbara Caputo\textsuperscript{$\clubsuit$}
%{\tt\small barbara.caputo@polito.it}
\and \quad
Dima Damen\textsuperscript{$\vardiamondsuit$}\\ \vspace*{-8pt} 
%{\tt\small dima.damen@bristol.ac.uk}
\and \textsuperscript{$\clubsuit$} Politecnico di Torino, Italy %\\
%{\tt\small {name.surname}@polito.it}
\and \textsuperscript{$\vardiamondsuit$}  University of Bristol, United Kingdom %\\
%{\tt\small {name.surname}@bristol.ac.uk} 
}

\maketitle
%\ificcvfinal\thispagestyle{empty}\fi

%%%%%%%%% ABSTRACT
\begin{abstract}
We propose and address a new generalisation problem: can a model trained for action recognition successfully classify actions when they are performed within a previously unseen scenario and in a previously unseen location? To answer this question, we introduce the Action Recognition Generalisation Over scenarios and locations dataset (\dataset), which contains 1.1M video clips from the large-scale Ego4D dataset, across 10 scenarios and 13 locations. We demonstrate recognition models struggle to generalise over 10 proposed test splits, each of an unseen scenario in an unseen location. We thus propose CIR, a method to represent each video as a Cross-Instance Reconstruction of videos from other domains. Reconstructions are paired with text narrations to guide the learning of a domain generalisable representation. We provide extensive analysis and ablations on \dataset\ that show \method\ outperforms prior domain generalisation works on all test splits. Code and data: \small{\url{https://chiaraplizz.github.io/what-can-a-cook/}}.
\end{abstract}
\vspace{-3mm}

\section{Introduction}
\label{sec:intro}

A notable distinction between human and machine intelligence is the ability of humans to generalise. We can see an example of the action ``cut'' performed by a cook in Italy, and recognise the same action performed in a different geographic \emph{location}, \eg India, despite having never visited. We can also recognise actions within new \textit{scenarios}, such as a mechanic cutting metal, even if we are unfamiliar with the tools they use.

This problem is known as domain generalisation~\cite{zhou2021domain}, where a model trained on a set of labelled data fails to generalise to a different distribution in inference. 
The gap between distributions is known as {\emph {domain shift}}.
% Domain Generalisation research aims to learn models which work when there is a shift between the domains seen in training and those encountered at inference.
% In general, this problem is referred to in the literature as \textit{domain shift}, meaning that a model trained on a set of source labelled data cannot generalize well on an unseen dataset, called target, when the latter belongs to a different distribution (or ``domain").
To date, works have focused on generalising over visual domain shifts~\cite{li2017deeper,torralba2011unbiased,munro2020multi,damen2018scaling,Planamente2022WACV}. %In order to demonstrate a more general understanding, models must be able to handle multiple shifts simultaneously.
In this paper, we introduce the \textit{scenario shift}, where the same action is performed as part of a different activity, impacting the tools used, objects interacted with, goals and behaviour.
We combine this with the location shift, generalising over both simultaneously. %In fact, cultural diversity has a huge impact not only on the appearance of different parts of the world, but also on the way people perform the same activity. E.g., in Italy, forks are used to spear pasta, eating, and hold food steady while cutting. On the other hand, in Japan, chopsticks are the primary utensil in a kitchen. %demonstrating a more general understanding by handling multiple shifts simultaneously.

\begin{figure}[t!]
    \centering
    \includegraphics[width=\columnwidth]{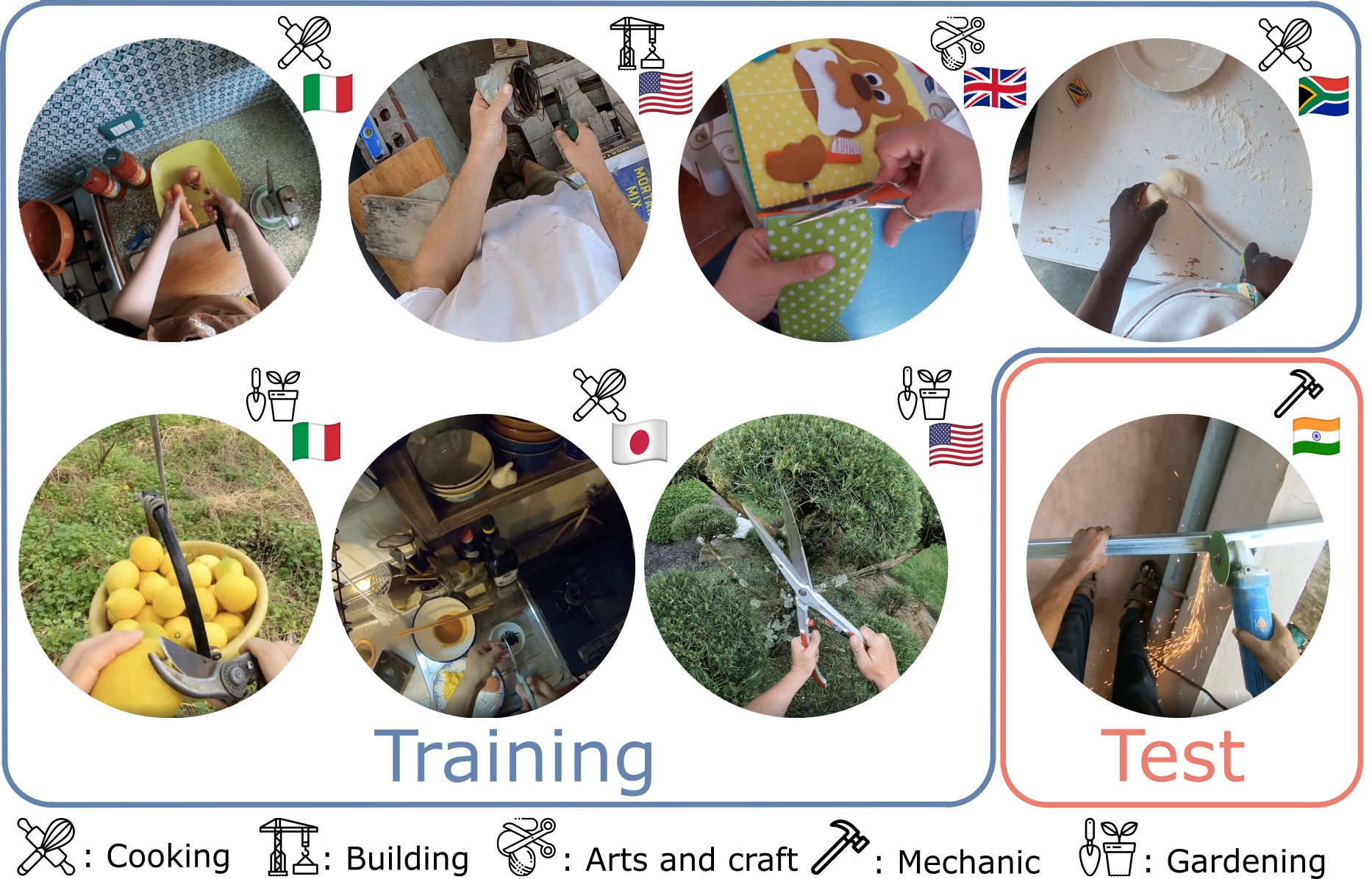}
    \caption{
    %\textbf{Top:} We introduce \dataset, a dataset for action recognition generalisation.
    Problem statement and samples from the \dataset\ dataset. The same action, e.g. ``cut", is performed differently based on the \textit{scenario} and the \textit{location} in which it is carried out. We aim to generalise so as to recognise the same action within a new scenario, \textit{unseen} during training, and in an \textit{unseen location}, \eg, \textit{Mechanic}~(\Mechanic) in \emph{India} (\India).  }%, in order to capture shared fine-grained information across domains. }
    \vspace*{-16pt}
    \label{fig:teaser}
    
\end{figure}

In Fig.~\ref{fig:teaser}, the action ``cut' is performed using a {knife} whilst cooking (\Cooking), {pliers} whilst building (\Building) and {scissors} for arts and crafts (\Sewing).
Tools are not specific for a scenario and can vary over locations -- e.g. in Fig.~\ref{fig:teaser}, seaweed sheets are cut with scissors while cooking in Japan.
Generalising would be best achieved by learning the notion of ``cutting'' as separating an object into two or more pieces, regardless of the tool or background location. 
Successful generalisation can thus enable recognising metal being ``cut'' by a mechanic in India using an angle grinder (Fig.~\ref{fig:teaser} Test).
%Generalising across scenarios has not been attempted before.

% learn models which can cope with the domain shift between training data 
% In general, this problem is referred to in the literature as \textit{domain shift}, meaning that a model trained on a set of source labelled data cannot generalize well on an unseen dataset, called target, when the latter belongs to a different distribution (or ``domain").
%Domain generalisation methods aim to learn models with this deeper level of understanding~\cite{zhou2021domain}.

% The aim of this paper is to generalise action understanding to new scenarios, locations, or both. 
% %Whilst there can be large appearance differences due to location, perhaps the biggest challenge 
% %is understanding the huge variety of ways in which many humans can perform the same action in different scenarios. 
% New scenarios impact the tools used, objects being interacted with, goals and behaviour.
% Generalising across scenarios has not been attempted before.
Our investigation is enabled by the recent introduction of the Ego4D~\cite{grauman2022ego4d} dataset of egocentric footage from around the world. 
%\tobya{We curate a subset called the Action Recognition Generalisation Over scenarios and locations (\dataset) dataset, which contains actions in 10 scenarios and 13 locations, shown in Fig.~\ref{fig:teaser} (top). We define a single training set, and 7 distinct test sets which assess the ability to generalise to different scenario and/or location shift combinations.}
We curate a setup specifically for action generalisation, called \dataset. It contains 1.1M action clips of 60 classes from 73  unique scenario/location combinations. %, forming the largest domain generalisation dataset for images or videos.
% We select 10 test splits, totalling 111K clips, with corresponding training that excludes all samples from the tested scenarios and locations. We showcase the challenge of generalising over scenarios and locations by analysing the impact of excluding these samples from training.

To tackle the challenge of \dataset, we propose a new method for domain generalisation. We represent each video as a weighted combination of other videos in the batch, potentially from other domains. We refer to this as Cross-Instance Reconstruction (\method). 
Through reconstruction, the method learns domain generalisable video features.
\method\ is supervised by a classification loss and a video-text association loss.
%\dN{Toby's next sentence assumes we return the teaser figure of method - we need to try with/without} 
%\tobya{In Fig.~\ref{fig:teaser} (bottom) we demonstrate the reconstruction of a video from \textit{gardening} using videos from the \textit{shopping}, \textit{cooking}, and \textit{sport} scenarios recorded at different locations. The reconstructed video~$\oplus v$ is paired with the text narration during training.}
%CIR outperforms previous domain generalisation methods by a large margin, showing a consistent improvement on generalisation performance across all the proposed test splits.  
\noindent To summarise, our key contributions are:
\vspace*{-6pt}
\begin{itemize}[leftmargin=3.5mm,itemsep=-1.5ex,partopsep=1ex,parsep=2ex]
   \item We curate the Action Recognition Generalisation dataset (\dataset) from videos and narrations from Ego4D. \dataset\ is the first to test action generalisation across both scenario and location shifts, and is the largest domain generalisation dataset across images and video.
    \item We introduce \method, a domain generalisation method which exploits Cross-Instance Reconstruction and video-text pairing to learn generalisable representations.
    \item We test \method\ on the proposed \dataset, showing that it consistently outperforms baselines and recent domain generalisation approaches on 10 test sets. 
\end{itemize}

\section{Related Work}
\label{sec:related}

%\subsection{Domain Generalisation \textit{vs} Domain Adaptation}

In this section, we review datasets and methods for Domain Generalisation. \textbf{Domain Generalisation (DG)} aims to generalise to any unseen target domain, where data from the target domain are not available during training~\cite{zhou2021domain}. We note the distinction from the \textbf{Domain Adaptation} setting, where unlabelled target samples are available during training~\cite{munro2020multi,Song2021CVPR,Kim2021ICCV}. Adaptation is out of scope for this paper. %This knowledge is provided by  features, allowing target distribution properties to be inferred.

% In this section, we review datasets and methods for \textbf{Domain Generalisation (DG)}, which aims to generalise to any unseen target domain, where data from the target domain are not available during training~\cite{zhou2021domain}. \textbf{Domain Adaptation}, where 
% unlabelled target samples are available during training~\cite{munro2020multi,Song2021CVPR,Kim2021ICCV}, is beyond the scope of this paper.

\subsection{Domain Generalisation (DG) datasets}

% \tobya{
% \cref{tab:datasets} presents a comparison of datasets used for domain generalisation, across images and video.
% Existing image datasets present a stylistic shift. 
% For example, PACS~\cite{li2017deeper}, Office-Home~\cite{venkateswara2017deep} and DomainNet~\cite{peng2019moment} include common objects in photos, paintings, clipart, cartoons and sketches. 
% % For example, common objects in photos, paintings, clipart, cartoons and sketches \cite{li2017deeper,venkateswara2017deep,peng2019moment}.
% Location shift was explored in~\cite{beery2018recognition} which contains animals photographed in different locations. 
% % Other datasets do not specify the shift, such as
% VLCS~\cite{torralba2011unbiased} tests recognition of common categories across datasets such as VOC~\cite{everingham2010pascal} and Sun~\cite{xiao2010sun}.  
% Image DG works typically test on a number of these benchmarks~\cite{gulrajani2020search}. %, %with no 
% }

%\vspace*{-3pt}
\cref{tab:datasets} presents a comparison of vision datasets used for domain generalisation.
Existing image datasets present a stylistic shift. 
% For example, PACS~\cite{li2017deeper}, Office-Home~\cite{venkateswara2017deep} and DomainNet~\cite{peng2019moment} include common objects in photos, paintings, clipart, cartoons and sketches. 
For example, common objects in photos, paintings, clipart, cartoons and sketches~\cite{li2017deeper,venkateswara2017deep,peng2019moment}, or common categories across datasets~\cite{torralba2011unbiased}.
Location shift was explored in~\cite{beery2018recognition} which contains animals photographed in different locations. 
% Other datasets do not specify the shift, such as
Image DG works typically test on a number of these benchmarks~\cite{gulrajani2020search}.
% For video, datasets noted various shifts including cross-dataset (UCF-HMDB)~\cite{chen2019temporal}, synthetic-to-real (Kinetics-Gameplay)~\cite{chen2019temporal}, viewpoint~\cite{choi2020unsupervised}, location~\cite{munro2020multi} and time~\cite{damen2020rescaling}.
For video, shifts include cross-dataset~\cite{chen2019temporal}, synthetic-to-real~\cite{chen2019temporal}, viewpoint~\cite{choi2020unsupervised}, location~\cite{munro2020multi} and the passage of time~\cite{damen2020rescaling}.

Compared to prior works, \dataset\ is $21 \times$ %\dN{avoid approx, if it's 19 say 19, if it's 19.5 say 20} 
the largest video DG dataset and $1.8 \times$ %\dN{again avoid approx} 
image DG dataset. Importantly, \dataset\ introduces the scenario shift, which it tests alongside the location shift, with many more domains ({up to 64 training domains and 10 test domains}).

\begin{table}[t]
\centering
\begin{adjustbox}{width=\columnwidth, margin=0ex 1ex 0ex 0ex}
\begin{tabular}[width=\textwidth]{llrrcrcc}
\toprule\noalign{\smallskip}
& &\multicolumn{2}{c}{Samples} &&\multicolumn{3}{c}{Domains}\\ \cline{3-4} \cline{6-8}
       &   Dataset   & \# Samples  & \# Cls  &  & \# Train    & \# Test           &   Domain Shift \\ \hline
\multirow{5}{*}{\rotatebox{90}{Images}} &PACS ~\cite{li2017deeper}                                   & 9,991      & 7  &  & 3  & 4                        &   Style \\
& VLCS ~\cite{torralba2011unbiased}     & 10,729   & 5  & & 3                              & 4                           &   N/A \\
& OfficeHome ~\cite{venkateswara2017deep}    & 15,588    & 65  &                             & 3 & 4                        &   Style   \\
& TerraIncognita ~\cite{beery2018recognition}    & 24,788     & 10 &                        & 3 & 4                        & Loc  \\
& DomainNet   ~\cite{peng2019moment}   & 586,575    & 345 & & 5 & 6                         &   Style \\ \hline
\multirow{5}{*}{\rotatebox{90}{Videos}} &UCF-HMDB~\cite{chen2019temporal}   & 3809  & 12 & & 1 & 2  & N/A \\
&Kinetics-Gameplay~\cite{chen2019temporal}                                   & 49,998     & 30  &    & 1 & 2                    &   Realism    \\
&MM-SADA~\cite{munro2020multi} & 10,094 & 8 & & 2 & 3  & Loc\\

&EPIC-Kitchens~\cite{damen2020rescaling}       & 48,139   & 86 &      & 11 & 1            & Time Gap \\ 
&\textbf{\dataset}              &  1,050,371     & 60   &       & 54-64 & 10                         & (Scenario, Loc) \\
\bottomrule
\end{tabular}
\end{adjustbox}
\vspace*{-12pt}
\caption{\textbf{Datasets for DG. }\dataset\ tests combined scenario and location shifts, and is the largest in \# of samples \& \# of domains.}% \dN{As discussed with Chiara separate into train and test domains and set the range for us with 10 being test. We never test 73 domains we should avoid saying that incorrectly.}}
\label{tab:datasets}
\vspace*{-8pt}
\end{table}
%This means it tests the more difficult and general problem of how well a model trained on actions performed whilst cooking in Italy, for example, can recognise actions performed by a mechanic in India.
%\subsection{Domain generalisation datasets}
%say we're covering generalisation and adaptation here, as you can test generalisation on an adaptation dataset

%\subsubsection{Image datasets}
%Stylistic shift, doesn't really take into account the semantics
%Only one shift 

%\subsubsection{Video datasets}
%All of these are based around one shift, or same environment (EK).
%What about UCF-HMDB (which shift is in it?)

%\tobyn{Not sure if we need this paragraph:} Note that there are two distinct domain shift problems - domain adaptation, where the whole test set is seen without labels, and domain generalisation, where only one test sample is seen at a time. \dataset contains splits for generalisation as it is a more realistic and challenging problem, but the scenarios we curate could easily be used to test adaptation.

\subsection{Domain Generalisation (DG) Methods}

%\vspace*{-3pt}
Previous approaches for DG are mostly designed around image data~\cite{carlucci2019domain,volpi2018generalizing,li2018domain,dou2019domain,li2018deep,bucci2020selfsupervised}.
\textit{Feature-based alignment} between training domains can be used to learn domain-invariant representations~\cite{li2018domain,sun2016deep,ganin2016domain,yang2022multi}. This can be achieved using a domain-adversarial network~\cite{ganin2016domain} or by minimising distances such as Maximum Mean Descrepancy (MMD)~\cite{li2018domain,gretton2012kernel}.
This has recently been extended in~\cite{yang2022multi}, which handles class and domain imbalance with a weighted loss. 
\textit{Data-based} methods augment training data to prevent overfitting~\cite{volpi2018generalizing,volpi2019addressing,zhou2021domain,zhang2022exact,chen2022compound,nam2021reducing,chen2022mix,wang2020heterogeneous,xu2020adversarial}. For example, data augmentation such as Mixup \cite{zhang2017mixup} has been shown to improve accuracy on unseen data.
\textit{Meta-Learning} methods simulate the distribution shift between seen and unseen environments~\cite{li2018learning,balaji2018metareg,dou2019domain,li2019episodic,li2019feature} using meta-train and meta-test domains.
\textit{Self-Supervision} \cite{carlucci2019domain,bucci2020selfsupervised} has been shown to learn generalisable representations, with unsupervised pretext tasks better capturing the shared knowledge among multiple sources.
%\tobyn{I think we need more on language, clip style association and how it helps generalise as it's a big part of our method.}
A recent trend is to learn \textit{domain prompts} from visual~\cite{zheng2022prompt,shu2022test} or text information~\cite{niu2022domain,zhang2021amortized}, or utilise \textit{cross-modal supervision}~\cite{min2022grounding}. For example,
DoPrompt~\cite{zheng2022prompt} %learns domain-specific prompts and trains a prompt adapter to generate a combination of these for each training image. The adapter is then used at test time to integrate knowledge from the source domains for each target image. % by generating a combined prompt for each target image.
learns training domain-specific prompts, and predicts prompts for test samples as linear combinations of training prompts.
There are limited works on video domain generalisation. \cite{Planamente2022WACV} relies on multi-modal alignment, and~\cite{yao2021videodg} uses adversarial data
augmentation. 
% VideoDG~\cite{yao2021videodg} is a DG framework for third person action recognition, which proposes an AdversarialPyramid Network performing generalisation on video features by capturing the local-relation, global-relation, and cross-relation features progressively.

% In this work, we rely on language to help the network capture semantic information in egocentric action recognition, motivated by the recent success of multi-modal training of images and language~\cite{radford2021learning}. Due to the wide diversity of objects and tools used to carry out the actions, text narrations in this context do indeed provide rich clues for fine-grained action understanding.

%These works were only tested with up to 11 different domains in training. %\dN{We agreed to change table 1 to number of training domains. EPIC should be 1 train and 1 test, so largest would be DomainNet with 5, but please check for correctness and if you agree with the shift. Primarily, avoiding to say others test one type of shift.}
For our comparative analysis, we extend a representative selection of prior works \cite{wang2020heterogeneous,li2018domain,sun2016deep,ganin2016domain,yang2022multi,zheng2022prompt} to the large number of training domains in \dataset, and showcase their limitation experimentally.
% Differently from these works, we reconstruct each
% video as a weighted combination of videos from various domains, which makes the most of the large number of training domains and samples in \dataset. 
% We exploit action narrations, which are frequently available with video datasets, to help learn generalisable representations. %Note that narrations lack the required detail for grounding as in \cite{min2022grounding}. Instead, we use narrations to help to guide the reconstruction learning.
% Authors of ~\cite{min2022grounding} also propose to use text. They use it to train for a grounding task by using fine-grained class-level textual descriptions specifically for the purpose of extracting domain-invariant primitives. 

%exploit the availability of a significantly higher number of training domains to obtain rich reconstructions without using domain labels. Instead, we rely on action labels and text narrations.  

%\dN{Not yet totally happy with the above}

\subsection{Cross-Attention for Reconstruction}

%\vspace*{-3pt}
The task of predicting masked tokens within one video is now common in many representation learning approaches, e.g.~\cite{feichtenhofer2022masked}.
We differ from these works in reconstructing from other videos in the batch.
Such cross-instance attention has been used to reconstruct query instances from examples of each class for few-shot learning~\cite{doersch2020crosstransformers,perrett2021temporal}.
In~\cite{perrett2023}, few-shot instances are reconstructed from samples of head classes.
% In~\cite{doersch2020crosstransformers}, cross-attention is used to reconstruct a query image using images from each class separately. \cite{perrett2021temporal} extends this idea to cross-attention between temporally-ordered sub-sequences. 
In cross-modal retrieval~\cite{patrick2020support}, reconstruction through cross-attention learns better video-text representations through a caption generation task.
Differently from prior works, we reconstruct each
video as a learned weighted combination of videos from \textit{various domains}. 

\begin{figure}[t]
    \centering
    \includegraphics[width=\columnwidth]{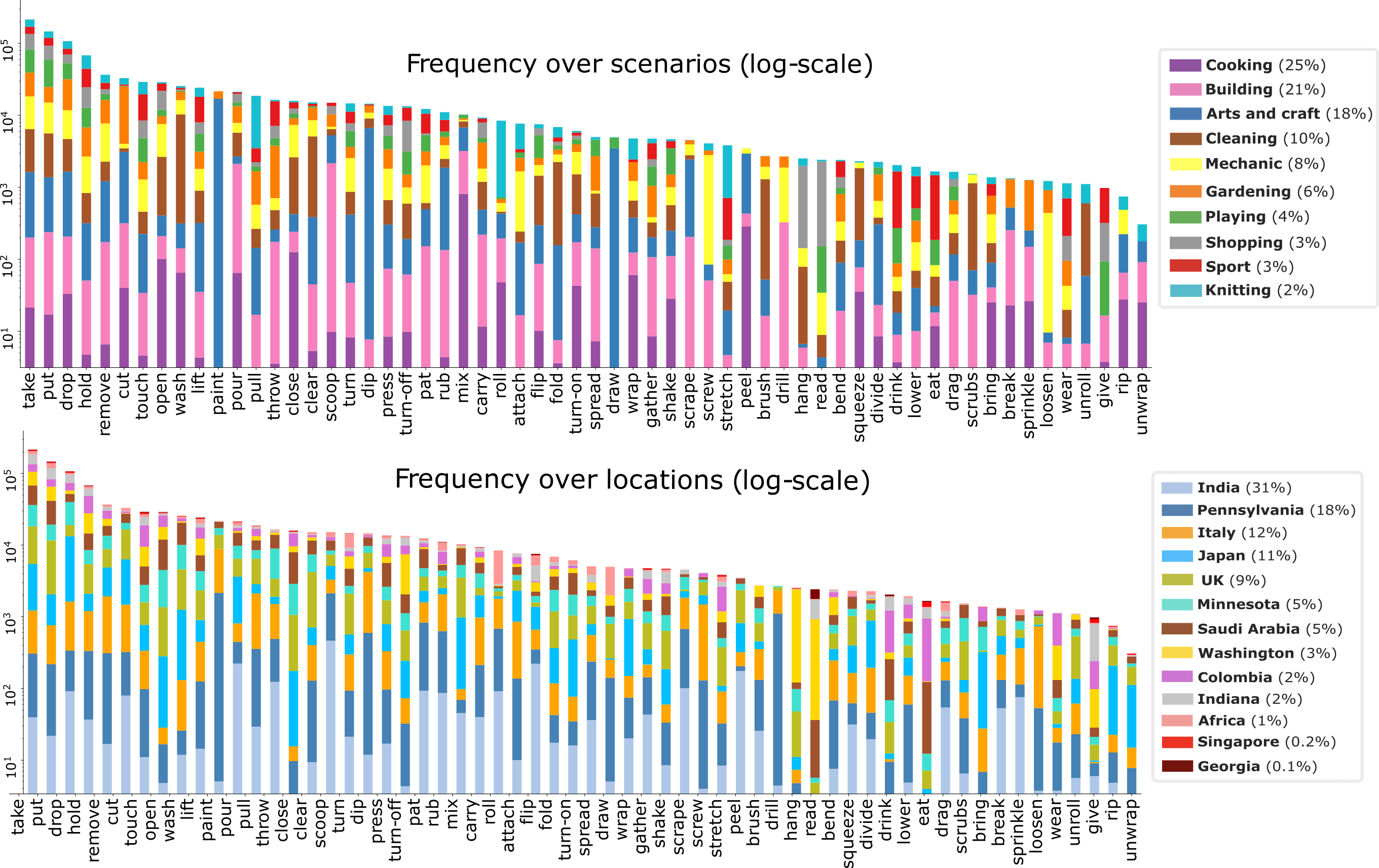}
    \vspace*{-20pt}
    \caption{Frequency (log-scale) of the 60 classes in \dataset\ across scenarios (top) and locations (bottom) - \% in legend. Scenarios and locations are linearly scaled within each bar. } % and actions (right). We highlight 2 corresponding patches. \textcolor{CadetBlue}{\textbf{(A)}}: features from one location (India) covering multiple scenarios. \textcolor{Green}{\textbf{(B)}}: features from one scenario (Playing), recorded in many locations. The action U-MAP (right) shows features of the same action are spread over scenarios and locations.} % to showcase that features are grouped for certain locations \textcolor{CadetBlue}{\textbf{(A)}} but from multiple scenarios at times \textit{vs} groupings by scenarios from different locations \textcolor{Green}{\textbf{(B)}}. Note that our target is action recognition. The action U-MAP (right) shows the problem complexity where action samples are spread over scenarios and locations.} % -- \ie features are more similar to others from the same scenario/location than examples of one action. }
    \vspace*{-8pt}
    \label{fig:distribution}
\end{figure}

\section{\dataset\ Benchmark}
\label{sec:dataset}

%To test action recognition generalisation, our training set contains common \dd{actions across scenarios and locations, and our test splits contains unseen scenarios and locations.}
In this section, we detail how we curated the \dataset\ dataset from videos of the Ego4D~\cite{grauman2022ego4d} dataset.

\noindent\textbf{Ego4D Background.} Ego4D~\cite{grauman2022ego4d} contains untrimmed egocentric videos totaling 3,670 hours collected from 8 non-US countries and 5 US states. These
%with 7 different devices, and 
represent a variety of daily life scenarios (\eg playing cards, cooking, fixing the car). 
Each video is associated with metadata reflecting the geographic location and the scenario it captures.
%Most videos capture one scenario, with 14.9\% marked as containing multiple scenarios.
Within each video, timestamp-level narrations of actions are provided. %\footnote{We focus on the first set of narrations throughout (2,509,565 in total).}. 

%Those describe the camera wearer’s actions and interactions with objects. For example, the narration “\#C C puts the scraper down” with the timestamp 3.70s.  

\noindent\textbf{\dataset\ Metadata.} 
% \tobya{While scenarios captured per video is provided in Ego4D, these are free-form, at times missing, and unstructured. Examples of those are ``Gardening", ``Gardener", ``Farmer", ``Car mechanic", ``Getting the car fixed". We exclude repetitive scenarios such as ``talking'' or ``on a screen'', as well as videos with missing or multiple scenarios.  We then manually cluster the free-form descriptions into 10 distinct scenarios. These are: \textit{Cooking}, \textit{Building}, \textit{Arts and craft}, \textit{Cleaning}, \textit{Mechanic}, \textit{Gardening}, \textit{Playing}, \textit{Shopping}, \textit{Sport}, \textit{Knitting}. For instance,  ``Gardening", ``Gardener", ``Gardener" will all fall under \textit{Gardening} scenario, and ``Cooking",  ``Car mechanic", ``Getting the car fixed" under \textit{Mechanic} scenario. All details about the grouping are in the Supplementary.
% }
The high-level scenario descriptions in Ego4D are free-form and at times missing. We exclude repetitive scenarios such as ``talking'' or ``on a screen'', as well as videos with missing or multiple scenarios. We then manually cluster the free-form descriptions into 10 scenarios. These are: \textit{Cooking} (\Cooking), \textit{Building} (\Building), \textit{Arts and crafts} (\Sewing), \textit{Cleaning} (\Cleaning), \textit{Mechanic} (\Mechanic), \textit{Gardening} (\Gardening), \textit{Playing} (\Playing), \textit{Shopping} (\Shopping), \textit{Sport} (\Sport), \textit{Knitting}~(\Knitting). As an example, the free-form descriptions ``Car mechanic", ``Getting the car fixed" and ``Bike mechanic'' are clustered into  \textit{Mechanic}.

%\begin{figure}[t!]
%    \centering
%    \includegraphics[width=\columnwidth]{imgs/pdfs/umaps.png}
%    \vspace*{-12pt}
%    \caption{U-MAPs of \dataset\ features across scenarios (left) and locations (right). %Color code is the same as Fig. \ref{fig:distribution}.} % and actions (right). We highlight 2 corresponding patches. \textcolor{CadetBlue}{\textbf{(A)}}: features from one location (India) covering multiple scenarios. \textcolor{Green}{\textbf{(B)}}: features from one scenario (Playing), recorded in many locations. The action U-MAP (right) shows features of the same action are spread over scenarios and locations.} % to showcase that features are grouped for certain locations \textcolor{CadetBlue}{\textbf{(A)}} but from multiple scenarios at times \textit{vs} groupings by scenarios from different locations \textcolor{Green}{\textbf{(B)}}. Note that our target is action recognition. The action U-MAP (right) shows the problem complexity where action samples are spread over scenarios and locations.} % -- \ie features are more similar to others from the same scenario/location than examples of one action. }
%    \vspace*{-8pt}
 %   \label{fig:umaps}
%\end{figure}

Similarly, while text narrations offer the ground-truth for the action in each video clip, they are also free-form sentences. 
%For example, the verbs: ``cut'' and ``chop'' appear in various narrations but all represent the same action.
We extract action labels by parsing the text narrations using spaCy~\cite{spacy2}. 
We take verbs as actions and convert these to closed-vocabulary classes, using modified clustering from~\cite{damen2018scaling} for the additional vocabulary.
We have 60 action classes shown in Fig. \ref{fig:distribution}. The distribution is long-tailed, and each action class appears in multiple scenarios and in multiple locations. On average each class appears in 8 scenarios and 11 locations.
%Actions distribute unevenly across the two, with some classes appearing only in some scenarios and/or locations, e.g., ``paint" appears only in \textit{Building}, \textit{Arts and Craft}, and \textit{Gardening}, and ``drill" appears only in 4 locations.

%More details on the Ego4D dataset are provided in~\cite{grauman2022ego4d}.
In summary, \dataset\ contains 1,050,371 video clips. 
% Each clip is associated with an action class (out of 60), a scenario (out of 10) and a geographic location (out of 13). Comparisons against other datasets are in Table~\ref{tab:datasets}.
% In \dataset, 
Each video {\emph{clip}} is captured in a given {\emph {scenario}} (out of 10) and geographic {\emph {location}} (out of 13), with associated {\emph {text narration}} and {\emph{action class}} (out of 60).
For example, the caption, ``\#Camera wearer (C) cuts the lemon strand.'' is associated to a clip recorded in ``Italy" and capturing ``Gardening" scenario, with associated action label ``cut".

\noindent \textbf{\dataset\ Splits.}\label{sec:splits} 
We curate 10 distinct train/test splits to evaluate generalisation over scenarios and locations.
We select these 10 test splits so \textit{all scenarios} are covered. 
For each scenario, we select the location with the largest number of samples to form the test split for robust evaluation.
Given paired scenario and location \textbf{(Sc, Lo)}, the corresponding training split excludes all samples from the scenario \textbf{(Sc)} as well as all samples from the location \textbf{(Lo)}. We show later in this section that these 10 splits present a variety of combined scenario/location shift properties.

% with one unique location per scenario based on the number of samples to }
%In order to evaluate the generalisation ability of our approach to both scenarios and locations, we follow the standard leave-one-domain-out protocol \cite{gulrajani2020search}. 
%Specifically, we create ten distinct test splits by excluding one unique \textit{(scenario, location)} combination, one for each of the ten available scenarios. We select the locations  by prioritizing selecting the ones with sufficient sample sizes to ensure a robust evaluation, while also taking into account the diversity of the selected locations, aiming to cover as many available domains as possible. In total, we cover 8/13 available locations. 
The selected test splits and their [number of samples] are:  \textit{Gardening} in \textit{Pennsylvania} (\textbf{Ga, US-PNA\footnote{We use ISO country codes and US state codes.}}) [16,410],  \textit{Cleaning} in \textit{Minnesota} (\textbf{Cl, US-MN}) [22,008], \textit{Knitting} in \textit{India} (\textbf{Kn, IND}) [13,250], \textit{Shopping} in \textit{India} (\textbf{Sh, IND}) [11,239], \textit{Building} in \textit{Pennsylvania} (\textbf{Bu, US-PNA}) [99,865], \textit{Mechanic} in \textit{Saudi Arabia} (\textbf{Me, SAU}) [11,700], \textit{Sport} in \textit{Colombia} (\textbf{Sp, COL}) [16,453], \textit{Cooking} in \textit{Japan} (\textbf{Co, JPN}) [82,128], \textit{Arts and crafts} in \textit{Italy} (\textbf{Ar, ITA}) [36,812], \textit{Playing} in \textit{Indiana} (\textbf{Pl, US-IN}) [17,379]. 
%\dN{We need: 1) logos of scenarios and flags for each and 2) number of test samples in each case}

%In each of these test splits, both the scenario and location are completely new, or ``unseen", as we exclude any other examples from that particular scenario or location from the training set. 
%All the remaining scenario and location combinations are used for training. We additionally provide a \textbf{validation split} of 10\% of the training clips to evaluate performance when both the scenario and the location are present in the training set. 
% Additional split details are in the Supplementary. 
%\dN{We might need this table here rather than in supplementary}

\begin{figure}[t!]
\begin{subfigure}[t]{0.46\columnwidth}
\includegraphics[width=1\columnwidth]{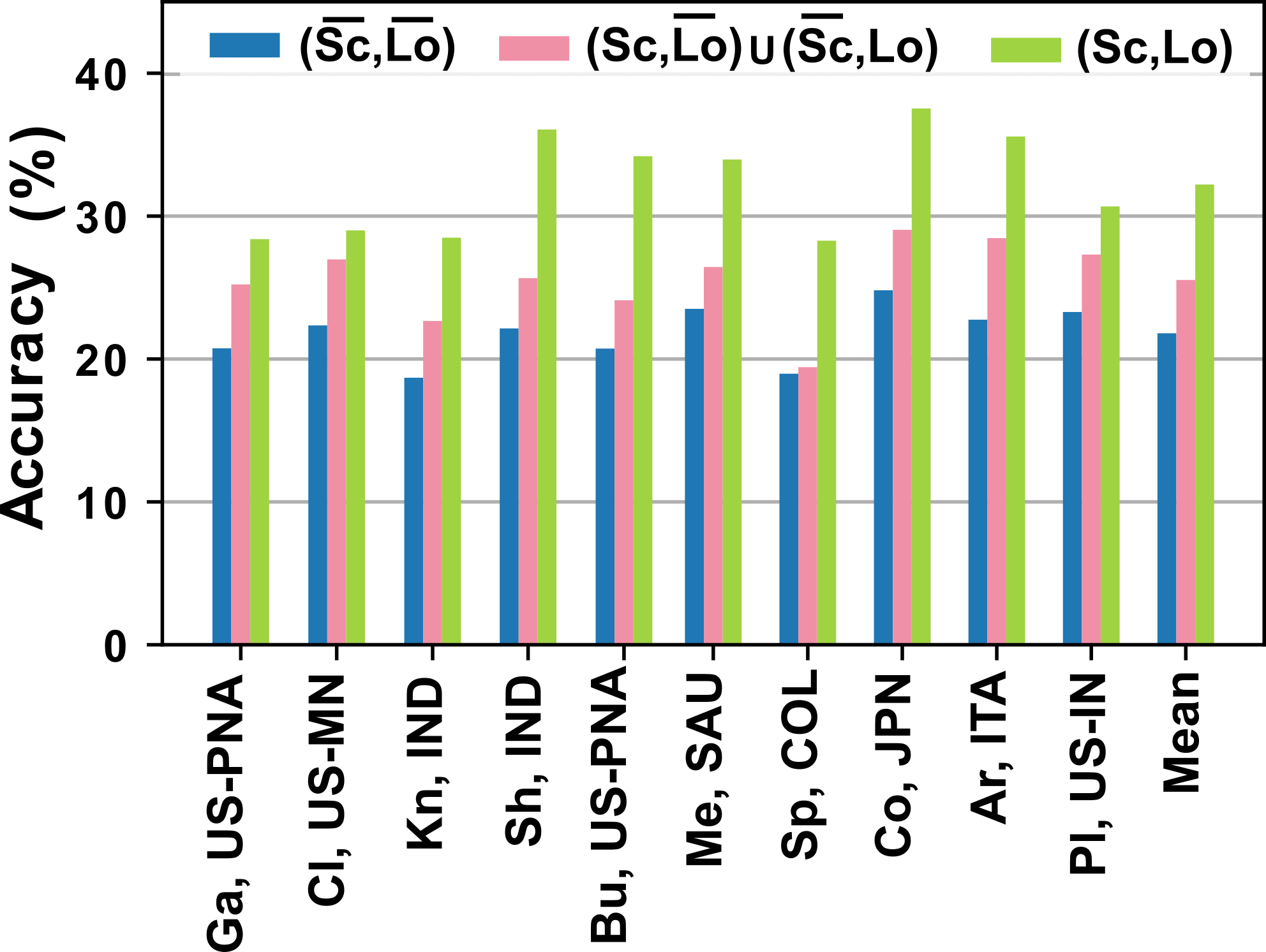}
\caption{Accuracy without samples from the test scenario or location {\textbf{($\overline{\text{Sc}}$, $\overline{\text{Lo}}$)}} as well as {\textbf{(Sc, $\overline{\text{Lo}}$)}}$\cup${\textbf{($\overline{\text{Sc}}$, Lo)}} and {\textbf{(Sc Lo)}}.}\label{fig:whole_drops}
\end{subfigure}\hfill% 
\hspace{0.03\columnwidth}
\begin{subfigure}[t]{0.50\columnwidth}
\includegraphics[width=1\columnwidth]{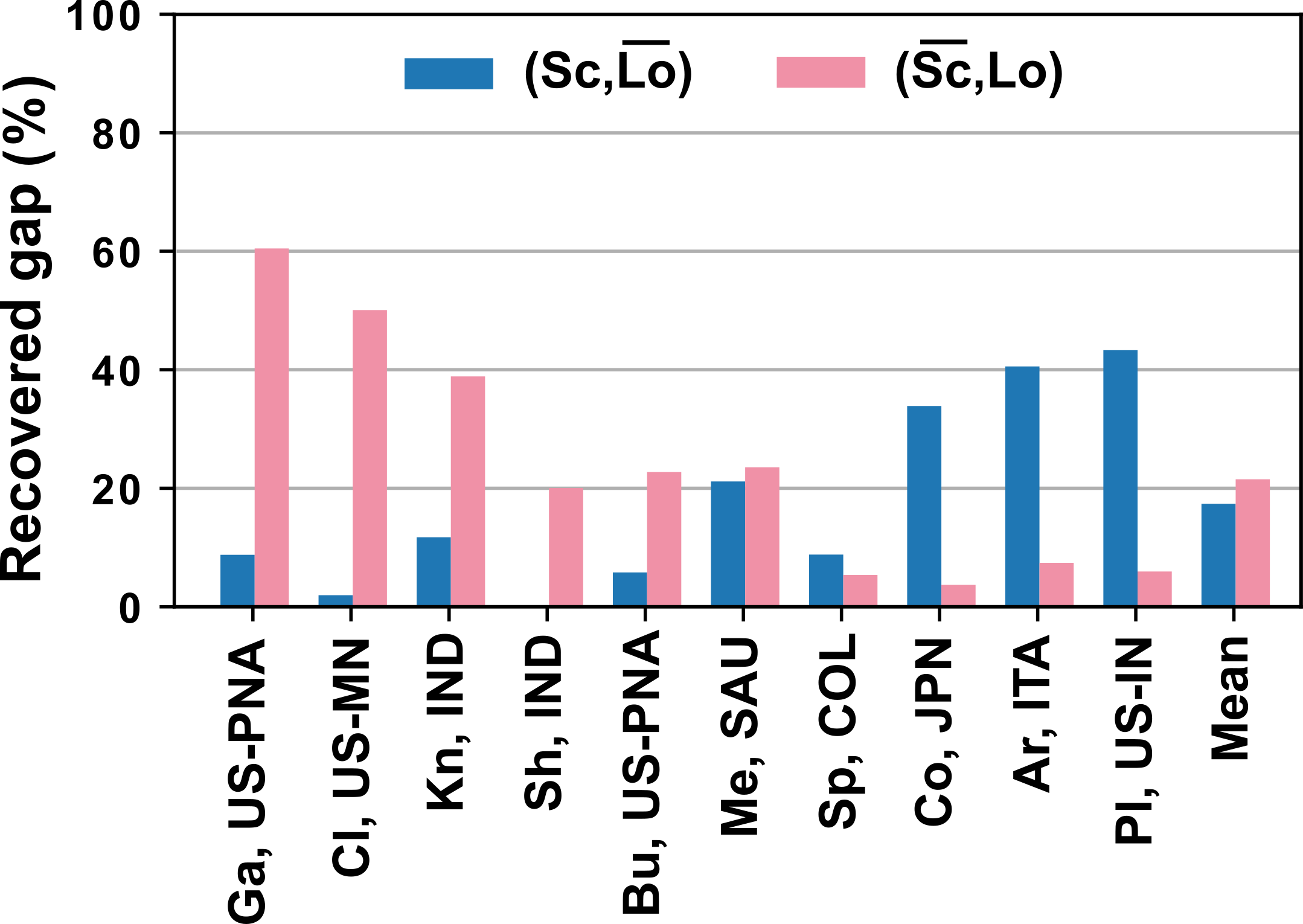}
\caption{\% of drop recovered when adding examples from either scenario {\textbf{(Sc, $\overline{\text{Lo}}$)}} or location {\textbf{($\overline{\text{Sc}}$, Lo)}}.}\label{fig:relative_drops}
\end{subfigure}
\vspace*{-8pt}
\caption{Analysis of scenario and location shifts on \dataset.}
\vspace*{-12pt}
\label{}
\end{figure}

%\begin{figure}
%\centering
%\subfloat[Top-1 accuracy (\%) on the ERM baseline, with no samples from the test scenario or location {\textbf{($\overline{\text{Sc}}$, $\overline{\text{Lo}}$)}}, either test scenario or location {\textbf{(Sc, $\overline{\text{Lo}}$)}}$\cup${\textbf{($\overline{\text{Sc}}$, Lo)}}, or both {\textbf{(Sc Lo)}}.]{\includegraphics[width=0.49\columnwidth]{imgs/pdfs/3a_iccv.png}}
%\subfloat[The percentage of the total drop recovered when adding examples from either the scenario {\textbf{(Sc, $\overline{\text{Lo}}$)}} or the location {\textbf{($\overline{\text{Sc}}$, Lo)}}.]{\includegraphics[width=0.49\columnwidth]{imgs/pdfs/3b_iccv.png} }
%\caption{Analysis of scenario and location shifts on \dataset.}
% \label{fig:shifts}
%\end{figure}

\noindent \textbf{\dataset\ Domain Shift Analysis.}\label{sec:da}
We analyse the impact of scenario and location shifts on the 10 test splits in \dataset\ by varying whether samples from the test scenario and/or location appear during training.

% \tobya{Full experimental details are provided later in Section \ref{sec:experiments}, but to understand the domain shifts in \dataset\ here, it suffices to say that standard cross-entropy training is used.
% }
%\tobyb{
For all experiments we use Empirical Risk Minimization (ERM) (\ie standard cross entropy training) - see Section \ref{sec:experiments} for full experimental details.
We present this early analysis so as to understand the domain shift in \dataset\ .
%}
We take the default setting (1) where no examples from the test scenario or the location appear during training. 
% This is the standard setup we evaluate in our experiments.
We denote this as {\textbf{($\overline{\text{Sc}}$, $\overline{\text{Lo}}$)}}, where overline indicates samples are excluded from the training split. We compare this against cases where (2) 
the training split also includes samples showcasing either the test scenario or the test location but not both, i.e. {\textbf{(Sc, $\overline{\text{Lo}}$)}}$\cup${\textbf{($\overline{\text{Sc}}$, Lo)}}, and (3) samples from the test scenario in the test location are included, i.e. {\textbf{(Sc, Lo)}}. In Figure \ref{fig:whole_drops},  performance improves from (1) $\rightarrow$ (2) with a bigger improvement (2) $\rightarrow$ (3). 
% This demonstrates that the unseen scenario and unseen location combination is particularly challenging.
This demonstrates that generalisation is particularly challenging when the combined test scenario and location do not appear during training.

Next, we analyse how much the scenario and location shifts individually contribute to this drop in performance. 
%To analyse the effect of scenario, we take the default ERM, and add examples from the same location of the test samples. Vice versa for location. 
We show the fraction of the drop recovered against (3)
when introducing training samples from either the test scenario {\textbf{(Sc, $\overline{\text{Lo}}$)}} or the test location {\textbf{($\overline{\text{Sc}}$, Lo)}}.
%We normalise the drop per shift to clarify these comparisons across our test splits.
%against the case where examples from the exact combination of scenario and location of the test are included in training. 
Fig.~\ref{fig:relative_drops} shows the impact of scenario and location varies widely for each test split. For example, on (\textbf{Sh, {IND}}), training with the test scenario \textit{shopping} does not help, whereas the location \textit{India} does. Conversely, on (\textbf{Ar, ITA}), training with \textit{arts and crafts} recovers 40\% of the drop, whereas the location does not help.
This showcases that both shifts are interesting and that our 10 test splits offer the diversity to study both.
% On average scenarios and locations on their own allow a recovery of 0.2 of the drop, showing that they are both important shifts.

\section{Method}

%Option 1
% Because of the complex nature of \dataset\, scenario and location shifts cannot be handled independently (recall that scenario clusters exist within location clusters in the dataset, and viceversa, as illustrated in Figure reffig:umaps). This means it is unclear how to apply methods designed to operate between labelled domains in the training set. Furthermore, the nature of those shifts differ significantly. While we are more likely to disregard background information (mostly dependent on the location), scenario-related information provides numerous clues about the action being performed [a citation could be EGO-TOPO?]. Thus, whether each (scenario, location) pair is considered as its own domain, or if scenario and locations are considered as two separate shifts, existing methods minimising distances between each domain feature distributions might ignore the complex relation between the two (\eg MMD~\cite{li2018domain}).

We propose Cross-Instance Reconstruction (\method) to represent an action as a weighted combination of actions from other scenarios and locations. 
We first formulate the input to our method in Section~\ref{sec:4.1}, then focus on our proposed \method\ in Section~\ref{sec:4.2}. We detail training in Section~\ref{sec:objectives} and inference in Section~\ref{sec:inference}.
%This section introduces \method, our method for domain generation. An overview is shown in Figure \ref{fig:model}. Given video clip-text pairs, \method\ first reconstructs video clip representations using video clips from multiple domains. This is followed by three learning objectives: video clip-text pairing (Section \ref{sec:v-t_ass}), video clip-action association (Section \ref{sec:v-a_ass}), and video clip classification. This leans a model for video clip classification in seen and unseen domains, which does not require text information at inference.
\begin{figure}[t!]
\centering
\includegraphics[width=1\columnwidth]{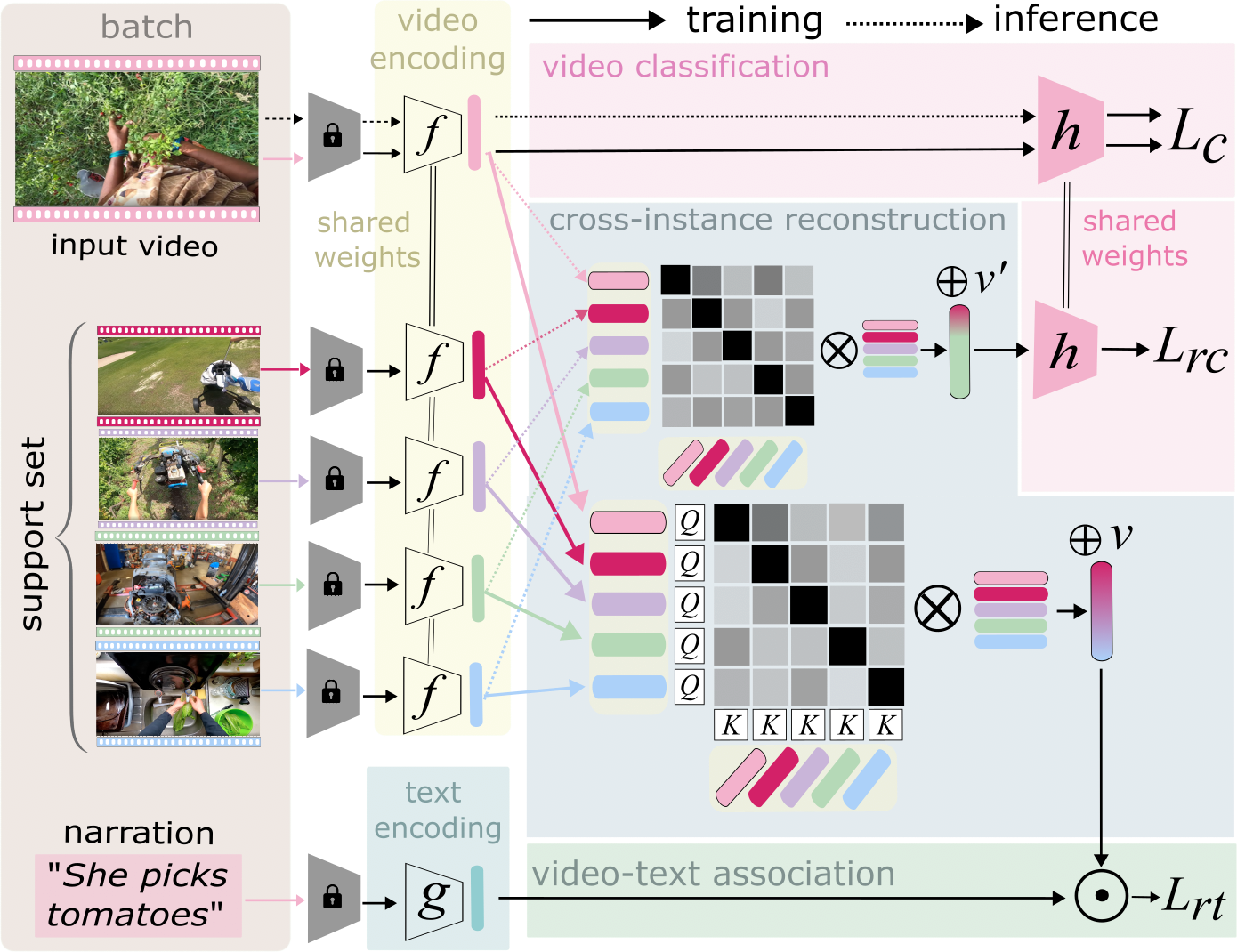}
\caption{\textbf{\method}. One clip and corresponding narration are shown along with the support set of other clips in the batch. Video $f(v)$ and text $g(t)$ embeddings are extracted using trained encoders on top of a frozen model. Cross entropy $\mathcal{L}_c$, and two CIR objectives $\mathcal{L}_{rt}$ and $\mathcal{L}_{rc}$ are minimized. For $\mathcal{L}_{rt}$, query $Q$ and key $K$ projections are learnt for clips in the batch, followed by self-masking.
Weights are multiplied by $f(v)$, and the reconstructed
$\oplus v$ is paired with the corresponding narration. For $\mathcal{L}_{rc}$, $\oplus v'$ is classified using the classifier $h$. At inference, only the video classifier $h$ is used. 
%Given video $f(v)$ and text $g(t)$ embeddings as input, cross entropy $\mathcal{L}_c$, and two CIR objectives $\mathcal{L}_{rt}$ and $\mathcal{L}_{rc}$ are minimized.
%The front pane shows $\mathcal{L}_{rt}$ where query $Q$ and $K$ projections are learnt between clips in the batch, followed by self-masking.
%Weights are multiplied by $f(v)$, and the reconstructed
%$\oplus v$ is paired with the corresponding narration. 
%The back pane shows $\mathcal{L}_{rc}$ where reconstructed $\oplus v'$ is classified using the classifier $h'$. At inference, only $h$ is used. 
}\vspace*{-12pt}\label{fig:reconstruction_a}
\end{figure}

\subsection{Proposed Setting}\label{sec:4.1}
\label{sec:problem}

\vspace*{-5pt}

Each training sample is a video clip $v$ with a free-form text narration $t$ and an action class label $y$: $(v,t,y)$. 
During testing, we only require an input video clip, to predict the action label. We use $\hat{y}$ to refer to the predicted label.

We consider a composite function to classify actions: 

\vspace*{-10pt}
\begin{equation}
\hat{y} = h \circ f (v) 
\label{eq:com}
\end{equation}
where $f$ is an encoder which learns a video representation suitable for domain generalisation, while $h$ specialises in learning an action classifier from that representation.

In addition to the cross-entropy loss $\mathcal{L}_c$ on $h$, we train the domain generalisable representation $f$ using two losses; one cross-modal and another classification loss. %We describe these next.

\begin{figure}[t!]
\includegraphics[width=\columnwidth]{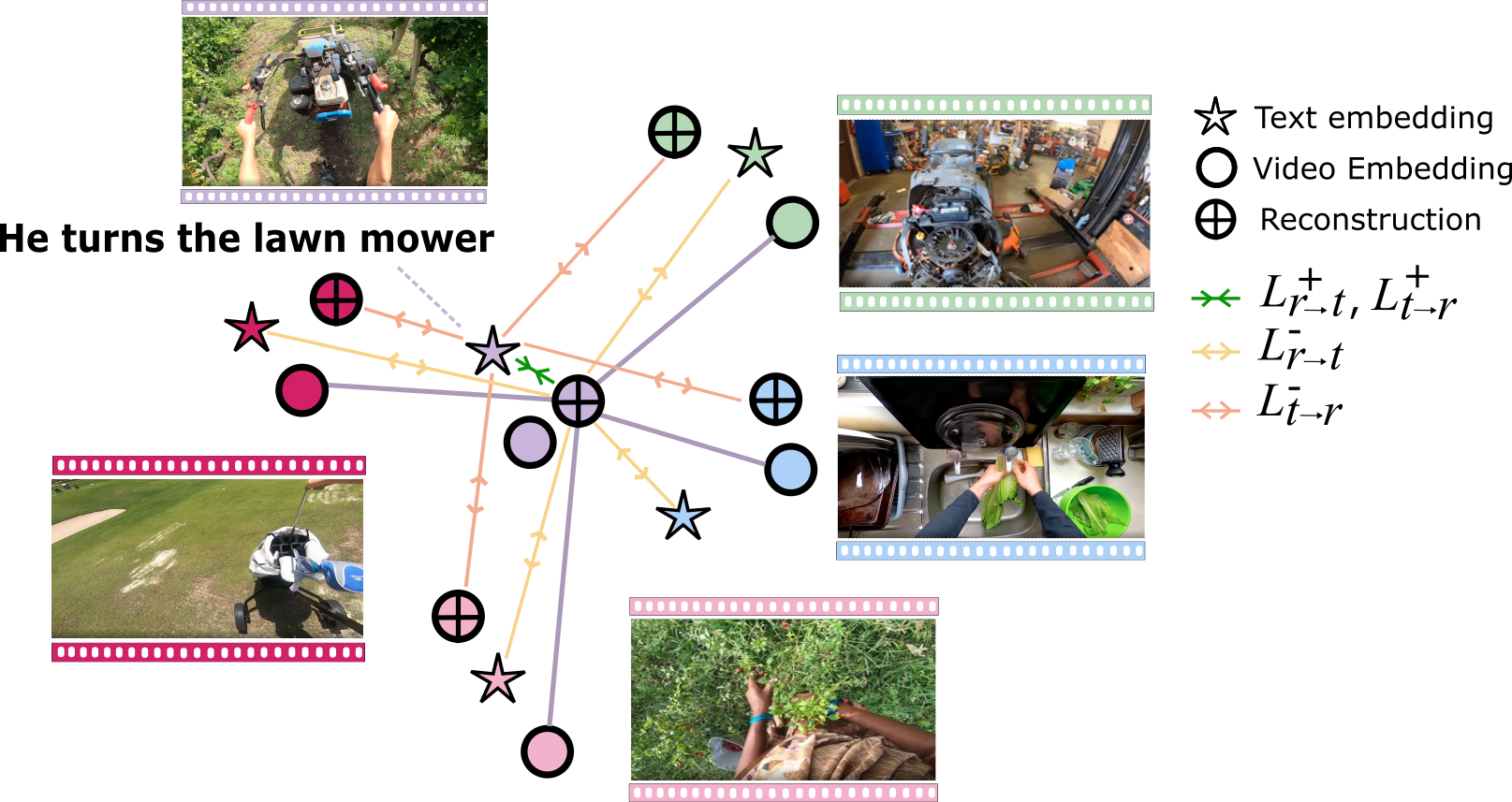}
\caption{\textbf{Video-text association.} The reconstructed clip $\oplus v_i'$ (\textcolor{Orchid}{violet}) is paired with its text representation. The reconstruction-text loss $\mathcal{L}_{r \rightarrow t}$ has the reconstruction $\oplus v_i'$ as positive and other text narrations as negatives, and the text-to-reconstruction loss $\mathcal{L}_{t \rightarrow r}$ has other reconstructions $\oplus v_j'$ as negative.}
\label{fig:reconstruction_b}
\vspace{-12pt}
\end{figure}
\subsection{Cross-Instance Reconstruction (\method)} \label{sec:4.2}

Our main premise in cross-instance reconstruction (CIR) is to encourage cross-domain representations of actions, where domains are scenarios and locations.
In doing so, these representations can be domain generalisable, as it reconstructs the same action from samples of {other} domains.

We learn-to-reconstruct any video clip from \textit{other} video clips in the randomly sampled batch, which we call the support set~$S$. %Batches are selected randomly - they contain a random set of samples from the training set, from any set of domains. 
%This cross-instance reconstruction~(CIR) is an approximate representation of the video clip itself, but reconstructed using representations from potentially other scenarios and locations.
% without domain knowledge.
% To achieve this, we use a cross-modal objective; \ie the CIR should match the caption of the original video. 
%To avoid trivial reconstructions, a video does not appear in its own support set, implemented with self-masking. 
We {\emph{jointly}} reconstruct all video clips in the batch, at the feature level. 
Each video clip appears in the support set of every other video clip in the batch. 
Before outlining the training objectives, we first describe the reconstruction process.
%Specifically,  each video is reconstructed as a weighted average of the representations of the remaining videos in the batch, excluding the query video we reconstruct, which we call the support set $S$.

We learn two projection heads, which we term the query and key heads, $Q$ and $K$, in line with standard works~\cite{vaswani2017attention}, along with a layer norm $L$. %~\cite{ba2016layer}.
%$S=\{v_j : j \neq i\}$. %We define query and key heads $Q$ and $K$  :  $\mathbb{R}^D \mapsto \mathbb{R}^{d_k}$, along with a layer norm $L$.  
We calculate the correlation between each pair of video clips, $v_i$ and $v_j$, in the training batch as:
\begin{equation} \label{eq:cij}
   c_{ij} = L(Q(f(v_i))) \cdot L(K(f(v_j)))
\end{equation}
The resulting weights $c_{ij}$ are softmaxed and self-masked  to avoid trivial reconstructions from the sample itself. 
The reconstructed representation $\oplus v_i$ is a weighted combination of all embeddings in its support set, using the weights $c_{ij}$:
%Then we weight the contribution of video clips in the support set to the reconstructed embedding $\oplus v_i$ for each clip $v_i$: 
\begin{equation}\label{eq:reconstruction}
%   \forall i: \quad \oplus{v}_i = \sum_{j \in S} \frac{\exp({c_{ij}})}{\sum_{k \in S}\exp({c_{ik}}) }\cdot  f(v_j) .
  \forall i: \quad \oplus{v}_i = \sum_{j \in S} \frac{\exp({c_{ij}}) f(v_j)}{\sum_{k \in S}\exp({c_{ik}}) }
\end{equation}
We directly weight $f(v)$ -- this is analogous to using the identity matrix for the value head in standard attention.
%encoder $f(v)$ as the value head.
%where $\cdot$ is the cross product between the two vectors. 
%Figure~\ref{fig:reconstruction_a} shows \method\ for a sample batch.
%\tobyb{
%\method\ is clearly distinct from mixup, as it (1) combines multiple videos, (2) reconstruction contributions are based on visual similarity, rather than random, (3) videos cannot contribute to their own reconstruction, and (4) the original video label is maintained.
%}

\subsection{Training CIR} \label{sec:objectives}

Fig.~\ref{fig:reconstruction_a} gives an overview of \method\ which we detail next.
We intend for reconstructions to learn to generalise, and backpropogate this ability to the video encoder $f$ (Eq.~\ref{eq:com}). 
We propose two reconstructions, each guided by a different objective. % two ways to utilise these reconstructions.
% The first obtains cross-domain instance-level reconstructions of video clips.
The video-text association reconstruction ($\oplus v$ in Fig.~\ref{fig:reconstruction_a}) uses text narrations so these cross-instance reconstructions are associated with the video clip's semantic description.
The classification reconstruction ($\oplus v'$ in  Fig.~\ref{fig:reconstruction_a}) is trained to recognise the clip's action class.
%Old: the first intends to reconstruct the specific instance of the action in the video.
%The second aims at finding reconstructions of the associated action classes across domains.
%The second aims at cross-domain action level reconstructions.
%Old: the second ensures the reconstructions correctly discriminate between action classes. 

%In the \textbf{first objective}, we reconstruct the video
% %using associated narrations \eg ``He turns the lawn mower'' (Fig.~\ref{fig:reconstruction_b}), as those provide a description of the instance which is not domain-specific and thus ensure the reconstruction to capture information which is invariant across domains. 
% we use associated narrations \eg ``He turns the lawn mower'' (Fig.~\ref{fig:reconstruction_b}) 
% . 

% old: In the \textbf{first objective}, we wish to reconstruct the video but avoid comparing the reconstruction to the video itself,as the video representation combines the action with the domain (i.e. the scenario and the location).
%Instead, use the narration associated with the video \eg ``turns the lawn mower'' (Fig.~\ref{fig:reconstruction_b}).
For the \textbf{video-text association reconstruction} $\oplus v_i$, 
we use contrastive learning to push $\oplus{v}_i$ towards the embedding of the text narration associated with the video, \eg~``He turns the lawn mower''. 
Given a batch of video-text pairs with corresponding reconstructions ${\mathcal{B}=\{(v_{i}, \oplus v_i, t_{i})\}^{B}_{i=1}}$, the resulting objective is formulated using Noise Contrastive Estimation~\cite{oord2018representation} over both reconstruction-text and text-reconstruction pairs. Specifically, the reconstruction-text loss considers the reconstruction $\oplus v_i$ as the anchor and the negatives as other text narrations in the batch, such that:
\begin{equation}
   \mathcal{L}_{r \rightarrow t}(\oplus{v}_i, g(t_i)) = - \frac{1}{B} \sum_i^B \log \frac{\exp \left( s(\oplus{v}_i, g(t_i))/\tau \right) }{\sum\limits^B_{j} \exp \left(s(\oplus v_i, g(t_j))/\tau \right)}
\end{equation}
% \begin{equation}
%    \mathcal{L}_{rt} = \frac{-2}{B} \sum_i^B \log \frac{\exp(s(\oplus{v}_i, g(t_i))/\tau)}{\sum\limits^B_{j} \exp((s(\oplus{v}_j, g(t_i))+s(\oplus v_i, g(t_j)))/\tau)} 
% \end{equation}
%\begin{equation}
%   \mathcal{L}_{rt} = \frac{-1}{B} \sum_i^B \log \frac{\exp \left( s(\oplus{v}_i, g(t_i))/\tau \right) }{\sum\limits^B_{j} \exp \left( s(\oplus{v}_j, g(t_i)) / \tau +s(\oplus v_i, g(t_j))/\tau \right)}
%\end{equation}
where $s(\cdot,\cdot)$ is the cosine similarity, $g$ is the text encoder, $g(t_i)$ is the encoded text narration, and $\tau$ is a learnable temperature. 
The analogous loss $\mathcal{L}_{t \rightarrow r}$ considers $g(t_i)$ as anchor and other reconstructions as negatives.  We showcase these in Fig.~\ref{fig:reconstruction_b}. Both are combined to form our reconstruction-text association loss $\mathcal{L}_{rt} = \mathcal{L}_{r \rightarrow t} + \mathcal{L}_{t \rightarrow r}$.

Note that we avoid pairing this reconstruction with the video embedding $f(v_i)$, instead of the text narration $g(t_i)$, as it may convey domain-knowledge (\ie scenario and location), which might bias the  reconstruction to videos from the same scenario or location.
%which can only be found in videos from the same scenario or location, thus biasing the reconstruction.
Instead, the associated narration offers an instance-level description of the action, which guides the reconstruction.

Our \textbf{classification reconstruction} $\oplus v'_i$ forms the input to the classifier $h$, so as to recognise the action class such that $\hat{y}' = h (\oplus v')$.
We train with cross-entropy loss, which we term~$\mathcal{L}_{rc}$ to imply classifying reconstructions.
%We also use a separate classifier ($h'$) as these reconstructions may have different distributions to the original video clips.
%Feeding these to the classifier we use at inference ($h$) can harm performance.
%Additionally, we use a separate classifier ($h’$) for reconstructions, as the classifier used at inference ($h$) will not see reconstructed video clips, and reconstructed video clips may have a different distribution to original video clips.
%Additionally, we use a separate classifier ($h'$ instead of $h$), as these reconstructions can be out-of-distribution (OOD) of real video clips. In other words, the distribution of the resulting reconstructed videos might be different from the ones of training domains. 
We share the weights between the classifier for videos and for reconstructions.
Additionally, for this reconstruction, we compute weights with cross-product attention: ${c'_{ij} = f(v_i) \cdot f(v_j)}$, \ie by replacing $c$ with $c'$ in Eq.~\ref{eq:reconstruction}. 
We thus do not learn additional query and key projections.
%This is important to ensure $f$ can directly be used for class predictions.
We ablate these decisions in \cref{sec:ablations}. % showcasing its superior performance.
% cross-instance reconstruction is the result of soft-attention. \tobyn{A reason why?}

%Importantly, this loss is used to train the generalisable representation $f$.

%While our first objective achieves generality across domains -- in our case scenarios and locations, the reconstruction might deviate too much from our action recognition goal.

%with the symmetric text-CIR objective $\mathcal{L}_{t2v}$ defined in the same way. 
%\begin{equation}
%   \mathcal{L}_{t2v} = -\frac{1}{B} \sum^B_I \log \frac{\exp(s(\bar{v}_i, t_j))}{\sum^B_i \exp(s(\bar{v}_j, t_i))} 
%\end{equation}
% The final video-text pairing loss is then formulated as: 
% \begin{equation}
%    \mathcal{L}_{vt} = \mathcal{L}_{v2t} +\mathcal{L}_{t2v} 
% \end{equation}
%where  is defined as the cosine similarity between the reconstructed video and the original text embedding.
% \begin{equation}
%   {s(\bar{v}_i, t_j)} = \frac{\bar{v}_j^\top t_i}{\lVert \bar{v}_j \rVert \lVert t_i \rVert}   .
% \end{equation}

%The resulting learning objective is thus the following: 

We combine our two losses with the cross-entropy video classification loss $\mathcal{L}_c$ (see Section \ref{sec:4.1}). Our overall training objective is: 
\begin{equation}
    \mathcal{L} = \mathcal{L}_{c} + \lambda_1 \mathcal{L}_{rt} + \lambda_2 \mathcal{L}_{rc}   .
\label{eq:loss}
\end{equation} 
where $\lambda_1$ and $\lambda_2$ weight the two reconstruction losses.

\subsection{Inference}\label{sec:inference}
Once training concludes, $f$ is capable of extracting domain generalisable representations that maintain action class knowledge without domain bias.
Accordingly, at test time, only video clips $v_i$ from the test split are processed by the encoder $f$ and classifier $h$. 
We do not require any narration during inference, and there is no reconstruction -- \ie each clip is classified independently.

\section{Experiments}\label{sec:experiments}

%\noindent \textbf{Dataset and metrics. } 

% If you use beamer only pass "xcolor=table" option, \ie \documentclass[xcolor=table]{beamer}
% If you use beamer only pass "xcolor=table" option, \ie \documentclass[xcolor=table]{beamer}

% \usepackage{multirow}

\setlength{\tabcolsep}{3.5pt}
\begin{table*}[]
\centering
\footnotesize
\begin{NiceTabular}{lllllllccccccccccc} 
\toprule
\multicolumn{1}{l}{}                   & \multicolumn{6}{c}{DG Strategies}                                        & \Gardening\US            & \Cleaning\US        & \Knitting\India                                                        & \Shopping\India                                                                                                                              & \Building\US                                                                                                                                                                            & \Mechanic\Saudi                                               & \Sport\Colombia                                                                         &\Cooking\Japan                 & \Sewing\Italy                                                                                               & \Playing\US                                                                                                &                        \\ \cline{2-7}
\multicolumn{1}{l}{}                   &                     &                     &                     &  &                                                                                &                                                                            &                                                                               &                                                                                 &                                                                            & \multicolumn{1}{c}{}                                                                            & \multicolumn{1}{c}{}                                                                            & \multicolumn{1}{c}{}                                                                            & \multicolumn{1}{c}{}  & \multicolumn{1}{c}{}                                                                              &                        \\
\multicolumn{1}{l}{\multirow{-3}{*}{}} & \multirow{-2}{*}{D} & \multirow{-2}{*}{A} & \multirow{-2}{*}{M} &  \multirow{-2}{*}{P} & \multirow{-2}{*}{R}  & \multirow{-2}{*}{T} & \multirow{-2}{*}{\textbf{\begin{tabular}[c]{@{}c@{}}Ga\\ US-PNA\end{tabular}}} & \multirow{-2}{*}{\textbf{\begin{tabular}[c]{@{}c@{}}Cl\\ US-MN\end{tabular}}} & \multirow{-2}{*}{\textbf{\begin{tabular}[c]{@{}c@{}}Kn\\ IND\end{tabular}}}& \multirow{-2}{*}{\textbf{\begin{tabular}[c]{@{}c@{}}Sh\\ IND\end{tabular}}}  & \multirow{-2}{*}{\textbf{\begin{tabular}[c]{@{}c@{}}Bu\\  US-PNA\end{tabular}}}& \multirow{-2}{*}{\textbf{\begin{tabular}[c]{@{}c@{}}Me\\ SAU\end{tabular}}} & \multicolumn{1}{c}{\multirow{-2}{*}{\textbf{\begin{tabular}[c]{@{}c@{}}Sp\\ COL\end{tabular}}}} & \multicolumn{1}{c}{\multirow{-2}{*}{\textbf{\begin{tabular}[c]{@{}c@{}}Co\\ JPN\end{tabular}}}} & \multicolumn{1}{c}{\multirow{-2}{*}{\textbf{\begin{tabular}[c]{@{}c@{}}Ar\\ ITA\end{tabular}}}} & \multicolumn{1}{c}{\multirow{-2}{*}{\textbf{\begin{tabular}[c]{@{}c@{}}Pl\\ US-IN\end{tabular}}}} & \multirow{-3}{*}{\textbf{Mean}} \\ \hline
Random                                 &                  &               &                  &                       &                &     & \textcolor{white}{0}{\color[HTML]{8d8d8d}}8.00   &      10.64                                &     \textcolor{white}{0}{\color[HTML]{8d8d8d}}9.13                                                    &  14.36                                                                                                                                                                                                                              & \textcolor{white}{0}{\color[HTML]{8d8d8d}}9.55                                                                         & 13.04                                                                         & \textcolor{white}{0}{\color[HTML]{8d8d8d}}8.35                                                                     & 10.13                                                                  & \textcolor{white}{0}{\color[HTML]{8d8d8d}}9.86                                                                          & 15.68                                                                          & {\cellcolor[HTML]{EFEFEF}}10.84           \\
ERM                                                      &                       &          &             &                &        &                       & 20.75      & 22.35          & 18.69                                                                                        & 22.14                                                                                                                                                                                         & 20.73                                                                                                                                                                                                & 23.51                                                                                               & 18.97                                                                                               & 24.81                                                                                               & 22.75                                                                                               & 23.29                                                                                                 & {\cellcolor[HTML]{EFEFEF}}21.80                                                  \\ \hline
CORAL* \cite{sun2016deep}          & \cmark &             &           &                       &       &                & 22.14       & 22.55        & 19.07                                                                                          & 24.01                                                                                                                                                                                           & 22.18                                                                                                                                                                                             & 24.31                                                                                               & 19.16                                                                                               & 25.36                                                                                               & 23.89                                                                                               & {\ul25.96}                                                                                                 & {\cellcolor[HTML]{EFEFEF}22.86}                                                   \\
DANN* \cite{ganin2016domain}       & \cmark & \cmark &           &             &                 &      & {\ul 22.42}        & {\ul 23.85}       & 19.27                                                                                       & 22.89                                                                                                                                                                                       & 22.23                                                                                                                                                                                        & 23.70                                                                                               & 18.64                                                                                               & 25.86                                                                                               & 23.86                                                                                               & 23.28                                                                                                 & {\cellcolor[HTML]{EFEFEF}}22.60                                                   \\
MMD* \cite{li2018domain}           & \cmark &            &            &                       &      &                 & {\ul 22.42}       & 23.60       & 19.66                                                                                  & {\ul 24.46}                                                                                                                                                                                     & 22.08                                                                                                                                                                                                & 24.64                                                                                               & {\ul 19.59}                                                                                         & 25.87                                                                                               & 23.84                                                                                               & 24.78                                                                                                 & {\cellcolor[HTML]{EFEFEF}}{\ul 23.09}                                             \\
Mixup \cite{wang2020heterogeneous} &                   &     & \cmark                      &  &       &                & 21.97      & 22.21       & {19.90}                                                                                           & 23.81                                                                                                                                                                                            & 21.45                                                                                                                                                                                       & 24.35                                                                                               & 19.01                                                                                               & {\ul 25.90}                                                                                         & 23.85                                                                                               & 24.41                                                                                                 & {\cellcolor[HTML]{EFEFEF}}22.69                                                   \\
BoDA*\cite{yang2022multi}         & \cmark &           &             &                       &       &                & 22.17   & 22.78         & 19.62                                                                                            & 22.94                                                                                                                                                                                            & 21.46                                                                                                                                                                                            & 23.97                                                                                               & 19.18                                                                                               & 25.68                                                                                               & 23.92                                                                                               & 24.90                                                                                                 & {\cellcolor[HTML]{EFEFEF}}22.66                                                   \\
DoPrompt* \cite{zheng2022prompt}   &  &         &               &                    \cmark   &      &                 & 21.92     & 22.77          & {\ul 20.40}                                                                                        & 23.67                                                                                                                                                                                           & {\ul 22.75}                                                                                                                                                                                & {\ul 24.67}                                                                                         & 18.24                                                                                               & 25.04                                                                                               & {\ul 24.74}                                                                                         &  25.24                                                                                           & {\cellcolor[HTML]{EFEFEF}}22.94                                                   \\ \hline
\RowStyle{\color{gray}}
%CIR   (w/o text)                                             &  &                       &   &           &      \cmark    &   & 24.16	&24.83&	21.00&26.95				&24.80	 	&	26.77	&	18.58		&25.06		&25.38		&29.50		 & {\cellcolor[HTML]{EFEFEF}}24.70\\

CIR   (w/o text)                                             &  &                       &   &           &      \cmark    &   & 23.39	&24.52&	21.02 &26.62				&24.64	 	&	27.00	&	19.66		&25.42		&25.71		& 30.17		 & {\cellcolor[HTML]{EFEFEF}}24.81\\

%CIR                                                      &  &                       &          &          &   \cmark  & \cmark & \textbf{23.91}     & \textbf{25.10}    & \textbf{20.40}                                                                                    & \textbf{26.18}                                                                                                                                                                          & \textbf{25.01}                                                                                                                                                                             & \textbf{26.79}                                                                                      & \textbf{20.43}                                                                                      & \textbf{26.74}                                                                                      & \textbf{25.70}                                                                                      & \textbf{30.67}                                                                                        & {\cellcolor[HTML]{EFEFEF}}\textbf{25.09}    \\ 
{CIR }                                                     &  &                       &          &          &   \cmark  & \cmark &\textbf{{24.10}}     & \textbf{{25.51}}    & {{\textbf{20.46}}}                                                                                   & \textbf{{27.78}}                                                                                                                                                                          & \textbf{{24.93}}                                                                                                                                                                             & \textbf{{26.83}}                                                                                      & \textbf{{19.75}}                                                                                      & \textbf{{26.34}}                                                                                      & \textbf{{25.67}}                                                                                      & \textbf{{30.94}}                                                                                        & {\cellcolor[HTML]{EFEFEF}}\textbf{25.23}    \\ 
\bottomrule                                     
\end{NiceTabular}
\vspace*{-6pt}
\caption{Top-1 accuracy on \dataset. Best results in \textbf{bold}, second best \underline{underlined} (omitting CIR w/o video-text association loss, which is greyed out but given for direct comparison showcasing strong performance w/o narrations).
$*$: Domain labels required during training. \textit{D}: distribution matching, \textit{A}: adversarial learning, \textit{M}: label-wise mix-up, \textit{P}: domain-prompts, \textit{R}: reconstruction \textit{T}: video-text association.
%learning textual representations. 
 }
 \vspace*{-12pt}
\label{tab:baselines}
\end{table*}
% \usepackage{multirow}
% \usepackage[normalem]{ulem}

%\tobyn{Chiara: add a little intro about what we want to show/test with experiments.}
We test the ability of \method\ to generalise over scenarios and locations by comparing it against baseline and state-of-the-art domain generalisation methods adapted for our setting. We then show ablations on its different components, and visualise its impact with qualitative examples. 

%In this Section we first describe the dataset and metrics used for evaluation, as well as the baseline methods for comparison and implementation details. We then show results of our method when compared to other baselines, and ablations on the proposed method. 

\noindent \textbf{Dataset and metrics.} 
We use the \dataset\ dataset introduced in \cref{sec:dataset} for all experiments. 
%For performance evaluation, we conduct experiments on the 10 distinct test splits that have been outlined in Section \ref{sec:dataset}. 
We report top-1 accuracy for each test split, as well as mean accuracy.
% For the ablations, we select the largest 5 test splits, which are \textbf{Cl,US-MIN}, \textbf{Bu,US-PNA}, \textbf{Co,JPN}, \textbf{Ar,ITA} and \textbf{Pl,US-IN}.

%For each method, we train once on the training set, then report Top-1 accuracy on each of the seven distinct test splits, which as a reminder test \slsa{seen scenario/seen location (SS/SL)}, \dlsa{seen scenario/unseen location (SS/UL)}, \slda{unseen scenario/seen location (US/SL)} and \dlda{unseen scenario/unseen location (US/UL)}. 
%Recall that the training dataset contains videos from 7 scenarios and 11 locations, with 73 distinct combinations. 

\noindent \textbf{Baselines.} 
We first compare our method with the Empirical Risk Minimisation (ERM) baseline \cite{vapnik1999overview}, as is standard practice in DG works~\cite{carlucci2019domain,gulrajani2020search}. This is cross-entropy~ ($\mathcal{L}_c$)
without a generalisation objective. 
We then compare against 6 methods for DG, all trained jointly with $\mathcal{L}_c$. 
%\method\ does not use domain (\ie scenario and location) labels to train, instead relying on narrations.
%(just narrations and videos)
%However, 

Most DG methods do require domain labels during training. We thus provide these labels when required and mark these methods with (*).  %In these cases, a baseline will train two generalisation objectives, one for scenarios and one for locations. %In supplementary, we provide separate results for when only one generalisation objective is used. 
At test time, all methods only use video clip input, and are not aware of any domain knowledge. Our baselines, ordered by publication year, are:

\vspace*{-8pt}
\begin{itemize}[leftmargin=3mm,itemsep=-2ex,partopsep=1ex,parsep=2ex]\item CORAL*~\cite{sun2016deep}: two mean and covariance distances are minimised. These are the distances between means and covariances of video representations from different scenarios, and the distances between means and covariances from different locations.
\item DANN*~\cite{ganin2016domain}: 2-fully connected layers  form an adversarial network to predict the location. A separate adversarial network predicts the scenario. 
\item MMD*~\cite{li2018domain}: same as CORAL w/ MMD distances~\cite{gretton2012kernel}.
\item Mixup~\cite{wang2020heterogeneous}: training data is augmented by performing linear interpolations of samples and labels. 
% Mixup has been shown to help generalisation~\cite{wang2020heterogeneous}. 
Note that Mixup is distinct from \method\, as it focuses only on pairs of videos selected randomly, rather than reconstructing from all videos in the batch based on visual similarity. Additionally, Mixup changes the output label, while in \method\ the video class label is maintained.
%\item Mixup w/text: We add our $L_{rt}$ loss acting directly on narrations and video input (\ie no \method) to mixup, which provides a baseline with access to exactly the same level of supervision (narrations and no domain labels). \tobyn{Say if this should appear in an ablation instead}.
\item BoDA*~\cite{yang2022multi}: minimises distances between domains, similar to MMD, weighted by both domain size and class size, in an effort to handle imbalance.
\item DoPrompt*~\cite{zheng2022prompt}: learns one domain prompt for each scenario and location to be appended to visual features before classification.
\end{itemize}
\vspace*{-12pt}
We also provide random chance averaged over 10 trials. 
%For an upper bound, we take ERM, but train with examples from the domains for the test split. For example, Sport in Colombia will contain examples from sport, examples from Colombia, but not examples from both. We donote this as ERM + domain examples.

\noindent \textbf{Implementation details.} We use SlowFast features~\cite{feichtenhofer2019slowfast}, pre-trained on Kinetics~\cite{carreira2017quo}, provided with the videos of Ego4D~\cite{grauman2022ego4d}. 
We represent the action by concatenating three features, forming a 6912-D vector, as in~\cite{Zhou2018}, taken from the action's onset as associated with the narration, halfway to the next action, and before the start of the next action.
%Because actions are different lengths, we concatenate the first, middle and last available features for the action as its feature representation.
For text features (512-D) we use the frozen text encoder of the pre-trained CLIP-ViT-B-32 model~\cite{reimers-2019-sentence-bert} 
%to obtain a 512-D vector
% \footnote{We replace \#C  with a random phrase, \eg  ``The user", ``The~wearer".}.

$f$ is implemented as 2 fully connected layers of hidden dimension 4096 and output dimension 512, with a ReLU activation function and a Batch Normalisation layer~\cite{ioffe2015batch}. $g$~is implemented as 2 fully connected layers with 512 hidden dimension and a ReLU activation function. The dimension of query and key embeddings for reconstruction is 128. 

We use a batch size of 128 for all experiments and methods, and train for 50 epochs using the Adam optimiser~\cite{kingma2014adam}.  The learning rate is set to $2e^{-4}$ for CIR, decaying by a factor of $10$ at epochs 30 and 40. We set $\lambda_1=1$ and $\lambda_2=0.5$ (Eq.~\ref{eq:loss}). Ablation on hyperparameters is in the Supplementary. %We optimise hyperparameters for all other methods on the biggest validation split (Pl, US-IND).
%To provide a fair comparison, we present results for the best learning rate and method/cross-entropy weighting for every baseline (see supplementary for full details). 
%We fix the seed for all experiments for repeatable results.
Training takes 8 hours on one Nvidia P100 GPU.

%\slsa{seen location/seen scenario}, \dlsa{seen scenario/unseen location}, \slda{seen location/unseen scenario} and \dlda{unseen scenario/unseen location}. We also report on a validation set (10\% of the training set). Note that the Cooking in Japan split is different from the validation, as no samples from Cooking Rwanda in Tokyo are seen at training time. 

 %\begin{figure}[t!]
%    \centering
%    \includegraphics[width=0.9\columnwidth]{imgs/pd%fs/bar_plots_1.pdf}
 %   \caption{Top-1 Accuracy on test splits of comparing ERM (light) to the proposed \method\ %(dark). 
%    }
%    \vspace*{-12pt}
%    \label{fig:bar_results}
%\end{figure}

\setlength{\tabcolsep}{2.8pt}

\begin{table}[t]
%\vspace{+10pt}
\centering
\footnotesize
\begin{tabular}{lcccccc}
\toprule
              & \textbf{\begin{tabular}[c]{@{}c@{}}Cl\\ US-MN\end{tabular}} & \textbf{\begin{tabular}[c]{@{}c@{}}Bu\\ US-PNA\end{tabular}} & \textbf{\begin{tabular}[c]{@{}c@{}}Co\\ JPN\end{tabular}} & \textbf{\begin{tabular}[c]{@{}c@{}}Ar\\ ITA\end{tabular}} & \textbf{\begin{tabular}[c]{@{}c@{}}Pl\\ US-IN\end{tabular}} & \textbf{Mean}                 \\ \hline
%$\mathcal{L}_v$     

% & 29.39                                                       & 25.70          & 26.31          & 27.66          & 22.32          & 23.79                                                        & 18.52                                                                                                                           \\

%CIR (ours) & \textbf{25.10}                                              & \textbf{25.01}                                               & \textbf{26.74}                                            & \textbf{25.70}                                            & {30.67}                                              & \cellcolor[HTML]{F3F3F3}{\textbf{26.64}} \\
CIR (ours) & {25.51}                                              & \textbf{{24.93}}                                               & {26.34}                                            & \textbf{{25.67}}                                            & \textbf{{30.94}}                                              & \cellcolor[HTML]{F3F3F3}{\textbf{26.68}} \\

$\quad \quad -\mathcal{L}_{rt}$  & 24.83                                                       & 24.80                                                        & 25.06                                                    & 25.38                                                     & 29.50                                                      & \cellcolor[HTML]{F3F3F3}25.91          \\

$\quad \quad -\mathcal{L}_{rc}$   & 23.13                                                       & 23.53                                                        & 25.87                                                     & 24.95                                                     & 26.59                                                       & \cellcolor[HTML]{F3F3F3}24.81          \\

$\quad \quad -\mathcal{L}_{rt} -\mathcal{L}_{rc}$ %= ERM 
& 22.35 &	20.73 &	24.81 &	22.75 &	23.29 &	\cellcolor[HTML]{F3F3F3}22.78              \\
$\quad \oplus v$ cross-product & \textbf{25.66} &	24.84 &	25.42 &	25.41 &	30.67	   & \cellcolor[HTML]{F3F3F3}26.40       \\
%$\quad \oplus v$ {cross-product} & {24.54}                                                       & {24.86}                                                        & {25.40}                                                     & {25.35}                                                     & {30.80}                                                       & \cellcolor[HTML]{F3F3F3}26.19          \\

$\quad \oplus v'$ learnt att. &22.58	& 22.55	&25.85 &	24.53 &	25.35 &		\cellcolor[HTML]{F3F3F3}24.17
	\\
%\quad \oplus v'$ {learnt att.} & {22.58}	& {22.55}	& {25.85} &	{24.53} &	{25.35} &		\cellcolor[HTML]{F3F3F3}24.17	\\
$\quad \oplus v = \oplus v'$& 23.47 &	23.33	&25.53 &	24.06& 28.74	& \cellcolor[HTML]{F3F3F3}25.03          \\
%$\quad \oplus v = \oplus v'$& {22.59}                                                       & {22.33}                                                        & {25.55}                                                     & {23.80}                                                     & {25.54}                                                       & \cellcolor[HTML]{F3F3F3}23.96          \\

$\quad h \neq h' $ & 24.47                                                       & 23.12                                                        & \textbf{26.74}                                                     & 24.74                                                     & 27.37                                                       & \cellcolor[HTML]{F3F3F3}25.29         \\ \bottomrule

%\method\  & L & S &
%29.31                         & 25.46   & 22.17 & \underline{19.74} \\

%& 29.23 &	{26.35} &	27.05 &	27.59 &	{23.43} &	23.88 &	18.70                                              \\

%$ & 29.61                                                       & {26.15} & \textbf{27.48} & \textbf{29.13} & 23.00          & \textbf{25.15}                                               & \textbf{20.09}                                           \\  
\end{tabular}
\vspace*{-6pt}
\caption{Ablation on \method, showing the contribution of the two reconstructions and alternative design choices.}
\vspace*{-12pt}
\label{tab:ablation}
\end{table}

% \begin{figure}[t!]
%     \centering
%     \includegraphics[width=\columnwidth]{imgs/pdfs/avg_class.pdf}
%     \caption{Average Top-1 accuracy over the 7 unseen test splits after training with ERM and CIR. 
%     }
%     \label{fig:bar_results}
% \end{figure}

\subsection{Results}
Table \ref{tab:baselines} shows \method\ outperforms all previous approaches, on every test split, by up to 4.9\%, and is on average {2.1}\% better than the second best method.
Compared to the ERM baseline, CIR outperforms by {3.4}\% on average and up to {7.7}\%.
The improvement varies across splits, with the smallest improvements occurring on harder splits -- those with lower ERM baselines, \eg (\textbf{Kn}, \textbf{IND}) and (\textbf{Sp}, \textbf{COL}).

% It is important to note that methods use different forms of supervision.
CIR does not use any domain labels during training, which is a common strategy for other methods (marked by $*$ in Table \ref{tab:baselines}), but instead assumes access to textual narrations. We also report results of CIR without text (\ie without $\mathcal{L}_{rt}$) or domain labels 
showcasing strong average performance for CIR with less supervision than other methods. 
%still outperforms all other methods on average, showing the importance of reconstruction in learning representations which generalise.

The second best performing method varies per split, showcasing the complexity of the problem as well as the need for multiple test splits to properly assess domain generalisation approaches.
%Interestingly, MMD~\cite{li2018domain} which is a classic approach performs second best overall, with newer methods struggling to outperform it. 
Methods that learn domain invariant visual features by matching distributions or via domain prompts seem to struggle with the scenario shift proposed in \dataset.
%Methods that learn domain-invariant visual features by matching distributions across different domains
%or prompt learning to embed the knowledge of source domains assume that the shift primarily lies in the appearance of the data. 
%Accordingly, they struggle to bridge the domain shifts in \dataset.
Results of \method\ show that reconstruction and usage of text narrations are an effective alternative.

\subsection{Ablations}\label{sec:ablations}

We use the 5 largest test splits for all ablation results. 

\begin{table}[t]
\centering
\footnotesize
%\resizebox{\linewidth}{!}{
\begin{tabular}{llllcccccc}
\toprule
SL &  SS& OL& OS&            \textbf{\begin{tabular}[c]{@{}c@{}}Cl\\ US-MN\end{tabular}} & \textbf{\begin{tabular}[c]{@{}c@{}}Bu\\ US-PNA\end{tabular}} & \textbf{\begin{tabular}[c]{@{}c@{}}Co\\ JPN\end{tabular}} & \textbf{\begin{tabular}[c]{@{}c@{}}Ar\\ ITA\end{tabular}} & \textbf{\begin{tabular}[c]{@{}c@{}}Pl\\ US-IN\end{tabular}} & \textbf{Mean}                 \\ \hline

% \cmark    & -                                                                          & -                                                                            & -                                                                              & -                                                                                 & 29.31                                                                    & 25.76                                                               & {\ul 23.45}                                                  & 18.82                                                                  \\
% \cmark    & \cmark                                                      & \cmark                                                        & \cmark                                                          & \cmark                                                             & 28.86                                                                    & 25.23                                                               & 21.89                                                        & 19.10                                                                  \\
% \midrule
\cmark                                                      & \cmark                                                        & \xmark                                                          & \cmark                                                             & 25.01	& 24.86 &	25.73 &	\textbf{25.99} &	30.69                                                 & \cellcolor[HTML]{F3F3F3}26.46                             
\\
\cmark                                                      & \cmark                                                        & \cmark                                                          & \xmark                                                              &25.00 &	25.05	& 26.07	& 25.62 & 	30.98 & \cellcolor[HTML]{F3F3F3}26.55                               \\

\cmark                                                      & \cmark                                                        & \xmark                                                          & \xmark                                                            & 24.87 &	24.68	& 25.77 &	25.38 &	30.07 & \cellcolor[HTML]{F3F3F3}26.15                               \\
  \midrule
\xmark                                                      & \cmark                                                    & \cmark                                                          & \cmark                                                             & 24.89 &	\textbf{25.13}	& 26.05 &	25.80 &	30.47                                              & \cellcolor[HTML]{F3F3F3}26.47                              \\
    
\cmark                                                      & \xmark                                                        & \cmark                                                          & \cmark                                                     & 25.22	& 24.99	& 26.34 &	25.84 &	30.25                                                       & \cellcolor[HTML]{F3F3F3}26.53                                        \\
\xmark                                                      & \xmark                                                        & \cmark                                                          & \cmark                                                            & 25.17	& 24.97	&\textbf{26.36}	&25.61 &	30.31                                                  & \cellcolor[HTML]{F3F3F3}26.48                               \\
    
\midrule
 \cmark    & \cmark & \cmark                                                 & \cmark     & \textbf{25.51}	& 24.93 &	26.34 &	25.67	&\textbf{30.94}                      & \cellcolor[HTML]{F3F3F3}\textbf{26.68}        \\               
              \bottomrule                       
\end{tabular}%}
\vspace*{-6pt}
\caption{Effect of masking samples in the support set used for reconstruction. Columns indicate whether the query can (\cmark) or cannot (\xmark) attend to samples from the \textbf{S}ame \textbf{S}cenario/\textbf{L}ocation (\textbf{SS}, \textbf{SL}) or \textbf{O}ther \textbf{S}cenario/\textbf{L}ocation (\textbf{OS}, \textbf{OL}) based on the domains they belong to. Note that \method\ (bottom) does not use any masking.}
\vspace*{-8pt}
\label{tab:mask}
\end{table}

\begin{figure}
    \centering
    \includegraphics[width=\columnwidth]{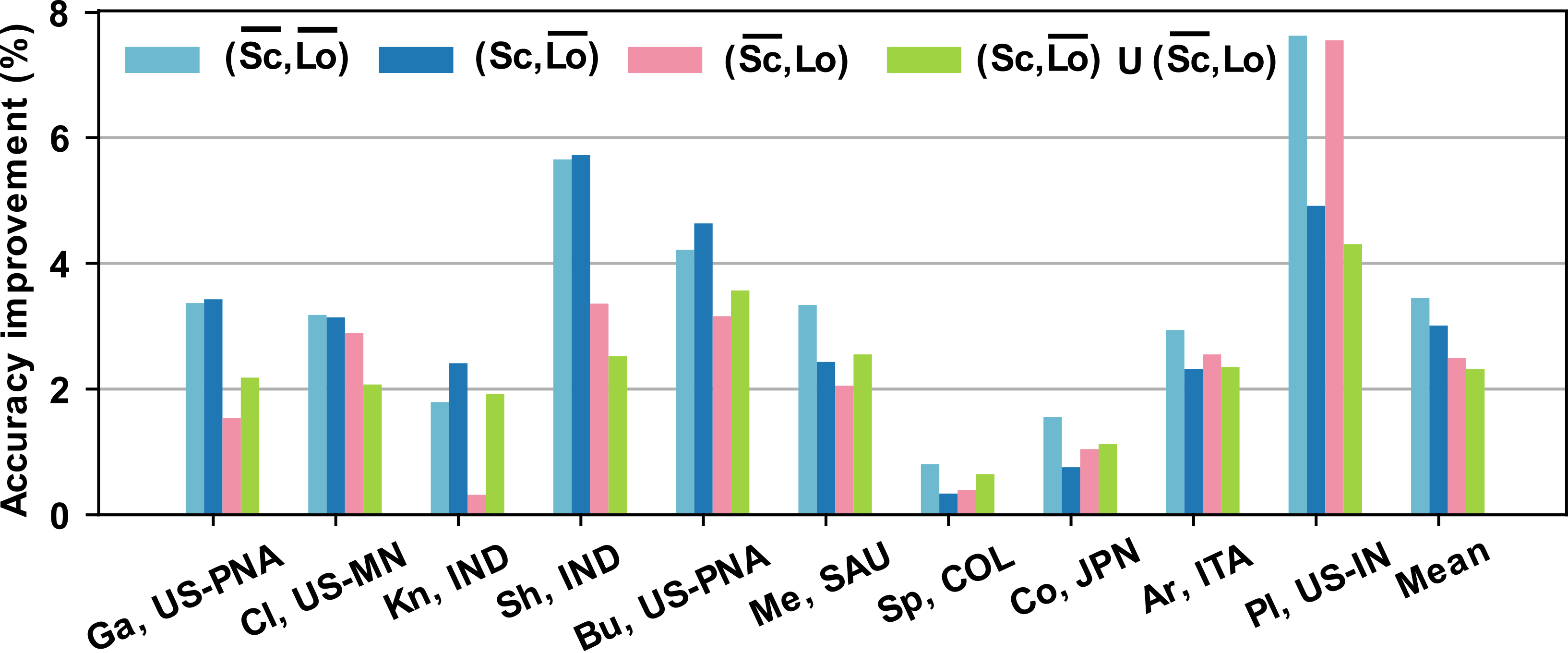}
    \caption{Accuracy improvement of CIR over ERM using the same training: (1) neither the test scenario nor location appears in training \textbf{($\overline{\text{Sc}}$,$\overline{\text{Lo}}$)}, (2) w/ scenario samples (\textbf{(${\text{Sc}}$,$\overline{\text{Lo}}$)}, (3), w/ location samples(\textbf{$\overline{\text{Sc}}$,Lo})), and (4) w/ both  ({\textbf{(Sc, $\overline{\text{Lo}}$)}}$\cup${\textbf{($\overline{\text{Sc}}$, Lo)}}).}
    \vspace*{-10pt}
    \label{fig:CIR_improvement}
\end{figure}

\noindent \textbf{CIR Ablation.} \method\ has two reconstruction objectives, and three architectural choices for reconstruction, which are ablated in \cref{tab:ablation}. 
For the two objectives, the one with the largest impact differs per split, with the classification reconstructions ($\mathcal{L}_{rc}$) performing better on average (shown by worse results when it is excluded). Both %however 
outperform the baseline (- $\mathcal{L}_{rc}$ - $\mathcal{L}_{rt}$) without reconstruction by a large margin. %Associating reconstructions with text ($\mathcal{L}_{rt}$) is complementary. Note that with neither of these, CIR is equivalent to ERM.
%that our method without text achieves a significant improvement, outperforming the baseline ERM baseline  (CIR-$\mathcal{L}_{rt}$-$\mathcal{L}_{rc}$) by 3.7\% on the selected splits. The reconstruction-text association loss improves by more than 2\% on ERM, while complementing $\mathcal{L}_{rc}$ to the overall CIR improvement.
%the reconstruction-text pairing loss ($\mathcal{L}_{rt}$) has a bigger impact, with the second reconstruction objective ($\mathcal{L}_{rc}$) being complementary. %This is because text models pre-trained on image-text pairs tend to focus on nouns (captured by $\mathcal{L}_{rt}$)~\cite{kazakos2021little}, whereas $\mathcal{L}_{rc}$ forces the reconstructions to be good verb (\ie action) representations.
We also ablate other decisions in the reconstruction.
Recall that $\oplus v$ is computed using learnt attention, while $\oplus v'$ is computed using cross-product attention.
We show the impact of reversing each of these decisions.
Finally, we show that sharing the same reconstruction ($\oplus v' = \oplus v$) and not sharing the classifier ($h \neq h'$) produces worse results.

\setlength{\tabcolsep}{3pt}

\noindent \textbf{Attention Masking.} 
%\tobyn{Fix table 3 first, and put in up to date results before I write anything.}
\method\ 
%relies on the network learning how to 
reconstructs each clip from others in the batch. On average,
a batch contains 11\% videos from the same scenario, 9\%
from same location and 3\% from both.
We do not restrict which samples to attend to, only avoiding reconstruction from the sample itself. 
In \cref{tab:mask}, we ablate possible masks of \textbf{S}ame \textbf{S}cenario/\textbf{L}ocation (\textbf{SS}, \textbf{SL}) or \textbf{O}ther \textbf{S}cenario/\textbf{L}ocation (\textbf{OS}, \textbf{OL}). 
% \tobya{Results obtained by attending to videos from different scenarios and/or locations performs best on average.
% On certain test splits, masking improves performance. For example, masking out examples from US-IN helps for (\textbf{Pl, US-IN}).
% % are generally better than the ones obtained by attending to the same scenario and/or location. 
% We do not use masking (which avoids the need for domain labels) but showcase its potential value when additional knowledge of the domain shift can be utilised.
% }
% \tobyb{
 Results obtained without masking are best on average, followed by results where the same/other scenario is masked. 
%When using a mask, results obtained by attending to videos from different scenarios and/or locations performs best on average. 
On certain splits, masking improves performance. For example, masking out samples from different locations helps for (\textbf{Ar, ITA}).
We do not use masking (which avoids the need for domain labels) 
but showcase its potential value when additional knowledge of the domain shift can be utilised. 
% are generally better than the ones obtained by attending to the same scenario and/or location. 
%We do not use masking, which obtains better overall results results and also avoids the need for domain labels, but showcase its potential value when additional knowledge of the domain shift can be utilised.
%}

\noindent \textbf{Effect of scenarios and locations on \method.} Figure \ref{fig:CIR_improvement} 
shows the top-1 accuracy improvement of \method\ over ERM 
when both methods have access to samples from test scenarios and locations. Four cases are evaluated: \textbf{($\overline{\text{Sc}}$,$\overline{\text{Lo}}$)}, \textbf{(${\text{Sc}}$,$\overline{\text{Lo}}$)}, (\textbf{$\overline{\text{Sc}}$,Lo}), and  {\textbf{(Sc, $\overline{\text{Lo}}$)}}$\cup${\textbf{($\overline{\text{Sc}}$, Lo)}}.  
CIR improves over ERM in every case and every split.
The improvement is largest on the hardest case \textbf{($\overline{\text{Sc}}$,$\overline{\text{Lo}}$)}. %However \method\ at times improves significantly with the additional knowledge of scenario (as in \textbf{Bu, US-PNA}) or location (as in \textbf{Pl, US-IN}).

%All cases show an improvement, with adding scenarios giving more of a benefit than locations.

%\noindent \textbf{Video representation.} In \cref{tab:video_model} we compare the MLP used for $f$ to a Transformer encoder (4 layer, 4 heads). While \method\ still improves by up to $3\%$, results are worse than using a MLP. 
%Transformers typically require massive data to train, which can explain this drop in performance. 
%This is because Transformer introduces additional complexity which is not helpful in our setting. 

%\noindent \textbf{Query and Key embedding dimensions. }We compare in \cref{tab:q_k_dim} different dimensions for query and key cross-attention embeddings, showing that our \method\ is not sensitive to embedding dimension variations. We choose $128$ as the one performing better on average across all splits. 

% \noindent \textbf{Impact of text.}

\noindent \textbf{Support-Set Size.} In \cref{tab:batch_size} we show how \method\ is affected by the size of the batch, which determines the size of the support set used for reconstruction. \method\ is relatively stable over a range of sizes, with slightly worse performance for very small or very large batch sizes. 

% If you use beamer only pass "xcolor=table" option, \ie \documentclass[xcolor=table]{beamer}
% \usepackage[normalem]{ulem}
\setlength{\tabcolsep}{5pt}

\begin{table}[t]
\centering
\footnotesize

\begin{tabular}{cccccc
>{\columncolor[HTML]{F3F3F3}}c }
\toprule
\multicolumn{1}{l}{} & \textbf{\begin{tabular}[c]{@{}c@{}}Cl\\ US-MN\end{tabular}} & \textbf{\begin{tabular}[c]{@{}c@{}}Bu\\ US-PNA\end{tabular}} & \textbf{\begin{tabular}[c]{@{}c@{}}Co\\ JPN\end{tabular}} & \textbf{\begin{tabular}[c]{@{}c@{}}Ar\\ ITA\end{tabular}} & \textbf{\begin{tabular}[c]{@{}c@{}}Pl\\ US-IN\end{tabular}} & \textbf{Mean} \\ \hline
\textbf{16}          & 23.90 &	22.99&		26.04&			23.87	  &28.46& 25.05                               \\
%\textbf{32}          & 24.58                                                       & 24.60                                                        & 26.06                                                  & 25.07                                                    & 28.59                                                      & 25.78                               \\
\textbf{64}          & 23.89 &	24.36 &	\textbf{26.54} &	24.98	& 28.97                  & 25.75                                \\
\textbf{128}         & \textbf{25.51}&	24.93 &	26.34 &	25.67&	\textbf{30.94}     & \cellcolor[HTML]{F3F3F3}\textbf{26.68}                    \\
\textbf{256}         & 25.00 &	\textbf{24.97} &	26.52 &	\textbf{25.96} &	30.61                                 & 26.61                                \\
%\textbf{512}         & 24.78                                                     & 24.72                                                       & 25.19                                                     & 25.28                                                     & 29.68                                                      & 25.93                                 \\
%\textbf{1024}        & 24.44                                                      & 25.15                                                        & 24.77                                                   & 25.4                                                    & 30.04                                                      & 25.96                               \\
\textbf{2048}        & 24.66 &	24.73 &	25.48 &	25.53 &	30.27 &	26.14                \\
\bottomrule
\end{tabular}
\vspace*{-6pt}
\caption{Effect of varying the batch size on \method.}
\label{tab:batch_size}
\resizebox{\linewidth}{!}{
\begin{tabular}{l|cccccc } 
\toprule
                    LM       & \textbf{\begin{tabular}[c]{@{}c@{}}Cl\\ US-MN\end{tabular}} & \textbf{\begin{tabular}[c]{@{}c@{}}Bu\\ US-PNA\end{tabular}} & \textbf{\begin{tabular}[c]{@{}c@{}}Co\\ JPN\end{tabular}} & \textbf{\begin{tabular}[c]{@{}c@{}}Ar\\ ITA\end{tabular}} & \textbf{\begin{tabular}[c]{@{}c@{}}Pl\\ US-IN\end{tabular}} & \textbf{Mean}  \\ \hline
\textbf{CLIP-ViT-B-32}  \cite{radford2021learning}   & \textbf{25.51}&24.93 & 	26.34& 25.67 &	\textbf{30.94}                                                     & \cellcolor[HTML]{F3F3F3}\textbf{26.68} \\
\textbf{all-mpnet-base-v2} \cite{song2020mpnet} & 25.15	& 25.01 &	26.30 & 	\textbf{25.73 }& 	30.71                                            & \cellcolor[HTML]{F3F3F3}26.58          \\
\textbf{all-miniLM-L6-v2} \cite{wang2020minilm} & 25.08	& \textbf{25.36} &	\textbf{26.36} &	25.45 &	30.50                                 & \cellcolor[HTML]{F3F3F3}26.55   \\
\bottomrule
\end{tabular}}
\vspace*{-6pt}
\caption{Comparison of pre-trained text models. }
\label{tab:text_model}
\footnotesize

\resizebox{\linewidth}{!}{
\color{black}\begin{tabular}{lc|cccccc}
\toprule
            & T    &  \textbf{\begin{tabular}[c]{@{}c@{}}Cl\\ US-MN\end{tabular}} & \textbf{\begin{tabular}[c]{@{}c@{}}Bu\\ US-PNA\end{tabular}} & \textbf{\begin{tabular}[c]{@{}c@{}}Co\\ JPN\end{tabular}} & \textbf{\begin{tabular}[c]{@{}c@{}}Ar\\ ITA\end{tabular}} & \textbf{\begin{tabular}[c]{@{}c@{}}Pl\\ US-IN\end{tabular}} & \textbf{Mean}  \\ \hline
ERM  &     & 22.35 &	20.73 &	24.81 &	22.75 &	23.29 &	\cellcolor[HTML]{F3F3F3}22.78          \\
MMD*  &    & 23.60                                                     & 22.08                                                     & 25.87                                                      & 23.84                                                        & 24.78                                                      & \cellcolor[HTML]{F3F3F3}24.03          \\
Mixup  &    &22.21                                                     & 21.45                                                     & \textbf{{25.90}  }                                                    & 23.85                                                       & 24.41                                                      & \cellcolor[HTML]{F3F3F3}23.56          \\ 
%\method       &       &\textbf{ 24.83}                                                       &\textbf{ 24.80  }                                                      &{ 25.06    }                                                & \textbf{25.38  }                                                   & \textbf{29.50  }                                                    & \cellcolor[HTML]{F3F3F3}\textbf{25.91}  \\
\method       &       &\textbf{{24.52}  }                                                     &\textbf{{24.64}   }                                                   &{ 25.42    }                                                & \textbf{{25.71}}                                                   & \textbf{{30.17}}                                                    & \cellcolor[HTML]{F3F3F3}\textbf{26.09}  \\\midrule
ERM  & \cmark  &  23.32  & 23.30 & 25.84  & 24.31      & 27.32                                                                                                          & \cellcolor[HTML]{F3F3F3}24.82          \\
MMD*  & \cmark    & 23.69 &	23.43	& 25.90	& 24.27	&	27.66	& \cellcolor[HTML]{F3F3F3}24.99          \\
Mixup  & \cmark &23.94	& 22.94 &	25.45 &	24.71	&	28.52                      & \cellcolor[HTML]{F3F3F3}25.11         \\

%\method       & \cmark    & \textbf{25.10}                                              & \textbf{25.01}                                               & \textbf{26.74}                                            & \textbf{25.70 }                                                    & \textbf{30.67  }                                                     & \cellcolor[HTML]{F3F3F3}\textbf{26.64} \\
\method       & \cmark    & \textbf{{25.51} }                                             &\textbf{{24.93}}                                               & \textbf{{26.34} }                                           & \textbf{25.67}                                                   & \textbf{{30.94} }                                                    & \cellcolor[HTML]{F3F3F3}\textbf{26.68} \\
\bottomrule
\end{tabular}}
\vspace*{-6pt}
\caption{Impact of adding text to existing DG methods. T indicates text supervision. * requires additional domain label supervision.}
\label{tab:with_text}
\end{table}

\begin{figure}[t]
        \centering    
    \includegraphics[width=\columnwidth]{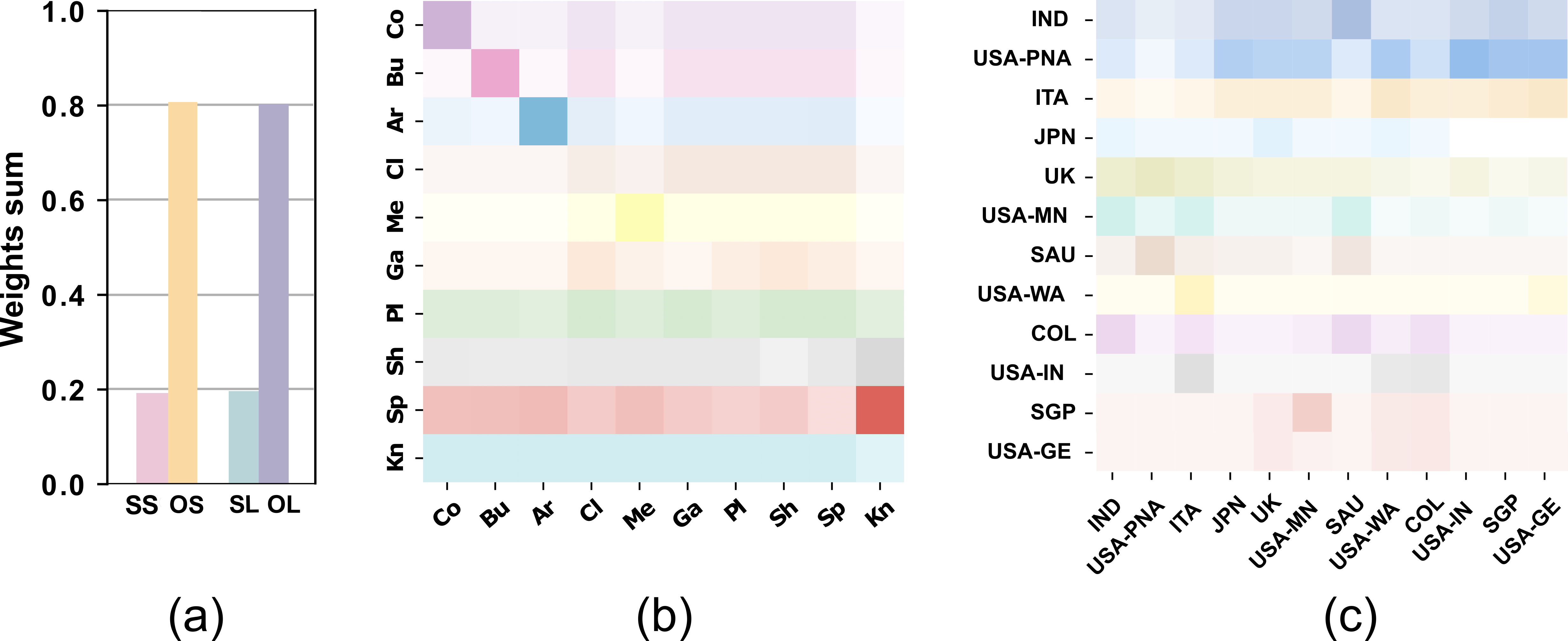}
    \vspace*{-16pt}
    \caption{Analysis of attention during reconstruction. (a) Normalised sum of attention weights over SS, OS, SL, OL. (b) Cross-scenario attention (c) Cross-location attention.}
        \vspace*{-10pt}
    \label{fig:cross-attention_1}
    
\end{figure}

 \begin{figure*}[t]
    \centering
    \includegraphics[width=\textwidth]{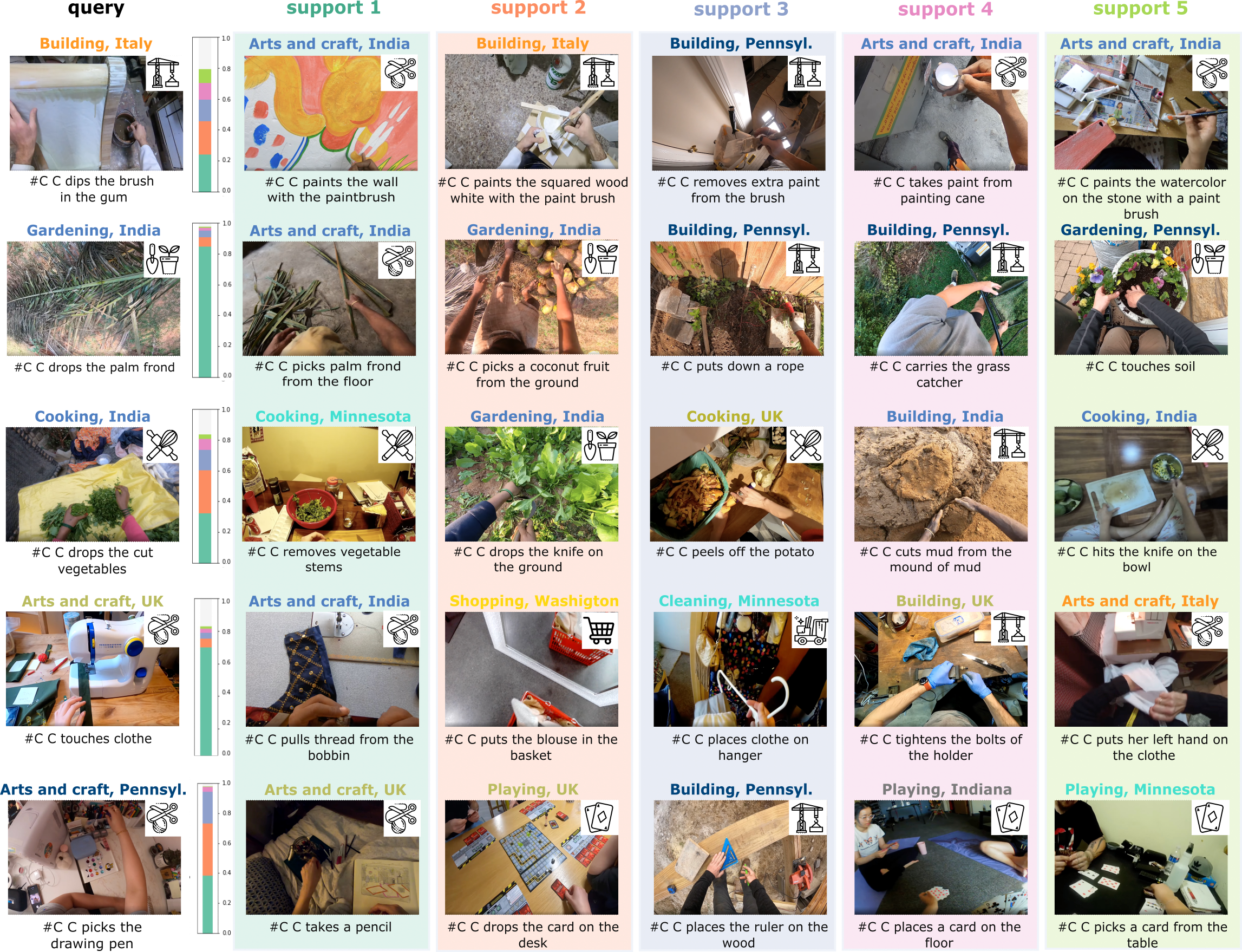}
    \vspace*{-18pt}
    \caption{\textbf{\method\ weights for reconstruction.} Five examples of cross-instance reconstruction from the training set. The query video is shown on the left. For each video, we show its corresponding scenario/location/narration. For each query, the bar shows the score of the \textit{j}-th support video (colour-matched) with white indicating the sum of the remaining scores from other samples. }
    \vspace*{-12pt}
    \label{fig:qualitative}
\end{figure*}

\noindent \textbf{Text models.} We compare the CLIP-ViT-B-32 text encoder to other pre-trained language models in \cref{tab:text_model}.  Results are comparable for different language models. 

\method\ exploits text narrations to help overcome domain shifts. Table \ref{tab:with_text} shows the benefit of this approach, and that merely adding video-text association to existing methods is insufficient. We add the text association loss $L_{rt}$, acting directly on video representations  (\ie no reconstruction) to existing DG methods.
We compare MMD, which performs second best after \method, and which requires domain labels. We also provide results for ERM and Mixup which do not require domain labels, and thus have the same level of supervision as \method. Importantly, \method\ \emph{without text} is better than other methods \emph{with text}.

\setlength{\tabcolsep}{3.3pt}

\setlength{\tabcolsep}{2pt}

\subsection{CIR Analysis}

Figure \ref{fig:cross-attention_1} analyses how videos attend to other videos during reconstruction-text association. (a) shows that videos primarily attend to other scenarios and locations, which helps to learn representations that generalise across domain shifts.
(b) shows attention between scenarios, with some strong self-attention (\eg cooking) as well as cross-attention (\eg sport attending to knitting). Certain scenarios attend evenly to all scenarios (\eg playing).
(c) shows attention between locations, which has fewer strong entries, suggesting that knowledge from all locations is helpful.

We show selected samples of our reconstructions during training in Fig.~\ref{fig:qualitative}. The Top-5 support set videos with the highest weights in the reconstruction (right) to the query video (left) obtained via \method\ ($c_{ij}$, \cref{sec:4.2}) are shown. \method\ is able to attend to samples belonging to other scenarios, other locations, and both.
For example, in the top row, a video of painting from a `Building' scenario in Italy is reconstructed using examples of `Arts and Crafts' in India, as well as `Building' from Italy. %The weights of the reconstruction are colour-coded in the bar. 

% \paragraph{CIR feature space.} We also visualise in \cref{fig:umaps_after} both training and test instances in the feature space after training with either ERM (left) or CIR (right). After training with CIR, the feature space is structured in a way that test instances overlaps with training instances, \eg {Rwanda} does not overlap with any of the training instances with ERM (left), while that does not happen with CIR (right). 
% %This shows that CIR helps matching each test instance from unseen domain with training instances of seen ones.
% This is because \method\ learns to represent instances as combinations of other domains, which in turn is helpful for out-of-distribution test samples (from unseen domains) to be matched to the seen distribution.
% %This is because the network has been trained to represent each instance as combinations of different domains. 
% %on both training and test features. 
% %The effect of 

\section{Conclusion}
\vspace*{-4pt}

In this paper, we introduced ARGO1M, a dataset for Action Recognition Generalisation Over scenarios and locations. %It contains over 1M video clips, capturing 60 action classes in various scenarios and geographic locations around the world.
We hypothesise that it is plausible to learn actions in a way that generalises to new scenarios (\eg an action `cut' in cooking can be used to recognise `cut' by a mechanic) in new locations (\eg the action `cut' in Italy can be used to recognise `cut' in India), as motivated by our paper's title.
%We showcase that current methods for generalisation, typically tested in images, do not produce favourable results in our complex settings.
We propose a method %, which we call Cross-Instance Reconstruction (\method), 
to reconstruct a video using samples from other scenarios and locations. In doing so, the learnt representation is generalisable to test splits with different scenarios and/or locations. \method\ consistently improves over baselines, and we offer extensive analysis and ablations.

The problem posed by \dataset\ is both practical and challenging. 
We hope this paper will foster further research on domain generalisation, which is under-explored in videos.

\vspace{-12pt}
%propose methods that work in the most challenging cases.

\paragraph{Acknowledgments.} Research at Bristol is supported by EPSRC Fellowship UMPIRE (EP/T004991/1) \& PG Visual AI (EP/T028572/1). We acknowledge the use of University of Bristol's Blue Crystal 4 (BC4) HPC facilities. We also acknowledge travel support from ELISE (GA no 951847).

%%%%%%%%% REFERENCES
{\small
\bibliographystyle{ieee_fullname}
\bibliography{egbib}
}

\appendix

\section{Dataset Curation}
\label{sec:dataset}

\begin{table*}[!ht]
\centering
\begin{tabular}{|p{2.5cm}|p{14cm}|}
\toprule
\toprule
{\textbf{Scenario}} & {\textbf{Ego4D Descriptions}} \\
\midrule
\rowcolor{cooking}
\textbf{Cooking}        & BBQing/picnics, Baker, Cooking, Making coffee, Outdoor cooking   \\                                                                                                                                                                                                 \rowcolor{building}                                                                           
\textbf{Building}       & Carpenter, Fixing something in the home, Handyman, Making bricks, Jobs related to construction/renovation company (director of work, tiler, plumber, electrician, handyman, etc)    \\                                                                               \rowcolor{arts}                                                                                      
\textbf{Arts and crafts} & Crafting/knitting/sewing/drawing/painting             \\                                                                                                                                                                                                              \rowcolor{cleaning}                                                                                     
\textbf{Cleaning}       & Car/scooter washing, Cleaning / laundry, Cleaning at the gym, Community cleaning, Daily hygiene, Household cleaners, Washing the dog / pet or grooming horse                                                                                                                                                                                            \\
\rowcolor{mechanic}          
\textbf{Mechanic}       & Assembling furniture, Bike mechanic, Blacksmith,  Car mechanic, Fixing PC, Getting car fixed,  Labwork, Maker Lab (making items in different materials, wood plastic and also electronics)- some overlap with construction etc. but benefit is all activities take place within a few rooms, Scooter mechanic, Working at desk,   Biology experiments \\
\rowcolor{gardening} 
\textbf{Gardening}      & Doing yardwork / shoveling snow, Farmer, Flower picking, Gardener, Gardening, Potting plants (indoor)                                                                                                                                                                                                                                                 \\
\rowcolor{playing} 
\textbf{Playing}        & Assembling a puzzle, Gaming arcade / pool / billiards, Playing darts, Playing board games, Playing cards, Playing games / video games, Practicing a musical instrument                                                                                                                                                                                \\
\rowcolor{shopping} 
\textbf{Shopping}       & Clothes and other shopping, Grocery shopping indoors, Working in milktea shop, Working in outdoor store                                                                                                                                                                                                                                                  \\
\rowcolor{sport} 
\textbf{Sport}          & Attending sporting events - watching and participating in, Baseball, Basketball, Bowling, Climbing, Cycling / jogging, Football, Going to the gym - (exercise machine, class, weights), Golfing, Hiking, Playing badminton, Roller skating, Rowing, Swimming in a pool/ocean, Working out at home, Working out outside                                  \\
\rowcolor{knitting} 
\textbf{Knitting}       & All videos from \textit{Arts and crafts} scenario, where \textit{at least} one narration contains keywords related to knitting activities.     \\                                                                                    \bottomrule
\bottomrule                                                                                
\end{tabular}
\caption{Our closed-form scenarios for \dataset, and corresponding Ego4D free-form descriptions.}
\label{tab:scenarios}
\end{table*}
In this section we detail our pipeline for curating \dataset. The process has three steps, (i) scenarios (\ref{sec:app:scenario}), (ii)~clip selection~(\ref{sec:app:clip}), and (iii) action classes (\ref{sec:app:class}).

%\def\arraystretch{1.5}

% \tobyn{This is duration. Maybe instances is better?}

\subsection{Scenarios} 
\label{sec:app:scenario}

We discard Ego4D videos with a missing scenario description (7.4\% of the total videos).
%, resulting in a set of 7484 videos with a unique scenario (77.6\% of the total), and 1441 videos (14.9\% of the total) with more than one scenario. 
Then, from the total of 136 free-form descriptions of scenarios provided by Ego4D, we choose the 62 that contain sufficient diversity and number of videos, excluding those that are repetitive and not representative of a specific activity, such as ``Talking," or ``On a screen". This results in a set of 6813 videos, which represents 83.1\% of the videos with at least one associated scenario. We also exclude videos marked to contain multiple scenarios.

We group the remaining scenario descriptions into 10 scenarios, each one containing similar activities, \eg ``brewing coffee" and ``making a sandwich" both belong to the scenario \textit{Cooking}. The resulting clustered scenarios are shown in Table \ref{tab:scenarios}.

% containing 65 percent of the total number of clips in Ego4D.

\subsection{Video Clips} \label{sec:app:clip}

Each selected video is provided with timestamp-level narrations, which describe the camera wearer’s actions and interactions with objects, for example the narration “\#C C puts the scraper down” with the timestamp 3.70s.  
We chose narrations from \textit{annotator\_1}, and only select actions which correspond to the camera-wearer, \ie those with narration tagged with \texttt{\#C}, ignoring those corresponding to actions performed by an external actor (tagged with \texttt{\#O}). 
%This results in 6799 videos and 1,701,189 narrations. 
We use a set of heuristics to filter out videos whose scenario metadata originally provided by Ego4D is incorrect.
%% OLD: We use a set of heuristics to filter out videos that are not associated with the correct scenario originally provided by Ego4D.
We do this by identifying a set of keywords that we expect to find in the corresponding scenario across video's narrations. 
%We also use the narrations to filter videos that are associated to with  incorrect scenario. 
We only keep videos whose narrations contain these keywords relevant to the scenarios which we manually curate. This yields a set of 6358 videos (93\% of the videos from the selected scenarios) and 1,637,810 narrations. 
%\begin{figure}[t]
%    \centering
%    \includegraphics[width=\columnwidth]{imgs/test_distribution.png}
%    \caption{\chiara{placeholder, needs to be nicer.} Frequency of the selected action verbs inside each test set.  }
 %   \label{fig:test_distribution}
%\end{figure}

Narrations in Ego4D are well-aligned with videos due to the use of a pause-and-narrate annotation procedure. This is noted in the Ego4D paper and by others~\footnote{Kevin Qinghong Lin, Alex Jinpeng Wang, Mattia Soldan, Michael
Wray, Rui Yan, Eric Zhongcong Xu, Difei Gao, Rongcheng Tu, Wenzhe
Zhao, Weijie Kong, Chengfei Cai, Hongfa Wang, Dima Damen, Bernard
Ghanem, Wei Liu, and Mike Zheng Shou. Egocentric Video-Language
Pretraining. NeurIPS, 2022}. To verify, we manually annotated action start times on a small subset and found an average offset of 0.6s between our action start times and the narration timestamps, and 0.9s between their endings.
This allows us to take the narration timestamp as the clip start time, and the timestamp of the next narration as the clip end time.
Like prior efforts, where action boundaries can be more relaxed given they contain the relevant action (\eg Kinetics \cite{carreira2017quo}), we find these boundaries to be sufficient for training and evaluation of action recognition. We next describe how clips are associated with class labels.

\subsection{Action Classes}
\label{sec:app:class}

%\tobyn{Why do we filter actions?}
%Our objective is to deal with the domain shift that occurs when the same action is performed on various objects, with various tools, and utilizing various motions, depending on the scenario in which it is performed. Due to this, we intentionally choose a subset of relevant actions whose intrinsics change depending on the context in which the action is carried out. 

Action labels are extracted from the verbs in the Ego4D narrations using spaCy \cite{spacy2}. We parse narrations into verbs and nouns. We take the  verb as the candidate action, and group these verbs using the EPIC-KITCHENS-100 \cite{damen2020rescaling} taxonomy, with some manual changes to handle the larger range of activities in Ego4D (Table \ref{tab:verbs}). For example, similar actions such as ``take" and ``pick" are grouped into one class.
We exclude ambiguous actions (\eg ``adjust") and those which do not interact with the surroundings (\eg ``look at"). We also exclude actions which occur too infrequently to train for domain generalisation. This process leaves a set of 60 action classes (shown in Fig. 2 in the main paper) and 1,050,371 instances. 

\dataset\ accordingly has 1,050,371 video clips from 5894 videos, which correspond to 42\% of all Ego4D clips and 61\% of all selected videos in Ego4D.

\noindent \textbf{Note:} While curating \dataset, we noticed a common pattern throughout Ego4D narrations: the actions ``put" and ``drop" were often used interchangeably, and often incorrectly. We hypothesise this is a result of non-native narrators, but could be due to the subjective choice of words. When we examined the average statistics on the validation set, we found that the network incorrectly predicted ``put" instead of ``drop" for approximately 16\% of the total number of ``drop" samples, and incorrectly predicted ``drop" instead of ``put" around 10\% of the time.
We acknowledge these annotation inconsistencies as a limitation. 
There might be other limitations in narrations we are not aware of. 
Importantly, we believe these ambiguities offer good practice in avoiding clips that achieve easy consensus. This allows for videos that present more challenging situations where actions are difficult to recognise~\footnote{M. Monfort et al. ”Moments in time dataset: one million videos for
event understanding.” T-PAMI, 2019.}.

\subsection{ARGO1M Feature Distribution}
\label{sec:umaps}
\begin{figure*}[t!]
    \centering
    \includegraphics[width=\linewidth]{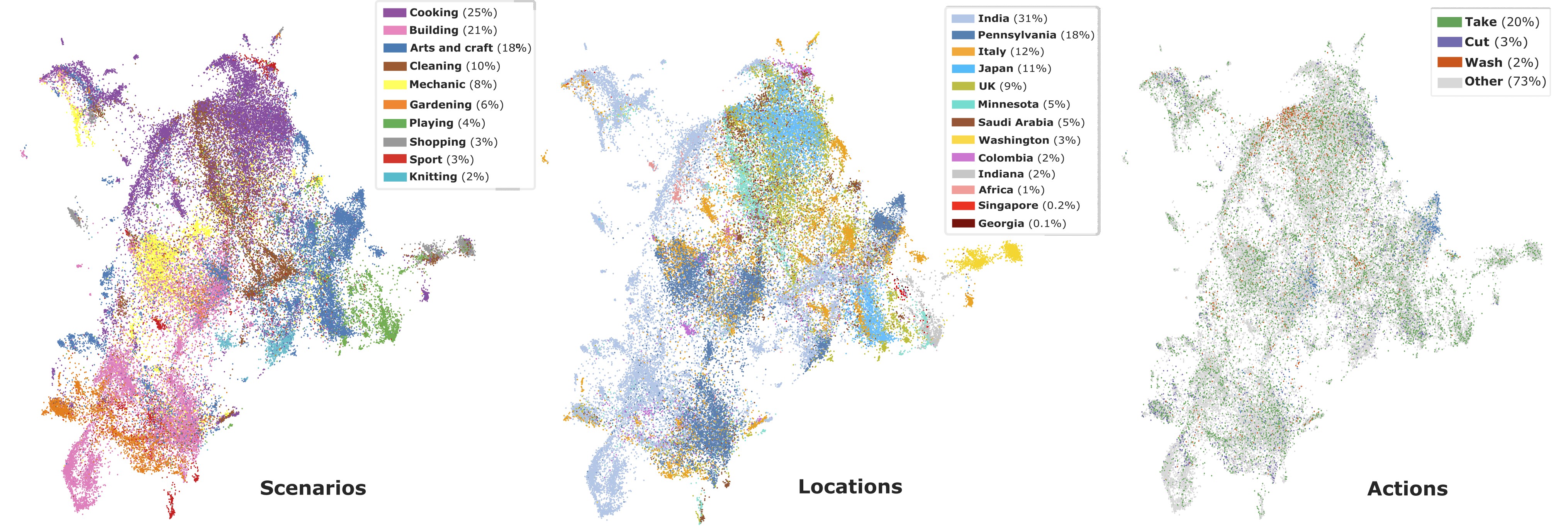}
   \caption{UMAPs of \dataset\ features across scenarios (left), locations (center) and for three action classes (right). We use the same projection to show correspondence across the three UMAP plots. 
}
\label{fig:umaps}
\end{figure*}
%\vspace{-15pt}
\label{sec:umaps}
Figure \ref{fig:umaps} visualises the feature distribution of all samples in \dataset\ across scenarios (left), geographic locations~(center) and action classes (right). For better clarity, in the action class plot we visualise 3 out of 60 classes and indicate the remaining ones as \textit{others}. These features are obtained by a SlowFast network~\cite{feichtenhofer2019slowfast} pre-trained on Kinetics~\cite{carreira2017quo} and are visualised through UMAP. 

There is evidence of \textit{scenarios clustering in different locations}, e.g., \textit{Playing} (green cluster at the right of the feature map) corresponds to different locations (\textit{United Kingdom}, \textit{Minnesota} and \textit{Indiana}), and \textit{locations clustering in different scenarios}, e.g., \textit{Minnesota} (yellow cluster on the right) corresponds to multiple scenarios (mostly \textit{Cleaning} and \textit{Shopping}). This shows that scenario and location shifts cannot be handled independently or disentangled easily, and that considering (scenario,  location) pairings as test domains better captures the combined scenario/location shift properties.

\begin{figure*}[!ht]
\centering
\begin{minipage}[t]{.49\linewidth}
    \subfloat{
    \resizebox {\linewidth} {!} {
\begin{tikzpicture}
\pgfplotsset{every axis legend/.append style={
at={(0.5,1.03)},
anchor=south}}
% LEFT AXIS
% =========
\begin{axis}[
  enlargelimits=false,
  ylabel={\large Top-1 Accuracy (\%)},
  xlabel={ $\lambda_1$},
   xmin=0, xmax=1.6,
   ymin=22, ymax=32,
  xtick={0.1,0.5, 1, 1.5},
  ytick={18,20,22,24,26,28,30},
  legend style ={ at={(1,1)}, 
        anchor=north west, draw=black, 
        fill=white,legend columns=1,legend cell align=left,draw=none},
  ymajorgrids=true,
  grid style=dashed,
  width=10cm,
  height=6cm,
  legend columns=1,
every axis plot/.append style={ultra thick},
every mark/.append style={mark size=50pt}
]
\addplot[
  color=MaterialGreen800,
  mark=*]
table[x index=0,y index=1,col sep=comma]
{imgs/tables/cl_1.txt};
% \addlegendentry{Single-DG-RNA} 
\addlegendentry{Cl,US-MN} 
\addplot[
  color=MaterialBlue800,
  mark=*]
table[x index=0,y index=1,col sep=comma]
{imgs/tables/bu_1.txt};
% \addlegendentry{Single-DG-RNA} 
\addlegendentry{Bu,US-PNA} 
\addplot[
  color=MaterialYellow800,
  mark=*]
table[x index=0,y index=1,col sep=comma]
{imgs/tables/jc_1.txt};
% \addlegendentry{Single-DG-RNA} 
\addlegendentry{Co,JPN} 
\addplot[
  color=MaterialOrange800,
  mark=*]
table[x index=0,y index=1,col sep=comma]
{imgs/tables/ar_1.txt};
% \addlegendentry{Single-DG-RNA} 
\addlegendentry{Ar,ITA} 
\addplot[
  color=MaterialRed800,
  mark=*]
table[x index=0,y index=1,col sep=comma]
{imgs/tables/pl_1.txt};
\addlegendentry{Pl,IND} 
\addplot[
  color=MaterialBlack,
  mark=*,dotted]
table[x index=0,y index=1,col sep=comma]
{imgs/tables/avg_2.txt};
% \addlegendentry{Single-DG-RNA} 
\addlegendentry{Mean} 
% \addlegendentry{Single-DG-RNA} 
%\addlegendentry{Sport in Colombia} 
%\legend{Japanese Cooking, Rwanda, Mechanic, Sport, Knitting, Mechanic in Colombia, Sport in Colombia}
% An \addplot for each line
\end{axis}
%\caption{Different performance (average Top-1 Accuracy (\%)) based on the value of $\lambda$ used to weight HNA and RNA$^{sub}$ and RNA losses.}
%\label{fig:lambda_var_1}
%\vspace{-0.4cm}
\node[above,font=\large\bfseries] at ([xshift=1pt]current bounding box.north) {$\lambda_2 = 0.5$};
\end{tikzpicture}
}
    \label{fig:ssa_vs_gc_params}
    }
\end{minipage}%
\begin{minipage}[t]{.49\linewidth}
    \subfloat{
    \resizebox {\linewidth} {!} {
\begin{tikzpicture}
\pgfplotsset{every axis legend/.append style={
at={(0.5,1.03)},
anchor=south}}
% LEFT AXIS
% =========
\begin{axis}[
  enlargelimits=false,
  ylabel={\large Top-1 Accuracy (\%)},
  xlabel={ $\lambda_2$},
   xmin=0, xmax=1.6,
   ymin=22, ymax=32,
  xtick={0.1,0.5, 1, 1.5},
  ytick={18,20,22,24,26,28,30},
  legend style ={ at={(1,1)}, 
        anchor=north west, draw=black, 
        fill=white,align=left,legend cell align=left},
  ymajorgrids=true,
  grid style=dashed,
  width=10cm,
  height=6cm,
  legend columns=1,
legend style={draw=none},
every axis plot/.append style={ultra thick},
every mark/.append style={mark size=50pt}
]
\addplot[
  color=MaterialGreen800,
  mark=*]
table[x index=0,y index=1,col sep=comma]
{imgs/tables/cl.txt};
% \addlegendentry{Single-DG-RNA} 
\addlegendentry{Cl,US-MN} 
\addplot[
  color=MaterialBlue800,
  mark=*]
table[x index=0,y index=1,col sep=comma]
{imgs/tables/bu.txt};
% \addlegendentry{Single-DG-RNA} 
\addlegendentry{Bu,US-PNA} 
\addplot[
  color=MaterialYellow800,
  mark=*]
table[x index=0,y index=1,col sep=comma]
{imgs/tables/jc.txt};
% \addlegendentry{Single-DG-RNA} 
\addlegendentry{Co,JPN} 
\addplot[
  color=MaterialOrange800,
  mark=*]
table[x index=0,y index=1,col sep=comma]
{imgs/tables/ar.txt};
% \addlegendentry{Single-DG-RNA} 
\addlegendentry{Ar,ITA} 
\addplot[
  color=MaterialRed800,
  mark=*]
table[x index=0,y index=1,col sep=comma]
{imgs/tables/pl.txt};
\addlegendentry{Pl,IND} 
\addplot[
  color=MaterialBlack,
  mark=*,dotted]
table[x index=0,y index=1,col sep=comma]
{imgs/tables/avg.txt};
% \addlegendentry{Single-DG-RNA} 
\addlegendentry{Mean} 
% An \addplot for each line
\end{axis}
%\caption{Different performance (average Top-1 Accuracy (\%)) based on the value of $\lambda$ used to weight HNA and RNA$^{sub}$ and RNA losses.}
%\label{fig:lambda_var_1}
%\vspace{-0.4cm}
\node[above,font=\large\bfseries] at ([xshift=1pt]current bounding box.north) {$\lambda_1 = 0.5$};
\end{tikzpicture}
}
    }
\end{minipage}
\caption{
Average Top-1 accuracy of CIR, over test splits, as we vary the loss weighting hyper-parameters. Left: Varying $\lambda_1$ (left) while keeping $\lambda_2 = 0.5$; as well as varying $\lambda_2$ (right) while keeping $\lambda_1=0.5$.}
\label{fig:lamda}
\end{figure*}
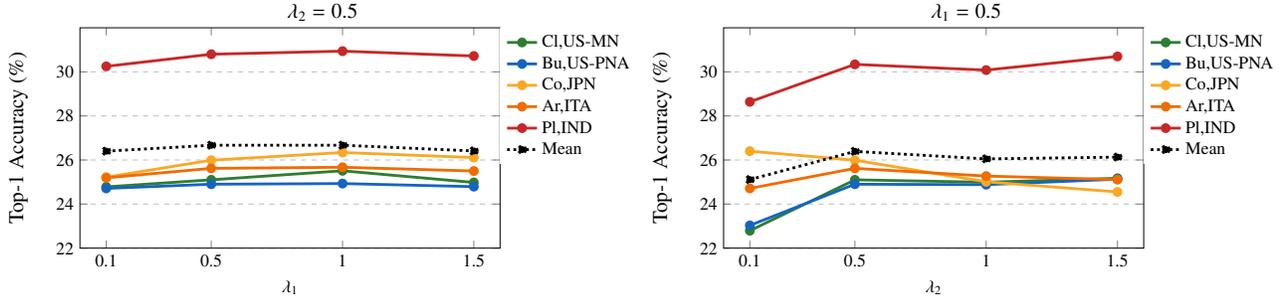

While it is easy to distinguish clusters of scenarios and locations, action classes are spread. We show that with the three actions `take', `cut' and `wash' that all are spread across the feature map. This shows the complexity of the proposed generalisation task.

\section{Additional Ablations}
\label{sec:ablations}

%\subsection{Details on hyper-parameter choice}
We use the \textbf{validation set} to select the best hyper-parameters for each algorithm. For each split, 
the validation set is a random 10\% of the training set, and thus contains no examples from the test scenario nor location.
Importantly, the split is on video basis, meaning that all clips from the same video are jointly present in either the training or the validation sets. 
We consider the performance over the split with biggest training and validation set (\textbf{Pl, US-IND}) for hyper-parameter optimisation.

%We use the \textbf{validation set} to select the best hyper-parameters for each algorithm. For each split, 
%the validation set is a random 10\% of the training set, and thus contains no examples from the test scenario nor location.
%Importantly, the split is on video basis, meaning that all clips from the same video are jointly present in either the training or the validation sets. 
%We consider the average over all the validation sets for hyper-parameter optimisation.

%The validation set allows us to assess performance when both the scenario and location are present in the training set.

\subsection{Ablation on $\lambda$ values}
\label{sec:lambda}

We assess how CIR results vary as we change $\lambda_1$ and $\lambda_2$, which weigh $\mathcal{L}_{rt}$ and $\mathcal{L}_{rc}$ respectively (Eq. 5 of the main paper). 
%Table \ref{tab:lambda} shows all the results for each weight combination of $\lambda_1$ and $\lambda_2$. 
For hyper-parameter selection, we chose the $\lambda_1$ and $\lambda_2$ values achieving the best results on the {validation} set ($\lambda_1$=1, $\lambda_2$=0.5).
In Fig.~\ref{fig:lamda}, we plot performance as we vary both $\lambda_1$ and $\lambda_2$ on the test splits.
We average performance on the test splits that we use for ablations in the main paper. When $\lambda_1$ variations are shown, $\lambda_2$ is set to $0.5$, and vice-versa.
Overall, performance is more sensitive to $\lambda_2$ than $\lambda_1$.
In both cases, performance drops for lower and higher values. 

%There is a clear trend of performance being lower for low or high values of $\lambda_1$ or $\lambda_2$, with performance being more stable on intermediate values. 
% \usepackage[normalem]{ulem}
% \usepackage[normalem]{ulem}

 \setlength{\tabcolsep}{3pt}

\begin{table}[ht!]
\resizebox{\linewidth}{!}{
\begin{tabular}{llccccc}
\bottomrule
\multicolumn{1}{l}{\textbf{Seed}} & \multicolumn{1}{l}{\textbf{Method}} &\textbf{\begin{tabular}[c]{@{}c@{}}Cl\\ US-MN\end{tabular}} & \textbf{\begin{tabular}[c]{@{}c@{}}Bu\\ US-PNA\end{tabular}} & \textbf{\begin{tabular}[c]{@{}c@{}}Co\\ JPN\end{tabular}} & \textbf{\begin{tabular}[c]{@{}c@{}}Ar\\ ITA\end{tabular}} & \textbf{\begin{tabular}[c]{@{}c@{}}Pl\\ US-IN\end{tabular}}  \\ \hline
                                  & \cellcolor[HTML]{FFF2CC}ERM & \cellcolor[HTML]{FFF2CC}22.35                   & \cellcolor[HTML]{FFF2CC}20.73                        & \cellcolor[HTML]{FFF2CC}24.81            & \cellcolor[HTML]{FFF2CC}22.75                & \cellcolor[HTML]{FFF2CC}23.29                \\
                                  & \cellcolor[HTML]{EAD1DC}MMD & \cellcolor[HTML]{EAD1DC}23.60                   & \cellcolor[HTML]{EAD1DC}22.08                        & \cellcolor[HTML]{EAD1DC}25.87            & \cellcolor[HTML]{EAD1DC}23.84                & \cellcolor[HTML]{EAD1DC}24.78                \\
\multirow{-3}{*}{\textbf{0}}      & \cellcolor[HTML]{D0E0E3}CIR & \cellcolor[HTML]{D0E0E3}\textbf{25.51}                   & \cellcolor[HTML]{D0E0E3}\textbf{24.93}                       & \cellcolor[HTML]{D0E0E3}\textbf{26.34}            & \cellcolor[HTML]{D0E0E3}\textbf{25.67}                & \cellcolor[HTML]{D0E0E3}\textbf{30.94}                \\ \hline
                                  & \cellcolor[HTML]{FFF2CC}ERM & \cellcolor[HTML]{FFF2CC}22.31                   & \cellcolor[HTML]{FFF2CC}21.09                        & \cellcolor[HTML]{FFF2CC}25.29            & \cellcolor[HTML]{FFF2CC}22.91                & \cellcolor[HTML]{FFF2CC}23.91                \\
                                  & \cellcolor[HTML]{EAD1DC}MMD & \cellcolor[HTML]{EAD1DC}23.87    & \cellcolor[HTML]{EAD1DC}{\color[HTML]{000000} 22.51} & \cellcolor[HTML]{EAD1DC}25.70             & {\cellcolor[HTML]{EAD1DC}}23.81 & {\cellcolor[HTML]{EAD1DC}}24.66 \\
\multirow{-3}{*}{\textbf{1}}      & \cellcolor[HTML]{D0E0E3}CIR & \cellcolor[HTML]{D0E0E3}\textbf{25.39}                   & \cellcolor[HTML]{D0E0E3}{\color[HTML]{000000} \textbf{25.01}} & \cellcolor[HTML]{D0E0E3}\textbf{25.83}            & \cellcolor[HTML]{D0E0E3}\textbf{25.79}               & \cellcolor[HTML]{D0E0E3}\textbf{30.41}                \\
\hline
                                  & \cellcolor[HTML]{FFF2CC}ERM & \cellcolor[HTML]{FFF2CC}22.30                   & \cellcolor[HTML]{FFF2CC}{\color[HTML]{000000} 20.86} & \cellcolor[HTML]{FFF2CC}24.89            & \cellcolor[HTML]{FFF2CC}22.91                & \cellcolor[HTML]{FFF2CC}23.40                \\
                                  & \cellcolor[HTML]{EAD1DC}MMD & \cellcolor[HTML]{EAD1DC}{\color[HTML]{FF0000} }23.66 & \cellcolor[HTML]{EAD1DC}{\color[HTML]{000000} 22.36} & \cellcolor[HTML]{EAD1DC}25.92            & {\cellcolor[HTML]{EAD1DC}}23.59 & {\cellcolor[HTML]{EAD1DC}}24.60 \\
\multirow{-3}{*}{\textbf{2}}      & \cellcolor[HTML]{D0E0E3}CIR & \cellcolor[HTML]{D0E0E3}\textbf{25.69}                   & \cellcolor[HTML]{D0E0E3}{\color[HTML]{000000} \textbf{25.02}}  & \cellcolor[HTML]{D0E0E3}\textbf{26.01}            & \cellcolor[HTML]{D0E0E3}\textbf{25.66}                & \cellcolor[HTML]{D0E0E3}\textbf{30.42}                \\ \hline
                                  & \cellcolor[HTML]{FFF2CC}ERM & \cellcolor[HTML]{FFF2CC}22.43                   & \cellcolor[HTML]{FFF2CC}{\color[HTML]{000000} 20.41} & \cellcolor[HTML]{FFF2CC}25.14            & \cellcolor[HTML]{FFF2CC}23.12                & \cellcolor[HTML]{FFF2CC}23.68                \\
                                  & \cellcolor[HTML]{EAD1DC}MMD & {\cellcolor[HTML]{EAD1DC}} 23.66   & \cellcolor[HTML]{EAD1DC}{\color[HTML]{000000} 22.22} & \cellcolor[HTML]{EAD1DC}25.96            & {\cellcolor[HTML]{EAD1DC}}23.60 & {\cellcolor[HTML]{EAD1DC}}24.53 \\
\multirow{-3}{*}{\textbf{3}}      & \cellcolor[HTML]{D0E0E3}CIR & \cellcolor[HTML]{D0E0E3}\textbf{25.49}                   & \cellcolor[HTML]{D0E0E3}\textbf{25.05}                        & \cellcolor[HTML]{D0E0E3}\textbf{26.16}            & \cellcolor[HTML]{D0E0E3}\textbf{25.28}                & \cellcolor[HTML]{D0E0E3}\textbf{30.41}  \\
\bottomrule
\end{tabular}}
\caption{Results of ERM and CIR on 4 different seeds.}
\label{tab:seeds}
\end{table}

\subsection{Seed variations}
\label{sec:seed}
\begin{figure}
    \centering
    \includegraphics[width=\columnwidth]{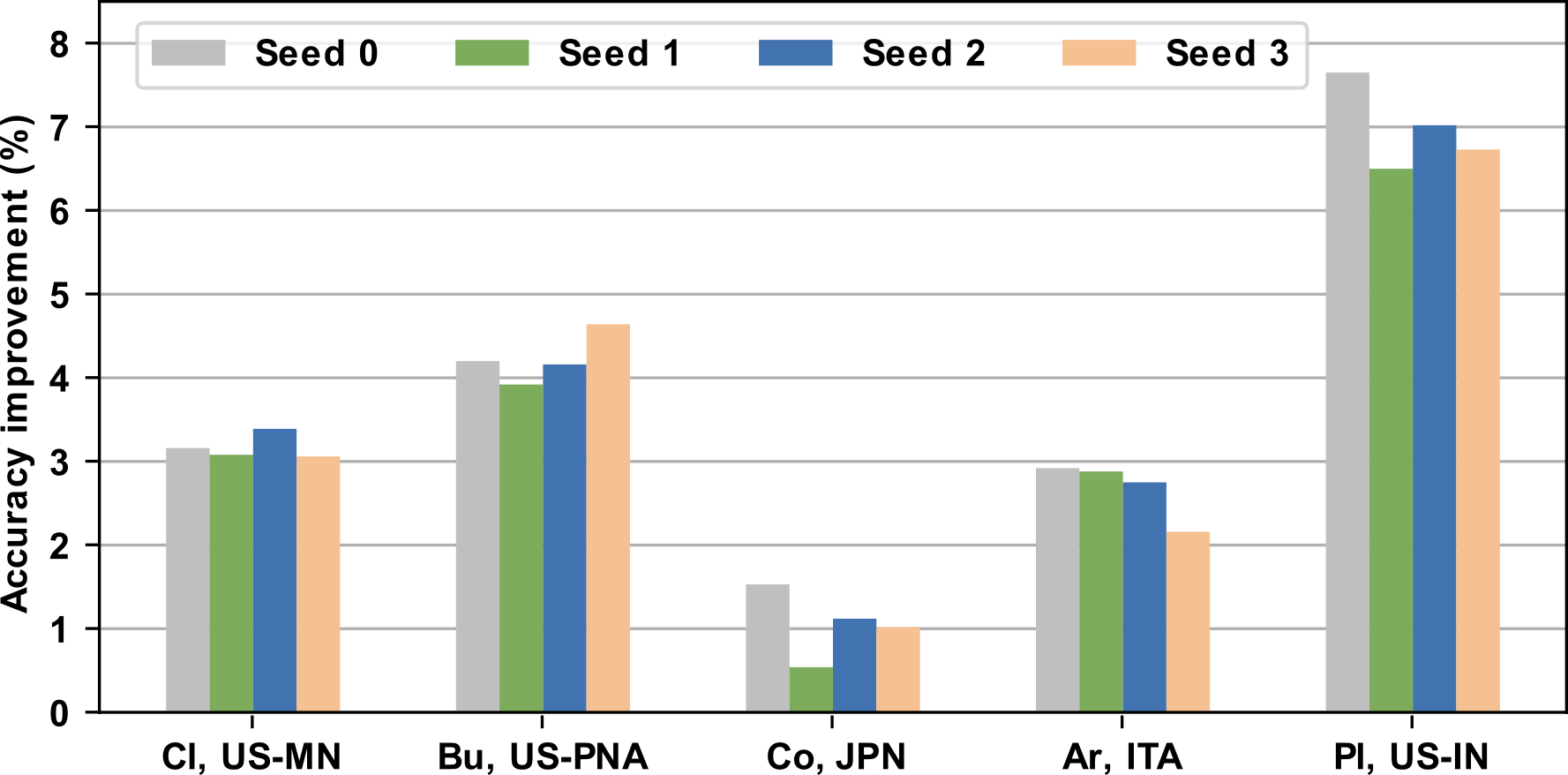}
    \caption{Improvement (\%) of \method\ w.r.t. ERM on the different seed. Main results are on seed 0, where the best ERM results are achieved on the validation set. }
    \label{fig:seeds}
\end{figure}

All the results in the main paper are compared on one fixed seed for direct comparison across baselines. We select the seed achieving best results on ERM by optimising on the validation set (seed 0). To show performance stability, we run results of ERM, \method, and MMD (the best performing baseline) on 4 seeds. 
In Table~\ref{tab:seeds}, we showcase the results confirming CIR is consistently achieving best performance on every split and every seed.
For easy comparison, we plot the relative improvement of \method\ over ERM over all seeds and on the five largest test splits in Fig. \ref{fig:seeds}.
Figure shows consistent improvement over ERM across the seeds.
%On every seed and each of our 5 test splits, our \method\ outperforms ERM.

\subsection{Support-Set Size}
\label{sec:support}
Due to space limitations, we do not include all batch sizes in the ablations of Table 5 in the main paper.
We provide the complete set of results in Table \ref{tab:batch_size}.
Results showcase continuous improvement as we increase the batch size up to 128, with a slight drop for larger batches.
% the full version on all batch sizes from Table 5 in Section 5.2 (``Ablations") in the main paper. 

\setlength{\tabcolsep}{5pt}

\begin{table}[!t]
\centering
\resizebox{\linewidth}{!}{
\begin{tabular}{cccccc
>{\columncolor[HTML]{F3F3F3}}c }
\toprule
\multicolumn{1}{l}{} & \textbf{\begin{tabular}[c]{@{}c@{}}Cl\\ US-MN\end{tabular}} & \textbf{\begin{tabular}[c]{@{}c@{}}Bu\\ US-PNA\end{tabular}} & \textbf{\begin{tabular}[c]{@{}c@{}}Co\\ JPN\end{tabular}} & \textbf{\begin{tabular}[c]{@{}c@{}}Ar\\ ITA\end{tabular}} & \textbf{\begin{tabular}[c]{@{}c@{}}Pl\\ US-IN\end{tabular}} & \textbf{Mean} \\ \hline
\textbf{16}          & 23.90	& 22.99	& 26.04 & 	23.87&	28.46	& 25.05           \\
\textbf{32}       &   23.54	& 22.78	& 26.40& 	24.38& 	28.12& 25.05                             \\
\textbf{64}          & 23.89 &	24.36 &	\textbf{26.54} &	{24.98} &	28.97 &	25.75        \\
\textbf{128}    &   \textbf{25.51}&	24.93&	26.34&	25.67	&\textbf{30.94}&	\textbf{26.68 }                \\
\textbf{256}         & 25.00	&24.97&	26.52&	\textbf{25.96}	&30.61&	26.61                              \\
\textbf{512}     &     24.95  &	24.82  &	26.15  &	24.02	  &30.88  &	26.16                                \\
\textbf{1024}        & 24.64	&\textbf{25.31}&	25.79&	25.87	&30.70&	26.46                  \\
\textbf{2048}        & 24.66 & 	24.73 & 	25.48 & 	25.53 & 	30.27 & 	26.14      \\
\bottomrule
\end{tabular}}
\caption{Effect of varying the batch size on \method.}
\label{tab:batch_size}
\end{table}

\section{Hyper-parameter search}
\label{sec:hyper-parameter}

In Table \ref{tab:hyper} we show the hyper-parameter search space for each of the baseline methods, highlighting the chosen hyper-parameters in \textbf{bold}. For CORAL~\cite{sun2016deep}, MMD~\cite{li2018domain}, DANN~\cite{ganin2016domain}, BoDA~\cite{yang2022multi} and DoPrompt~\cite{zheng2022prompt}, the overall loss is $\mathcal{L}=\mathcal{L}_c+\gamma_1 \mathcal{L}_{scen}+\gamma_2 \mathcal{L}_{loc}$, where $\mathcal{L}_c$ is the cross-entropy loss and $\mathcal{L}_{scen}$ and $\mathcal{L}_{loc}$ are the losses from these methods applied to scenarios and locations respectively. For example, $\mathcal{L}_{scen}$ is the MMD loss between scenarios when training MMD. %The search space for $\gamma_1$ and $\gamma_2$ is shown Table \ref{tab:hyper}. 
For domain-alignment methods (MMD, CORAL, BoDA), we apply the scenario and location alignment loss on the last layer features. 

For BoDA, we also set the hyper-parameter \texttt{nu} controlling the calibration distance to ${\texttt{nu}=1}$ (see details in \cite{yang2022multi}). 

The discriminator-based method DANN utilises two domain discriminators, consisting of 2 fully connected layers each. One discriminator is responsible for classifying scenarios, and one for classifying locations. Each of them has a gradient reversal layer with momentum term $\beta_1=0.5$\footnote{We followed A. Radford et al., 
Unsupervised Representation Learning with Deep Convolutional Generative Adversarial Networks. ICLR 2016}.%following~\cite{radford2015unsupervised}. 

For DoPrompt, we learn two separate prompts, one for each scenario, and one for each location. In addition to the weights $\gamma_1$ and $\gamma_2$ used for prompt learning, we also perform a grid search on the prompt length \textit{l}. 

For Mixup~\cite{wang2020heterogeneous}, which is the only baseline method that does not require domain labels, we perform an optimisation of the hyper-parameter $\alpha$ controlling the interpolation of mixed samples. 

For each method, we also optimise the learning rate in a search space of $\{10^{-6}, 10^{-5}, 10^{-4}, 10^{-3}\}$. We show in Table~\ref{tab:lr} the chosen learning rate (LR) for each baseline method.

\begin{table}[t!]
\centering
\footnotesize
\begin{tabular}{lcc}
\toprule
\rowcolor[HTML]{FFFFFF} 
{\color[HTML]{000000} \textbf{Method}}              & {\color[HTML]{000000} \textbf{Hyper-parameter}}                            & {\color[HTML]{000000} \textbf{Grid Search}}                                                   \\ \hline\noalign{\medskip}
CORAL &
$\gamma_1$,$\gamma_2$ & \begin{tabular}[c]{@{}c@{}}\{\textbf{0.1},0.5,1,1.5\},\{\textbf{0.1},0.5,1,1.5\}\end{tabular}\\\midrule
\begin{tabular}[c]{@{}c@{}}MMD\end{tabular} &
\begin{tabular}[c]{@{}c@{}}$\gamma_1$,$\gamma_2$\end{tabular}                                         &
\begin{tabular}[c]{@{}c@{}}\{{0.1},0.5,\textbf{1},1.5\},\{{0.1},\textbf{0.5},1,1.5\}\end{tabular} \\ \midrule
DANN                                                & \begin{tabular}[c]{@{}c@{}}$\gamma_1$,$\gamma_2$\\ Adam $\beta_1$\end{tabular} & \begin{tabular}[c]{@{}c@{}}\{\textbf{0.1},0.5,1,1.5\}, \{\textbf{0.1},0.5,1,1.5\}\\ \textbf{0.5}\end{tabular} \\ \midrule
Mixup                                               & $\alpha$                                                                   & \{0.1,\textbf{0.2},0.5\}                                                                      \\\midrule
BoDA                                                & \begin{tabular}[c]{@{}c@{}}$\gamma_1$,$\gamma_2$\\ \texttt{nu} \end{tabular}      & \begin{tabular}[c]{@{}c@{}}\{\textbf{0.1},0.5,1,1.5\}, \{\textbf{0.1},{0.5},1,1.5\}\\ \textbf{1}\end{tabular}   \\\midrule
DoPrompt                                            & \begin{tabular}[c]{@{}c@{}}$\gamma_1$, $\gamma_2$\\\textit{l}\end{tabular}           & \begin{tabular}[c]{@{}c@{}}\{{0.1},0.5,\textbf{1},1.5\}, \{{0.1},0.5,\textbf{1},1.5\}\\\{\textbf{4},{16},32\}\end{tabular} \\    
\bottomrule
\end{tabular}
\caption{Hyper-parameter search space for different algorithms. Best ones are in \textbf{bold}.}
\label{tab:hyper}
\resizebox{\linewidth}{!}{
\begin{tabular}{llcccccc}
\toprule
              & \textbf{ERM} & \cellcolor[HTML]{FFFFFF}{\color[HTML]{000000} \textbf{CORAL}} & \cellcolor[HTML]{FFFFFF}{\color[HTML]{000000} \textbf{DANN}} & \cellcolor[HTML]{FFFFFF}{\color[HTML]{000000} \textbf{MMD}} & \cellcolor[HTML]{FFFFFF}{\color[HTML]{000000} \textbf{Mixup}} & \cellcolor[HTML]{FFFFFF}{\color[HTML]{000000} \textbf{BoDA}} & \cellcolor[HTML]{FFFFFF}{\color[HTML]{000000} \textbf{DoPrompt}} \\ \hline\noalign{\smallskip}

LR & $10^{-4}$    & $10^{-5}$                                                     & $10^{-5}$                                                    & $10^{-5}$                                                   & $10^{-5}$                                                     & $10^{-5}$                                                    & $10^{-6}$            \\                             \bottomrule              
\end{tabular}}
\caption{Chosen learning rate (LR) for each baseline method.}
\label{tab:lr}
\end{table}
 \setlength{\tabcolsep}{5pt}

\begin{figure}[t]
    \centering
    \includegraphics[width=\columnwidth]{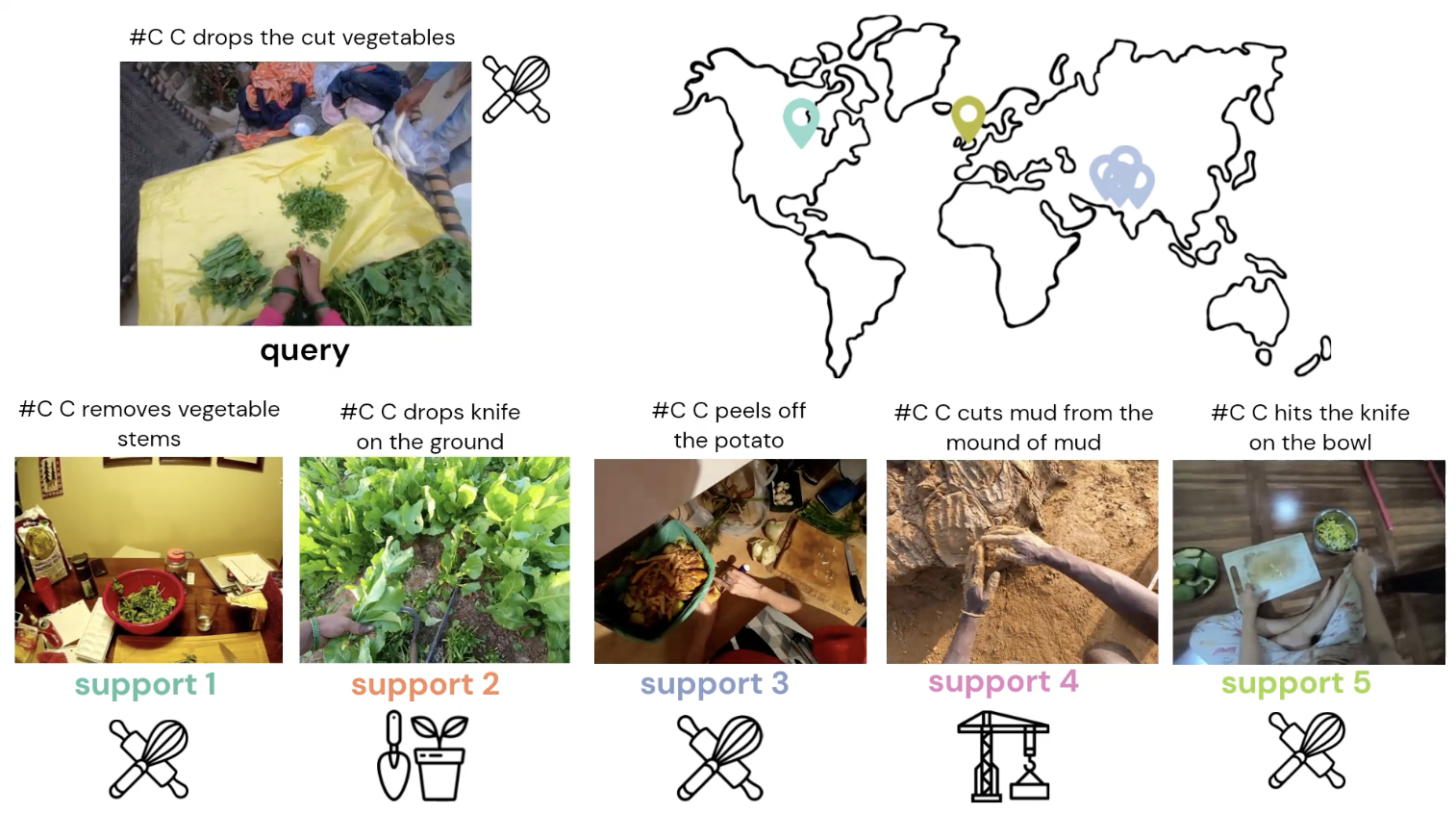}
    \caption{Preview of the video of qualitative results.}
    \label{fig:video}
\end{figure}

\section{Qualitative Results}
\label{sec:video}

In the video of qualitative results available at \url{https://chiaraplizz.github.io/what-can-a-cook/}, we visualise reconstructed instances from the training set by \method\ and their support set, which correspond to the examples in Figure 8 of the main paper. A preview of the video is shown in Fig. \ref{fig:video}. On the top left, we show the query video clip, along with the corresponding narration (top of the video), scenario (icon on the top-right of the video), and location (pin on the top-right map). Note that pin colours correspond to location colours in Fig. 2 of the main paper. On the bottom row, we show the \textit{j}-th support video clip, along with its narration (top of the video), scenario (icon below) and location (pin on the map).

The video, as in Figure 8 of the main paper, aims to showcase how one video clip during training is reconstructed from other video clips in the batch, potentially from other scenarios and/or locations.

\setlength{\tabcolsep}{5pt}

%\begin{figure*}[!ht]
%    \centering
 %   \includegraphics[width=\linewidth]{imgs/pdfs/bar_seeds.pdf}
 %   \caption{Improvement (\%) of \method\ w.r.t. ERM on the different seeds (x-axis). Main results are on seed 0, where the best ERM results are achieved on the Seen test set. The black horizontal line indicates the average improvement (\textbf{avg}) over all seeds for that specific test set. 
 %   Seed 0 is the fixed seed in all other experiments. }
  %  Seed 0 (best on ERM) is followed by other seeds, ordered by increasing average improvement over all test sets.}
%\label{fig:seed}
%\end{figure*}

% \usepackage[normalem]{ulem}

% \usepackage[normalem]{ulem}

%\nopagebreak
%\clearpage
\onecolumn
\nopagebreak
\renewcommand{\arraystretch}{1.5}
\setlength{\LTpre}{0pt}
\setlength{\LTpost}{0pt}

\section{Expanded version of Table 5}
\label{sec:full_tables}
\begin{tabularx}{\textwidth}{p{1.5cm}|p{15cm}} %{@{}lX@{}} 
%\label{tab:verbs}
%\bottomcaption{Selected action classes and corresponding open-vocabulary verbs. }
%\label{tab:verbs}
%\begin{longtable}{\textwidth}{|p{1.5cm}|p{15cm}|}
%
%\begin{tabular}{|p{1cm}|p{15cm}|}
\toprule
\textbf{Class} &\textbf{ Open-vocabulary Verbs}  \\ 
\midrule
%\endhead
\textbf{take}     & takes-along, takes-out, take-off, takes-inside, takes-at, takes-behind, takes-around, takes-underneath, cctakes, takes-without, take-with, takes-for, take-out, takes-from, takes-back, takeout, takers, takes-near, takes-into, takes-against, takes-like, takes-of, takes-by, takes-to, takes-toward, takes-beside, takes-below, take-at, takes-on, takes-towards, takes-down, takes-over, take-from, take, takes-among, takes-up, takes, partakes-in, takes-under, takes-fro, takes-in, takes-atop, takes-outside, overtakes, takes-off, takes-beneath, take-in, takes-with,pickes-on, picks-atop, picks-as, picks-by, picked-at, picking, picks-around, picks-near, pickup-from, picks-at, picks-fro, picked, picks-against, picks-down, picks-below, picks-on, picks-over, picks-without, cpicks, picks-beside, picks-into, picks-beneath, pick-from, picks-inside, picked-from, picks-for, picksfrom-with, picks-before, picks-back, picked-up, picking-from, picks-off, picks-from, picks-amongst, pick-on, picks-with, picked-with, picks-to, picks-outside, picka, pick-out, pick, pick-up, picking-on, picks-fom, picks-out, picks-under, picks-amidst, picks-up, pick-in, picks-toward, pick-with, pickes-with, picks, picks-onto, picking-up, picks-in, picks-underneath, picks-of, picks-wit, picked-on, picks-behind, picked-of, picked-beside, pickss-with, fetches-inside, fetches-into, fetch-with, fetch, fetches-with, fetches-to, fetches, fetches-under, fetches-from, fetchs, fetches-on, fetching-into, fetches-in, fetch-in, grabs-in, grabs-at, grabs-beside, grabbed, grabs-from, grabs-around, grabs-below, grabs-within, grabs-inside, grabs-for, grabs-on, grabs-of, grabs-with, grab, grabs, grabs-off, grabs-by, grabs-under ,gets-from,  pull-off, pulls-off, draws-out  \\  \nopagebreak
\textbf{put}      & puts-through, puts-agaisnt, put-of, inputs-into, puts-off, put-in, put-from, puta-into, puts-at, put-over, put-back, put-into, puts-alongside, puts-behind, putts, put-to, putting, {[}puts, put-down, puts-among, puts-under, puts-together, putting-on, put-under, puts-round, puts-beside, puts-to, puts-inot, puta-down, puts-away, puts-out, puts-against, put-on, puts-beneath, putting-in, puta-on, puts-n, puts-between, puts-towards, puta-in, sheputs-down, puts-underneath, puts-below, put-at, puts-of, put, puts-near, putting-down, oputs-in, put-underneath, p{[}puts-on, puts, puts-aside, puts-in, put-inside, put-beside, puts-with, puts-back, puts-into, inputs-on, puts-from, puts-by, puts-across, puts-around, put-between, puts-over, putting-into, puts-along, puts-above, puts-onto, puts-on, puts-down, puta, sheputs-in, puts-inside, places-round, places-onto, places-below, place-for, places-before, places-from, sheplaces, place-under, displaces-on, replaces-on, placers, places-by, places-underneath, places-under, places-at, places-back, places-with, places-near, placed-in, places-within, replaces-into, place-with, places-across, places-off, replaces-to, placed-beside, places-infront, places-behind, places-through, places-opposite, replace, place-on, sheplaces-on, replaces-with, places-against, replaces-from, places-of, placed, places-among, placers-on, places-over, placed-under, place, places-like, placers-down, places-along, places-above, places\#unsure-on, places-towards, places-atop, replaces, placed-with, displaces, places-down, places-into, place-into, places-beneath, places-beside, place-in, places-in, replaces-in, places-around, places-to, displaces-with, places-up, places-on, places-for, places-inside, places-between, place-down, placed-on, places, places-out, places-unto, displaces-in,repositions-against, reposition-in, positions-on, positions-under, repositions-with, repositions-in, reposition-with, repositions-across, repositions-on, repositions-at, repositions-amongst, positions-inside, repositions-out, positions-beside, positions-at, positions-along, repositions-under, positions-with, repositions-from, reposition, repositions-to, positions-against, repositions, positions-in, positions, repositions-atop, repositions-up \\ \nopagebreak
\textbf{drop}     & drops-off, drop-inside, drops-amongst, dropped-in, drops-om, drops-across, drops-outside, drops-forth, drops-back, dropes-into, dropes-in, drops-with, drops-from, drops-under, drops-infront, drops-by, shedrops-in, dropes, drop, drops-on, drops-like, drops-oon, drops-to, drops-in, drop-down, drope-into, drops-above, drops-inside, drops-onto, drops-beneath, dropd, drops, drops-between, drop-in, drops-unto, drops-down, drops-at, drops-below, drops-against, drops-near, drops-of, drops-around, drops-for, drop-into, dropping-on, drops-spout, drops-behind, drops-without, drop-with, drops-beside, dropes-on, drops-up, dropped-on, drops-out, drops-over, drops-into, shedrops, drop-on, ccdrops-on, dropes-down, drops-fro, drops-along, dropping, dropped, drops-towards, shedrops-on, drops-underneath, drops-atop  \\
\textbf{hold}     & holds-for, holds-around, holds-up, holds-into, holds-against, holds-beside, holds-by, hold-against, holds-between, holds-along, withholds, hold-in, holding, holds-to, holds-inside, holds-v, holds-down, holds-towards, holds-onto, holds, hold, holds-near, hold-on, holds-of, holds-at, holds-unto, holds-under, holds-over, hold-with, hold-up, holds-on, sheholds, holds-out, holds-from, holds-through, upholds, holding-with, holdswall-on, holding-in, holds-with, holds-in, holder, holds-atop, holds-w, hold-down, unholds-on, holder-with \\
\textbf{touch}    & touches-along, touches-behind, touches-near, touched-with, touches-against, touches-on, touches-f.with, touches-beneath, shetouches-in, touches-before, touches-below, toucheses, touches-to, touch-on, touches-off, touches-across, touchers, touches-beside, touchses\#unsured-with, touches-round, shetouches-on, touches, touches-inside, touches-of, touched-on, touches-up, touches-around, toucher, touches-underneath, touching, touches-down, touches-from, touches-by, touches-under, touch-in, touched, touch, touches-into, touches-at, shetouches, touches-above, touches-in, touch-with, touches-over, touches-with  \\
\textbf{remove}   & removes-beside, remove-around, removes-from, removes-under, remove-to, removes-of, remove-in, removes-towards, sheremoves-in, removes-off, removes-to, removes-round, remove-with, removes-over, removed, removets, removes-at, remove, removes-with, removes-on, removes-around, removes-underneath, removes-in, removers, removes-like, removes-near, removes-among, removes-below, remove-under, remover-on, removes-down, remover, removes-fom, removes-by, remove-from, removes-up, removes-out, removes-inside, removes-into, removes-against, removes, removers-from, remove-on, removes-between, remover-from, removes-for, sheremoves, removes-onto, removes-beneath, removed-from, removes-behind, takes-out, take-off, take-out,takeout,pick-out,picks-out, unplug, unplugs-in, unplugs-from, unplugs-on, unplugs-with, unplug-from, unplugs, draws-out,disconnects-in, disconnect-with, disconnects-to, disconnect-from, disconnects-from, disconnects-on, disconnect, disconnects-with, disconnects                                                                   \\
\textbf{lift}     & lifts-by, lifts-toward, lift-off, lift-into, lifts-alongside, liftss-with, lifts-beneath, lifts-out, lifts-to, lift-from, lifts-with, lifts-off, lifts-towards, lift-in, lifted-on, liftes-with, lifts-into, lifts-on, uplifts, lifts, lifts-over, lifts-above, lifts-down, lifts-up, lifts-across, lifts-at, lifts-beside, lifts-onto, lifts-around, lifts-in, lifts-for, lift-up, lifts-under, lift, lift-on, lifts-of, lifts-from, lift-with,  raise-in, raises-over, raised-with, raise, raises-toward, raises-above, raise-off, raises-at, raises-towards, raise-with, raises-up, raise-towards, raises-along, raises-beneath, raises, raises-for, raises-off, raises-with, raises-by, raised, raises-from, raises-of, raises-in, raises-on, raises-around, raises-underneath, raises-beside, raises-across, raises-down, raises-to, raise-to, puts-up \\
\textbf{open}     & opens-wiith, opens-through, opens-up, opens-in, opens, opens-opposite, sheopens, opens-below, opening-with, opens-underneath, open-on, opening, opens-beneath, sheopens-with, opens-over, opens-by, opens-atop, opened-in, opens-for, open-up, openst, opens-forth, opens-at, opens-onto, opens-inside, opens-back, opens-outside, opens-with, opens-around, flips-open, opens-behind, opens-to, open, opened-with, opens-down, opens-on, opens-under, open-with, opens-after, open-beside, opened, opents, opens-beside, opens-along, opens-from, opens-near, opens-out, open-in, opens-into \\
\textbf{pull}     & pull-in, pulls-for, pulls-past, pulls-back, pulls-down, pulls-with, pull-on, pulls-beside, pulls-by, pulls-on, pulls-through, pulls-in, pulled-beside, pulls-under, pulls-across, pullout, pulls-out, pulls-inside, pulls-up, pulling-from, shepulls, pulls-rom, pulls-around, pulls-towards, pulls-at, pulls-against, pulls-behind, pulls-near, pull-from, pulling-on, pulling-through, pulls-toward, pulls-wit, pulls-into, pull-with, pulls-after, pull-up, pull-out, pulls-from, pulls, pulls-over, pulls-along, pulls-to, pulls-outside, pull, pulls-beneath,pulls-out, pull-out, pulls-outside, pullout \\
\textbf{turn}     & turned-to, turns-towards, turn-over, overturns-in, turns-up, turn, turns-inside, overturns, turns-toward, turning-inside, turns, turns-with, upturns, turned-on, turning-over, turns-underneath, turns-out, turns-onto, overturns-alongside, turning-in, turn-out, turn-at, turns-from, overturns-with, turn-with, turns-about, turns-at, overturn, turn-around, upturns-on, turns-behind, overturns-atop, turns-into, turns-back, upturns-in, sheturns, overturns-on, turning-on, turns-over, sheturns-around, turn-towards, turn-to, turns-above, turns-beside, turn-in, turns-down, turns-by, turned, turns-to, turns-around, turns-in, turns-under, overturn-from, turns-outside     \\ 
\textbf{press}    & presses-above, pressesthe, pressing-onto, pushes-along, spress, presses, pushes-in, press-into, pushes-towards, pushes-at, pushes-out, pushes-up, press, pushes-on, compresses-on, presses-on, presses-for, pushes-round, compress-with, pushes-beside, pushes-behind, pushes-around, press-down, push, compresses-into, pushes, push-with, presses-to, pressed-into, pushes-beneath, press-with, compresses-with, pushes-over, compressor, presses-off, pushes-onto, push-up, pushes-outside, pushes-through, presses-under, push-out, pushing-out, pushed-into, push-to, pushes-underneath, pressure-up, presses-inside, push-on, pushers-in, pushes-under, pushes-inside, pushes-off, presses-from, pushes-from, presses-around, presses-by, presses-round, pushes-down, pushes-across, presses-against, pushes-by, pushes-past, presses-wit, pushes-toward, pushes-against, press-in, pushes-to, presses-atm, presses-in, presses-beside, presses-up, press-on, pushes-away, pushes-back, press-to, push-in, presses-down, presses-at, presses-like, pressing-with, presses-out, compresses-in, compresses, presses-onto, pushes-with, pushes-before, pressing, presses-into, compresses-inside, press-up, pushesb, presses-over, presses-with \\
\textbf{turn-off} & turn-off, turns-around, turns-down, turns-in, turns-of, turns-on, turns-to, turns-towards, switch-off, switched-off, switches-off  \\
\textbf{throw}    & throws-toward, throws-to, throw-up, throws-between, throw-down, throw-with, throws-across, throws-beneath, throws-towards, thrown-inside, throws-in, throwing-on, throws-up, throws-outside, throws-inside, throws-unto, throws-from, throw-on, throws-of, throw, thrown-to, throws-with, throws-by, throws-into, thrown-in, throws-under, throws-back, throws-off, throws-out, throws-over, throws-on, throws-down, throws-onto, throw-in, throws, throws-away, cthrows, throws-underneath, throw-into, shethrows, throw-from, throws-behind, throw-to, throws-through, throws-beside, throws-at  \\
\textbf{wash}     & cleanses, cleans-inside, clean-beneath, cleans-on, cleans-outside, cleanses-in, cleanses-with, clean-off, cleans-across, cleans-along, clean-in, cleanssink-with, cleans-at, cleaning, cleaning-on, cleans-over, cleanses-on, cleans-off, cleans-underneath, cleans-under, cleans-out, cleans-beneath, cleans-against, clean, cleans-near, cleans-to, cleans-in, cleans, cleans-back, cleans-from, cleans-with, cleans-by, cleans-beside, clean-with, cleans-around, cleansthe, cleaning-with, cleanes-with, cleans-below, cleans-towards, cleans-of, cleans-down, rinse-under, rinses-at, rinse-with, rinses-from, rinses-inside, rinses-on, rinses-underneath, rinses-under, rinse, rinses-over, rinses, rinses-with, rinses-through, rinses-off, rinses-in, rinse-in, sherinses-on, rinsed-in, rinses-out, rinses-of, rinses-into, washes-by, washes, washes-off, washing-with, wash, washed-with, washing-on, washes-in, washes-at, wash-with, washes-beside, washes-behind, washing-in, washes-through, washes-inside, washes-with, washes-under, washes-underneath, washed, washes-out, washes-on, washers-in, washes-into, washing, washed-inside, wash-in, washes-from    \\
\textbf{pour}     & pour-to, pours-across, pours-around, pours-with, pours-from, pours-inside, pouring-on, poured-on, pours-outside, pours-of, pours-over, pours-on, pours-off, pours-into, pour-in, pours-down, pour-from, pours-fro, pours-back, pours, pour-on, pours-in, pour-into, pours-to, pours-towards, pours-away, pours-out, pours-through, pour, pours-onto, pours-between, poured-into, pours-at, poured-in, pours-along,sieves-into,sieve-in, sieves-in                                                                                                \\
\textbf{close}    & close, closes-back, close-on, closes-into, closes-near, closes-in, closes-from, closes-opposite, closes-onto, closed-with, close-up, closes-beneath, closes-at, closes, encloses-within, closes-with, encloses-with, closes-under, close-in, closes-on, closes-withy, closes-of, closes-atop, encloses, closes-beside, closes-up, closes-to, close-with, closes-behind, closes-through, closed, closes-above, closes-by \\
\textbf{pat}      & hits-between, hits, hits-through, hits-inside, shits-in, hits-down, hits-onto, hits-against, hits-behind, hits-towards, hits-on, hits-to, hits-from, hits-around, hits-with, hits-beneath, shehits-against, hits-at, shits, hits-in, hits-into, hits-out, hit-in, hit-with, hits-beside, shits-from, hit, hitting-with, hits-under, hit-on, hits-over, hits-by, taps-from, untaps, taps-into, taps-to, taps-onto, tap-on, staps-with, taps-by, taps-against, taps, tap, taps-at, taps-on, taps-around, taps-beside, taps-with, taps-in \\
\textbf{cut}      & cutes, cuts-beside, cut-into, cuts-of, cuts-wit, cuts-along, cutting-on, cutting-around, cuts-at, cuts-up, cutting-out, cuts-unto, cuts, cuts-outside, cuts-from, cut-in, cutting-into, cuts-into, cuts-through, cuts-out, cut-under, cuts-between, cuts-with, cut-with, cutting, cuts-around, cut, cuts-to, cuts-by, cuts-in, cut-out, cut-on, cut-off, cutting-with, cuts-without, cuts-on, cuts-under, cuts-across, cutback-with, cuts-off, cut-from, cuts-down, cuting, cuts-over, cute, cuts-inside, chopping, chops, chops-off, chopped-on, chop-with, chops-from, chops-with, chops-to, chops-over, chopped-from, chops-into, chops-in, chopping-on, chops-on, chops-at, chop, chopped,  slices-in, slices-by, slices-to, slices-with, slice, slices-through, slices-from, slices-onto, slices-on, slices-inside, slices-into, slices-off, slices, trim, trimming-down, trims-on, trims, trims-off, trims-to, trims-in, trims-with, trims-into, trims-around, trims-out, trims-from, trims-at, trim-with    \\
\textbf{carry}    & carried, carries-wiith, carries-unto, carries-beside, carriy, carries-on, carrirs, carries-down, carries, carries-with, carriers-up, carries-underneath, carriers-with, carries-under, carries-into, carries-off, carrries, carries-by, carries-at, carries-through, carries-outside, carrying-from, carriers, carries-around, carries-up, carried-on, carrying, carriers-from, carries-towards, carry-in, carries-in, carries-over, carries-to, carries-between, carries-out, carrues, carries-of, carry, carries-along, carries-from, carries-across, carried-from, carrys  \\
\textbf{clear}    & wipes-across, wipes-in, wipes-off, shewipes-in, wipes-up, wiped-on, wipes-beneath, wipes-by, wipes-inside, wipes-around, wipes-into, wipes-of, wiped-with, wipe, wipe-off, wipes-as, wipes-at, wipes-out, wipes-from, wipes-with, wipes-under, wipes-behind, wipes-over, wipes-against, wipes-underneath, wipes, wipe-in, wipes-down, wipes-onto, wipes-on, wipes-to, wipe-with, clears-with, clear-in, clears-inside, clears, clears-from, clearing, clears-by, clears-on, clears-under, clears-beside, clears-in, clears-of, clears-before, clears-around, clears-to, clear, clear-with, clearing-with, clears-out, clears-off        \\  
\textbf{rub}      & rubs-off, rubs-through, rubs-into, rubs-onto, rubs-in, rub, rub-between, rubs-alongside, rub-against, rubs-over, rubs-v, rubbed-on, rubs-aganist, rubs-under, rub-with, rubs-above, grubs-from, rubs-from, rubes-on, rubs-up, rubbing-with, rubs-across, rubs-at, rubs-of, rubs-against, rubs-inside, rubs-between, rubs-with, grubs-with, rubs-before, rubs-by, rubs, rub-on, grubs, rubbing, rubs-on, rubs-to, rubs-around, rubbing-on, scratches-off, scratches-in, scratches-by, scratches-behind, scratches-from, scratches-to, scratchers, scratches-on, scratches-between, scratch, scratchs, scratches-with, scratchs-with, scratches, scratch-with, scratch-off  \\
\textbf{fold}     & fold-into, folding, folds-under, folds-over, folds-on, folds-into, folds-down, folds-onto, refolds, folds-off, folds-with, folds-out, folds-above, ufolds, fold, folds-across, folds, folds-to, folds-back, folds-inside, fold-with, folds-from, folds-up, folds-at, folds-against, folds-around   \\
\textbf{gather}   & gathered-on, gathers-into, gathers-to, gathers-near, gather-with, gathers-around, gather-on, gathers-behind, gathers-on, gathers-inside, gathers-round, gathers-with, gather-in, gathers-under, gathers-over, gather, gathers-in, gathered-with, gathers-onto, gathers-from, gathers-up, gathers-out, gathers, collect-with, collect, collects-by, collects-inside, collect-from, collects-from, collects, collects-into, collects-with, collects-in, recollects, collects-to, collects-on  \\
\textbf{stretch}  & stretches-around, stretches-into, stretches-towards, stretch-out, stretches-to, stretches-with, stretches-down, shestretches, stretches-in, stretchers, stretches-from, stretchers-outside, stretches-across, stretches-before, stretches, stretchs, stretches-out, stretches-above, stretches-under, stretches-along, stretches-for, stretches-up, stretch-with, stretchers-towards, stretches-over, stretch, stretches-on, stretches-outside, stretches-at, stretches-toward, stretches-past \\
\textbf{attach}   & attaches-onto, reattaches, attaching-to, attaches-into, attaches-under, attach, attaches-against, attaches-to, attaches-from, attaches-with, reattaches-to, attaches-underneath, attach-on, attaching-in, attaches-in, reattaches-on, attach-to, attaches, attaches-on, attaches-by, reattaches-in,connects-in, connect-with, connects-from, connects-into, connects-to, connect-to, connects, connects-through, connect-from, connected-on, connects-with, connects-on, connect, inserts-into, insert-in, inserts-at, inserts-to, insert-with, inserts-inside, inserts-under, inserts-back, reinserts-to, inserts-by, inserting-into, inserts-between, insert-on, reinserts, inserts-in, inserts-through, inserts-beneath, inserted-to, inserts-below, inserts-with, inserts-against, inserts-on, insert-into, inserts, insert, inserted-in, inserts-from, inserts-onto, inserts-near, reinserts-into, inserts-around, plugs-into, plug-in,plugs-inside,plugs-in,pushes-in,pushed-into,pushers-in,pushes-inside,push-in,pushes-into\\
\textbf{flip}     & flips-by, flips-onto, flips-from, flips-inside, flips-off, flips-back, flips-on, flip-on, flips-over, flips-with, flips, flips-in, flips-between, flips-amongst, flips-out, flips-through, flips-open, flips-at, flip-in, flipping-on, flip, flips-towards, flips-forward, flips-up, flips-around, flips-down, flips-into, flipping, flips-to,turn-over,turning-over,turns-over \\
\textbf{turn-on}  & turn-on,turned-on,turns-on,turning-on,switches-on,switch-on \\
\textbf{scoop}    & scoops-up, scoops, scoopes-with, scoops-onto, scoops-on, scoops-by, scoops-from, scoop-in, scoops-into, scoops-with, scoops-off, scooped-with, scoop-into, scoop-with, scoops-inside, scoops-in, scoop-out, scoops-to, scoop-from, scoops-out, scoop, scoops-wit   \\
\textbf{shake}    & shake-off, shake, shakes-with, shakes-over, shakes-between, shakes-in, ashaker, shakes-out, shakes-above, shakes-by, shakes-under, shakes-into, shakes, shakes-of, shakes-off, shake-with, shakes-inside, shakes-from, shake-in, shakes-on, shakes-at, shakes-around   \\
\textbf{bend}     & bends-beside, bends-up, bends-down, bends, bends-at, unbends, bends-near, bends-in, bends-to, bends-towards, bends-against, bends-toward, bends-by, bends-over, bends-around, bends-into, bends-beneath, bends-behind, bends-on, bends-out, bends-along, bends-forward, bend-towards, bends-under, bends-with  \\
\textbf{dip}      & dips-from, dips-into, dips-onto, dip-into, dips-with, dipped, dips-inside, dips-in, dips-beside, dips-beneath, dips-on, dipping-in, dip-in, dips, dips-under, dips-to, dips-through, shedips-in \\
\textbf{roll}     & rolls-across, rolles, rolls-in, rolls-round, rolled, rolling, roll-out, rolls-under, rolls-inside, roll, rolls-on, rolls-through, strolls-on, strolls, rolls-against, rolls-around, rolls-onto, roller-in, rolls-down, rolls-back, rolls-up, rolls-off, rolls-of, rolls, rollls-on, roll-with, rolled-into, rolls-to, rolls-between, srrolls-on, rolls-out, rolls-over, roll-on, rollers, rolls-with, rolls-towards, rolls-into, rolls-from, rolling-into   \\
\textbf{wrap}     & wraps, wrapps, wraps-from, wraps-in, rewraps-around, wraps-around, wrapping-around, wrap, wrapper, wrap-on, wraps-up, wraps-on, wraps-inside, wraps-round, wraps-at, wraps-over, wrapped-on, wraps-with, covers-up, coverses-with, covers, covers-back, covers-of, coverers-with, covers-by, covers-on, cover-in, cover-with, covers-in, covers-down, recovers, cover, covers-from, covers-with, rolls-up \\    
\textbf{lower}    & lowers-with, lowers-from, lowers-underneath, flower-from, lowers-to, lowers-over, lowers-unto, lowers, lowers-onto, lowers-into, lowers-above, lowers-towards, lowers-along, lowers-down, lowers-in, lowers-toward, lowers-by, lower-from, lowers-under, lowers-on                                                                                               \\
\textbf{drink}    & drinks, drinking, drinks-out, drink-from, drinks-in, drinks-on, drinking-from, drinks-from, drink, drinks-with, drinks-up, drink-on \\
\textbf{spread}   & spreads-inside, spreads-around, spreads-from, spreads-in, spread-in, spreads-to, spreads-out, spreads-at, spreads-into, spreads-on, spreads-down, spread, spreads-with, spread-with, spreads-up, spreads-under, spreads-beside, spread-from, spread-out, spread-around, spreads-over, spreads-by, spreads-onto, spreads, spreading-on, spread-on, spreads-towards, spreads-across \\
\textbf{drag}     & drags-down, drags, drag, drags-with, drag-in, drags-on, drags-into, drags-inside, drags-along, drag-with, drags-from, drags-in, drags-around, drags-to, drags-across, drags-off, drags-out, drags-towards, drags-beneath, drags-through, drags-up \\
\textbf{mix}      & mixes-from, mixing, mixing-on, mixes-inside, mixes-into, mixes-in, mix, mix-with, mixes-with, mixed-with, mixes-on, mixes-up, mixing-with, mixex-with, mixes-by, mixes, stirs-inside, stirring-with, stirrs, stirs-on, stirs, stirs-from, stirs-into, stir-with, stirring-in, stirred-in, stirs-with, stirs-in, stir-on, stir-in, stirring-on, stirring, stirring-inside, stir, stirred-with, stir-into, stirs-around, stirs-up, stirs-at, whisks-on, whisks-in, whisks-with, whisks,scrambles-in, scrambles, scrambles-for, scrambles-with,folds-in,fold-in  \\
\textbf{wear}     & wears-for, wears-before, wear, wears-in, wears, wears-under, wear-in, wears-around, wears-to, wears-from, wears-v, wears-onto, wears-with, wears-back, wears-on, wears-beneath, wears-by, wear-on   \\
\textbf{divide}   & separate-with, separates-in, separates-with, separates-over, separates-on, separate, separates-into, separates-from, separated-with, separates-to, separates, separated-from, separate-from, separate-for, detaches-on, detaches-rom, detaches-with, detach, detaches-off, detaches-from, detaches-into, detach-from, detaches-in, detached, detaches-back, detaches, detaches-behind, detaches-at, splits-with, splits-into, splits-from, splits-on, splits-to, split-with, splits-in, splits, splits-by, splits-inside  \\
\textbf{eat}      & eats, eats-from,eats-at,eat,eats-off,eats-with,eats-in,eats-on,eats-out,bites-in, bites, bite, bites-with, bites-from \\
\textbf{bring}    & brings-under, brings-up, brings-on, brings-in, bring-out, brings-from, bring-from, brings-down, brings-towards, brings-out, bring-on, brings, brings-to, brings-with \\
\textbf{hang}     & hangs-by, hangs-with, hang-on, hangs-to, hangs-onto, hangs-from, hangs-over, hangs-at, hangs-back, hangs, hangs-up, hangs-beside, hangs-inside, hanged, hang, hangs-on, hangers-on, hangs-in, hang-in, hangs-against, hanging, hangers   \\
\textbf{read}     & reads-back,read,reads-at,reads-with,reads-through,reads-out,reads,read-on,reads-in,reads-to,reads-on,reading,reads-from,reads-to \\
\textbf{scrape}   & scrapes-from, scrapes-round, scrapes-into, scrapes-inside, scrapes-on, scrapes-beside, scrapes-out, scrapes-against, scrapes, scrapes-underneath, scrapes-at, scrape-on, scrape, scrapes-through, scraped, scrape-in, scrapes-in, scrapes-beneath, scraped-on, scrapes-off, scrapes-with, scraped-in, scrapes-onto, scraper-with, scrapes-of, scrape-from, scraped-inside, scrapes-to  \\
\textbf{brush}    & brush-on, brushes-from, brushes-over, brush-with, brushes-against, brushes-onto, brush, brushes-into, brushes, brushes-on, brushes-through, brushes-across, brush-through, brushes-with, brushes-to, brushes-off, brushes-under, brushes-in,sweeps-out, sweeps-back, sweeps-off, sweeps-into, sweeping, sweeps-of, sweeps-wit, sweeps-towards, sweeps-inside, sweeps-on, sweeps-onto, sweeps-along, sweeps-under, sweeps-with, sweeps-from, sweeps-behind, sweep-on, sweeps-to, sweeping-outside, sweeps-outside, sweeps-in, sweeps, sweep-into, sweep-with, sweeps-down, sweeps-around, sweeps-across, sweeping-with, sweep  \\
\textbf{screw}    & tightens-under, tightens-in, tightens-into, tightens-from, tightens-to, tightens, tighten-at, tighten, tightens-at, untightens, tightens-on, tightens-with, untightening, tightening, tightening-with, tightens-against, tighten-with, tighten-to, tighten-on, tightens-behind, tightens-around, tightens-up, tightens-underneath, screws-up, screw-from, screw, screws-on, screws-out, screws-at, screws-with, screw-through, screws-back, screws-under, screws-into, screws-to, screws-through, screw-into, screwing-with, screw-with, screw-on, screws-beneath, screw-in, screws-onto, screws-inside, screws-in, screws, screwed-to \\
\textbf{squeeze}  & squeezes-through, squeezes-from, squeezes-against, squeeze-with, squeeze, squeezed-out, squeezes-up, squeezes-between, shesqueezes-with, squeezes-to, squeezes, squeezes-inside, squeezes-on, squeezes-onto, squeeze-in, squeeze-on, squeezes-into, squeezes-with, squeezes-over, squeezes-out, squeezes-in, squeezes-around, squeezes-under \\
\textbf{scrubs}   & scrub, scrubs-off, scrubs-in, scrubs-with, scrubs-by, scrubs-on, scrubbing-with, scrubs-beside, scrubs-from, scrubs-under, scrubs-into, scrub-with, scrubs-inside, scrubs-of, scrubs, scrubs-beneath, scrubs-out  \\
\textbf{unroll}   & unfolds-in, unfolds-to, unfolds, unfolding, unfolding-with, unfolds-on, unfold, unfolds-from, unfolds-with, unfolds-around,unrolls-in, unrolls-with, unrolls-on, unrolls, unrolls-from        \\
\textbf{give}     & gives-on, give-in, gives-from, gives-in, gives-with, give-to, gives-back, gives-up, give-with, gives, gives-to, give, gives-through, gives-out, gives-w, gives-towards \\
\textbf{draw}     & draws-back, draws, draws-across, draw-across, draws-around, draws-on, draws-near, draws-down, draws-in, draws-by, drawing-on, draw, draws-through, draw-on, draws-out, draws-from, draws-along, drawing, draws-as, draws-up, draws-above, draws-at    \\
\textbf{loosen}   & loosens-by, loosen-on, loosen, loosens-from, loosens-against, loosening-with, loosens-round, loosens-behind, loosens-around, unloosens, loosens-at, loosening-on, loosen-with, loosens-into, loosens-out, unloosens-from, loosens, loosens-up, loosens-under, loosens-on, loosens-in, loosens-with  \\
\textbf{break}    & breaks-up, break-on, break-with, breaks-of, breaks-at, breaks, breaks-unto, break, breaks-on, break-off, breaks-in, breaks-down, break-apart, breaks-for, breaks-into, breaks-off, breaks-by, breaks-from, breakes, breaks-with, breaks-out                                                                                                 \\
\textbf{peel}     & peeling, peels-into, peel, peels-with, peels-from, peel-out, peels-of, peels-over, peeling-from, peels-on, peels-under, peels-out, peels-in, peels, peels-off, peels-onto, peels-around  \\

\textbf{paint}    & painting-by, paint-from, paints-opposite, paints-to, painting-in, paints-around, paints-in, paints-onto, painting-above, paint-with, paint-inside, paint-to, painting-near, paints-from, paint, paint-on, paints-beside, paints-beneath, paint-in, painted-with, paints-inside, paints-before, paints-on, painting-around, paints, paints-at, paints-with, painting, paints-over, painting-with, paintboard, paints-by, paints-near, paintss-on, paints-above, paint-down, painting-on \\
\textbf{rip}      & tears-on, tear-on, tears-in, tears-by, tears-apart, tears-under, tears-off, tears-into, tears-up, tears, tears-around, tears-with, intearacts-with, tears-inside, tears-from, tears-out,rips-in, ripping, trips-with, rippes-with, rips-off, rips-with, rips-into, rips                                                                                                   \\
\textbf{sprinkle} & sprinklers, sprinklers-on, sprinkles-from, sprinkles-into, sprinkle-on, sprinkles-in, sprinkles, sprinkles-with, sprinkle-inside, sprinkles-to, sprinkle, sprinkle-into, sprinkles-on, sprinkle-from, sprinkles-before, sprinkles-onto, sprinkles-beside, sprinkles-over       \\

\textbf{drill}    & drilling-with, drill-with, drills-around, drills-onto, drills-into, drilling-on, drills-to, drilling-through, drills, drills-inside, drills-out, drilling, drills-under, drill-on, drill-to, drills-with, drills-through, drill, drills-across, drills-up, drill-into, drills-in, drills-on, drills-by, drill-in, drill-from, drilled  \\

\textbf{unwrap}   & unwraps-on, unwraps, unwraps-around, unwraps-over, unwraps-from, unwraps-with, unwraps-in, unwrap        \\                                                     \bottomrule                                                                   
%\end{tabular}
\caption{\dataset\ action classes and their corresponding open-vocabulary verbs. }
\label{tab:verbs}
%\label{tab:verbs}
\end{tabularx}
%\end{multicols*}{1}
\clearpage
\twocolumn

%%%%%%%%% REFERENCES

 \end{document}